%% file: main_ArXiv_xelatex.tex
\documentclass[11pt]{article}

\usepackage[preprint]{acl}

 \usepackage{microtype}

\usepackage{inconsolata}

\usepackage{graphicx}
\usepackage{tcolorbox}
\usepackage{courier}
\usepackage{subfig} 
\usepackage{ragged2e}
\usepackage{amssymb}
\usepackage{colortbl}
\usepackage{booktabs}
\usepackage[dvipsnames,x11names]{xcolor}
\usepackage{hyperref}
\hypersetup{
    colorlinks=true,
    linkcolor=Blue,
    filecolor=Blue,
    urlcolor=Magenta,
    citecolor=Blue,
    }
\usepackage{array}
\usepackage{longtable}
\usepackage{xspace}
\usepackage{amsmath}
\usepackage{multirow}
\tcbuselibrary{skins}
\usepackage{varwidth}
\usepackage{subcaption}
\usepackage{ulem}
\usepackage{balance}
\usepackage{fontawesome5}

\usepackage{enumitem}

\usepackage{fontspec}

\usepackage[english]{babel} 

\babelfont{rm}[Path = ./, Extension = .otf, UprightFont = tgreg, BoldFont = tgbold, ItalicFont = tgit, BoldItalicFont = tgbit]{texgyretermes}

\babelprovide[import]{hindi}
\babelfont[*devanagari]{rm}[Path=./, Extension=.ttf]{lohit_dev}

\babelprovide[import]{bengali}
\babelfont[*bengali]{rm}[Path=./, Extension=.ttf]{lohit_ben}

\babelprovide[import]{urdu}
\babelprovide[import]{arabic}

\babelfont[*arabic]{rm}[Path=./, Extension=.ttf, Scale=0.75]{scheherazade}

\newcommand{\new}[1]{\textcolor{black}{#1}}

\newcommand{\xhdr}[1]{\vspace{1mm}\noindent{{\bf #1.}}}
\newcommand{\xhit}[1]{\vspace{1mm}\noindent{{\it #1:}}}

\newcommand{\secref}[1]{\S\ref{#1}}

\newcommand{\llamaS}[0]{{\tt LAM-8B}\xspace}
\newcommand{\mistralS}[0]{{\tt MST-7B}\xspace}
\newcommand{\mistralM}[0]{{\tt MST-24B}\xspace}
\newcommand{\llamaM}[0]{{\tt LAM-70B}\xspace}
\newcommand{\avg}[0]{{\tt Average}\xspace}

\newcommand{\mam}[0]{{\tt \textbf{MAM}}\xspace}
\newcommand{\sem}[0]{{\tt \textbf{SEM}}\xspace}
\newcommand{\sgm}[0]{{\tt \textbf{SGM}}\xspace}

\newcommand{\en}[0]{{\tt EN}\xspace}
\newcommand{\de}[0]{{\tt DE}\xspace}
\newcommand{\hi}[0]{{\tt HI}\xspace}
\newcommand{\bn}[0]{{\tt BN}\xspace}
\newcommand{\ur}[0]{{\tt UR}\xspace}

\newcommand{\mtrex}[0]{{\bf mTREx}\xspace}
\newcommand{\gmmlu}[0]{{\bf G-MMLU}\xspace}

\newcommand{\metric}[0]{\texttt{TPHR}\xspace}

\title{Do LLM hallucination detectors suffer from low-resource effect?\thanks{~This work has been accepted at EACL 2026 (Main).}}

\author{
\textbf{Debtanu Datta,\textsuperscript{1}}
\textbf{Mohan Kishore Chilukuri,\textsuperscript{1}}
\textbf{Yash Kumar,\textsuperscript{1}}\\
\textbf{Saptarshi Ghosh,\textsuperscript{1}}
\textbf{Muhammad Bilal Zafar\textsuperscript{2,3} }\\ \\  
\textsuperscript{1}Indian Institute of Technology Kharagpur, India
\\
\textsuperscript{2}Ruhr University Bochum, Germany,
\\
\textsuperscript{3}UAR Research Center for Trustworthy Data Science and Security, Germany \\
 \small{
   \textbf{Correspondence:} \texttt{\href{mailto:debtanudatta04@gmail.com}{debtanudatta04@gmail.com}}
 }
}

\begin{document}
\maketitle


\input{00_abstract}

\input{01_intro}

\input{02_related_work}
\input{03_datasets}

\input{04_HD_methods}

\input{05_experiments}
\input{06_conclusion}

\clearpage
\input{07_limitations_ethics}


\balance
\bibliography{main}


\clearpage
\noindent {\bf \large Appendix}
\appendix

\input{08_appendix}

\end{document}

%% file: 00_abstract.tex
\begin{abstract}

LLMs, while outperforming humans in a wide range of tasks, can still fail in unanticipated ways. We focus on two pervasive failure modes: (i)~hallucinations, where models produce incorrect information about the world, and (ii)~the low-resource effect, where the models show impressive performance in high-resource languages like English but the performance degrades significantly in low-resource languages like Bengali. 
We study the intersection of these issues and ask: do hallucination detectors suffer from the low-resource effect?
We conduct experiments on \textit{five tasks} across \textit{three domains} (factual recall, STEM, and Humanities).
Experiments with \textit{four LLMs} and \textit{three hallucination detectors} reveal a curious finding: As expected, the task accuracies in low-resource languages experience large drops (compared to English). 
However, \textit{the drop in detectors' accuracy is often several times smaller than the drop in task accuracy}.
Our findings suggest that even in low-resource languages, the internal mechanisms of LLMs might encode signals about their uncertainty. 
Further, the detectors are robust within language (even for non-English) and in multilingual setups, but not in cross-lingual settings without in-language supervision.

\texttt{\faGithub \  \url{https://github.com/aisoc-lab/low-resource-hallucination-detection}}

\end{abstract}




%% file: 01_intro.tex
\section{Introduction}
\label{sec:intro}

\begin{figure}[t]
\centering
\includegraphics[width=0.85\linewidth, height=4.55cm]{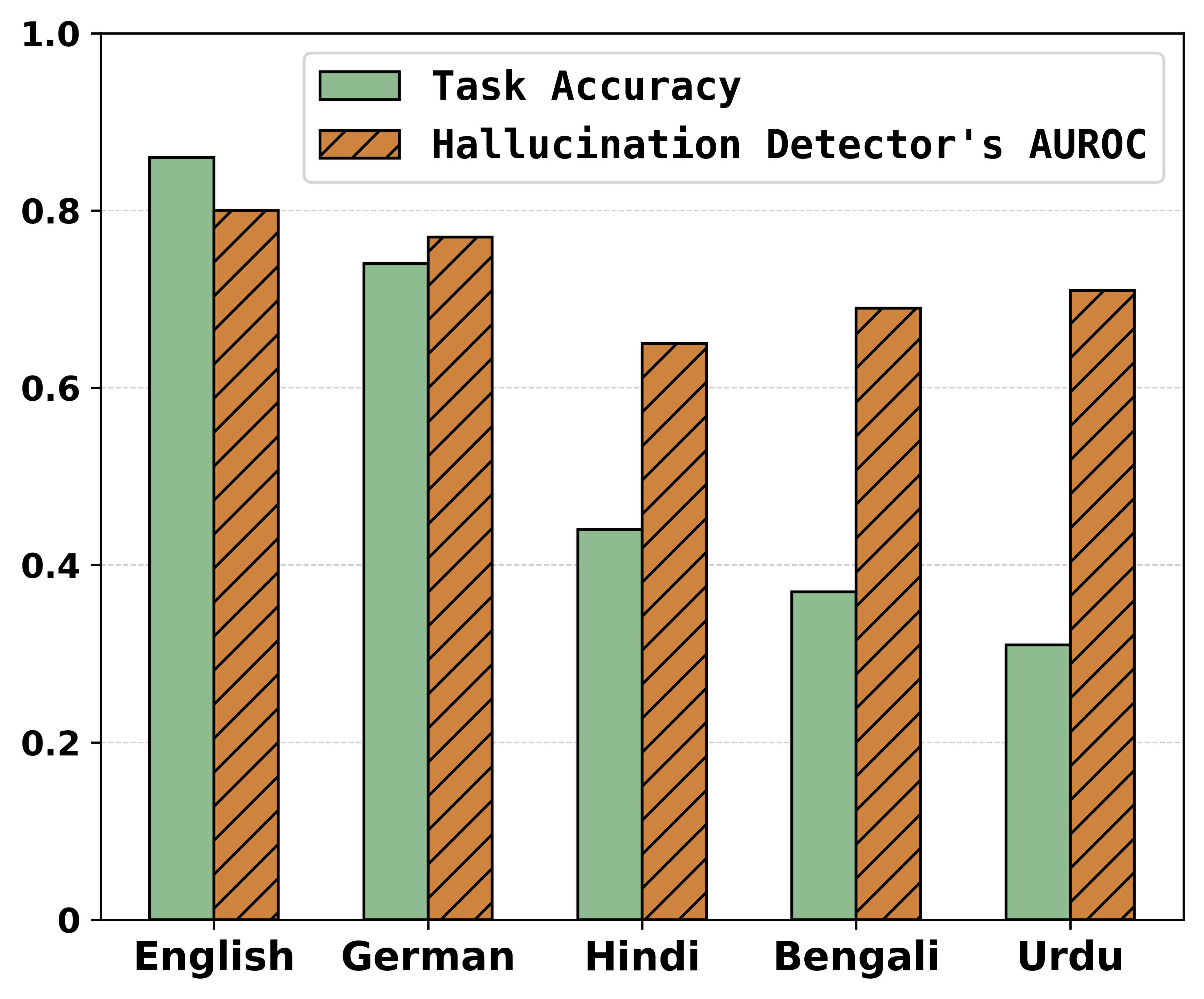}
\vspace{-2.5mm}
\caption{Comparison of task accuracy ($1$ denotes a correct answer, $0$ denotes a hallucination) of a model (\llamaM) with hallucination detector's (\mam) performance across languages on \texttt{mTREx-Capitals} dataset. When going from English to lower-resource languages, the task accuracy drops significantly. But the hallucination detector's performance remains relatively stable.}
\label{fig:task-hallu-comparison}
\vspace{-6mm}
\end{figure}

LLMs have demonstrated remarkable capabilities across a wide range of tasks like web search~\cite{google_ai_search}, coding~\cite{Peng2023TheIO} and scientific research~\cite{gottweis2025aicoscientist,shao-etal-2024-assisting}. In  these applications, the correctness of answers and the ability of models to cater to various languages is of paramount importance. However, a long line of work suggests that LLMs face issues along both  dimensions.

First, although LLMs generate highly fluent and coherent text, this fluency often masks a critical problem: LLMs can confidently generate inaccurate information, a phenomenon commonly referred to as \textit{hallucination}~\citep{ji2023survey, xiao-wang-2021-hallucination}. This behavior severely undermines user trust and the reliability of LLMs, and has prompted a flurry of research on detecting and mitigating hallucinations~\cite{simhi2025trustmeimwrong, snyder_early_2024, semantic_entropy_paper}.

Second, the performance of LLMs varies significantly across languages. Models that perform well in high-resource languages like English often degrade substantially in low-resource ones~\citep{low-resource-paper}, a phenomenon often referred to as the \textit{low-resource effect}. Despite the advancement of multilingual pretraining and alignment, significant performance gaps persist. A number of prior studies have focused on addressing this issue~\citep{song2025llmsilverbulletlowresource, jiang-etal-2020-multilingual}.

In this work, we study the intersection of these two issues. Specifically, we ask: 
\vspace{-1.5mm}
\begin{center}
    \textbf{If LLMs hallucinate more in low-resource languages, do hallucination detectors perform equally worse in these languages?}
\end{center}
We conduct extensive experiments on \textit{five question-answer (QA) tasks} 
over \textit{five languages} -- English (\en), German (\de), Hindi (\hi), Bengali (\bn), and Urdu (\ur) -- with \textit{four LLMs} 
of varying sizes (ranging from 7B to 70B parameter models).
Due to the lack of multilingual datasets, we translate the three factual recall tasks from English into the four other languages.
These five languages span \textit{different resource levels} and \textit{geographic regions}, and are widely spoken across the world. Moreover, they use a \textit{typologically diverse set of four scripts}: Latin, Devanagari, Perso-Arabic, and Bengali.
We tested \textit{three hallucination detectors} (HD methods) from two popular families: (i)~methods that examine the model's internal artifacts during generation, and (ii)~sampling-based black-box methods that utilize model responses only.

We find that although the task accuracy of LLMs drops sharply for low-resource languages, the degradation in HD performance is often much smaller (Figure~\ref{fig:task-hallu-comparison}). In fact, in some cases, the HD performance is even better than in English. To further quantify this phenomenon, we design a metric, \metric, that compares the drop in task accuracy with the drop in detection performance across languages relative to the \en baseline.
To summarize, our key contributions are:
\begin{enumerate}[leftmargin=*, nolistsep]
    \item Due to the scarcity of factual QA datasets in low-resource languages, we developed a novel multilingual factual QA benchmark, \mtrex (detailed in \secref{subsec:mtrex-main}), by translating the original English text into \de, \hi, \bn, and \ur. Figure~\ref{fig:hallu-intro} shows an example in all languages. 
    \item We perform a comprehensive evaluation of three HD methods using four LLMs over five tasks across five languages. 
    \item We further conduct extensive 
    cross-lingual and multilingual analyses with ablation studies. This provides new insights into HD methods' behavior, demonstrating that detectors perform poorly in pure cross-lingual transfer, but the multilingual training with in-language supervision mitigates the performance gap. 
    
    \item We introduce the \metric metric to compare task accuracy with HD performance across languages, which further validates that HD is relatively robust even when task accuracy drops significantly in low-resource languages.
    \item We find that hallucination detectors that utilize the model's internal artifacts outperform sampling-based black-box methods, even in multilingual and low-resource settings. 
\end{enumerate}


%% file: 02_related_work.tex
\begin{figure}[t]
\centering
\begin{tcolorbox}[colback=gray!15, colframe=black, boxrule=1pt, fontupper=\normalfont\small, boxsep=1pt, left=2pt, right=2pt, top=1pt, bottom=1pt, halign=justify, enhanced]
\textbf{Question (\en):} What is the capital of Irion County?
\\
\textbf{Ref. Ans. (\en):} Mertzon
\\
\textbf{Model Ans.:} \textcolor{ForestGreen}{Mertzon (\textbf{no hallucination})}
\tcbline
\textbf{Question (\de):} Was ist die Hauptstadt von Irion County?
\\
\textbf{Ref. Ans. (\de):} Mertzon
\\
\textbf{Model Ans.:} \textcolor{ForestGreen}{Mertzon (\textbf{no hallucination})}
\tcbline
\textbf{Question (\hi):} {\fontsize{8}{8}\selectfont \foreignlanguage{hindi}{इरियन काउंटी की राजधानी क्या है?}}
\\
\textbf{Ref. Ans. (\hi):} {\fontsize{8}{8}\selectfont \foreignlanguage{hindi}{मर्टज़ोन}}
\\
\textbf{Model Ans.:} {\fontsize{8}{8}\selectfont \textcolor{Red}{\foreignlanguage{hindi}{अल्बेमार्ल}}} \textcolor{Red}{(\textbf{hallucination})}
\tcbline
\textbf{Question (\bn):} {\fontsize{8}{8}\selectfont \foreignlanguage{bengali}{আইরিয়ন কাউন্টি এর রাজধানী কি?}}
\\
\textbf{Ref. Ans. (\bn):} {\fontsize{8}{8}\selectfont \foreignlanguage{bengali}{মার্টজন}}
\\
\textbf{Model Ans.:} {\fontsize{8}{8}\selectfont \textcolor{Red}{\foreignlanguage{bengali}{হার্বার বিচ}}} \textcolor{Red}{(\textbf{hallucination})}
\tcbline
\textbf{Question (\ur):} \foreignlanguage{urdu}{ایریون کاؤنٹی کا دارالحکومت کیا ہے؟}
\\
\textbf{Ref. Ans. (\ur):} \foreignlanguage{urdu}{میرٹزون}
\\
\textbf{Model Ans.:} \textcolor{Red}{\foreignlanguage{urdu}{ایری، پنسلوانیا} (\textbf{hallucination})}
\end{tcolorbox}
\caption{Example from \texttt{mTREx-Capitals} along with the response from \llamaM across five languages. The model answers correctly in English (\en) and German (\de) but hallucinates in the low-resource languages.}
\label{fig:hallu-intro}
\end{figure}

\section{Related Work}

Performance disparities for under-represented groups or languages are a well-known problem in broader ML~\cite{barocas-hardt-narayanan} as well as for language models~\cite{liang2023holistic,atari2023humans,moayeri2024worldbench}. Causes for these disparities have been attributed to a variety of factors, including lack of training data, lack of sufficient coverage at a linguistic and cultural level~\cite{G-MMLU_paper,bender2021dangers}, and modeling choices like capacity~\cite{chen2018my} and tokenization~\cite{schwobel-etal-2023-geographical, zhou2022richer,neitemeier2025hierarchical}. Prior work has also found that LLMs can struggle in answering the same question in different languages~\cite{jiang-etal-2020-multilingual,islam2025llmshallucinatelanguagesmultilingual,wang-etal-2025-lost-multilinguality}, which is the focus of our paper.

Benchmarking model performance across languages can lead to a number of challenges, such as translation quality issues, lack of cultural and domain-specific context, and evaluation issues; see \cite{G-MMLU_paper, 10.1145/3748313, datta-etal-2023-mildsum} and references therein. In this work, we translate a well-known factual recall benchmark called TREx~\cite{trex_paper}. Inspection by native speakers reveals similar challenges; see~\secref{sec:datasets}.

With the emergence of generative models, hallucinations have come to the fore as an important issue. Hallucinations can manifest in a variety of language modeling tasks such as summarization, translation, and question-answering~\cite{lin2021truthfulqa,rawte2023survey,ji2023survey}.  In this work, we focus on the question answering (QA) setting. 

Plenty of recent work has looked into detecting and mitigating hallucinations. 
Most popular Hallucination Detection (HD) approaches operate either by repeatedly looking inside the model~\cite{snyder_early_2024,azaria-mitchell-2023-internal,ferrando2024know} or by inspecting the model output~\cite{manakul-etal-2023-selfcheckgpt,semantic_entropy_paper}. In our analysis, we consider HD techniques from both types and analyze their performance in low-resource scenarios. In the multilingual context, \citet{islam2025llmshallucinatelanguagesmultilingual}  introduced a multilingual dataset to study hallucination across languages, showing that models covering more languages tend to hallucinate more.
Recently, \citet{vazquez-etal-2025-semeval} and \citet{Abdaljalil2025HalluVerse25FM} also addressed multilingual hallucination, but their language overlap with our study is very limited, and their task is to detect spans of text corresponding to hallucinations.
To our knowledge, there is no prior work on comparing HD performance across high- and low-resource languages for the factual QA task.

%% file: 03_datasets.tex
\section{Multilingual QA Datasets}
\label{sec:datasets}

While QA datasets can span many domains, we focus on two popular categories: \textit{knowledge of the facts about the real world} and \textit{knowledge about academic disciplines}, such as STEM and humanities. We chose these two datasets: 
(1)~\textbf{TREx}~\citep{trex_paper} and (2)~\textbf{Global MMLU}~\citep{G-MMLU_paper}. 
TREx offers structured factual triples grounded in encyclopedic knowledge, enabling an evaluation of factual recall. Global MMLU provides questions spanning a wide range of disciplines. 
Notably, the \textit{response generation style} for these two datasets is also different, allowing us to study hallucination detection performance in different generation settings. TREx requires concise, fact-based short-form answers, whereas for Global MMLU, the model has to select one answer from four available options.

\subsection{Multilingual TREx (mTREx) dataset}
\label{subsec:mtrex-main}

The existing TREx dataset~\citep{trex_paper} is originally available only in English. 
\textbf{We developed a multilingual version of TREx}, which we refer to as \textbf{mTREx}, 
by translating parts of TREx into four additional languages: German (\de), Hindi (\hi), Bengali (\bn), and Urdu (\ur). 
Note that \bn and \ur are \textit{resource-poor languages}~\citep{haddow-etal-2022-survey}. 
Our objective in constructing mTREx is to benchmark hallucination detectors (HDs) across typologically and resource-wise diverse languages. 
To the best of our knowledge, there is no directly comparable, existing multilingual QA benchmark for hallucination detection, which motivated our creation of this mTREx.

\xhdr{Subject Categories} TREx comprises over 11 million factual triples sourced from Wikipedia and covers more than 600 unique Wikidata predicates. Each triple encodes a factual relationship between entities in the form of subject-predicate-object. For this study, we focus on three factual relations: 

\begin{tcolorbox}[colback=white, colframe=black, boxrule=1pt, width=0.49\textwidth, fontupper=\normalfont\small, boxsep=1pt, left=2pt, right=2pt, top=2pt, bottom=1pt, halign=justify] 
\noindent \textbf{(1) Capitals:} The capital of X is Y. 
\\ [0.3em]
\noindent \textbf{(2) Country:} The country X is located in is Y. 
\\ [0.3em]
\noindent \textbf{(3) Official Language:} The official language of X is Y. 
\end{tcolorbox}
Other predicates are excluded due to data imbalance, ambiguity, or poor translation fidelity, as discussed in Appendix~\ref{appendix:trex_filtering}. 
For each of the five languages, the number of samples present in Capitals, Country, and Official languages of mTREx, are 2500, 2500, and 2374, respectively.

\xhdr{Selection of Translation Model via Language-specific Considerations}
Both the \textit{subject} and \textit{object} entities in TREx are translated into the four languages stated above.
For this, we try a range of translation (MT) models, including open-source neural MT models (e.g., \textit{IndicTrans2}~\citep{gala2023indictrans2highqualityaccessiblemachine}), and proprietary multilingual LLMs (e.g., \textit{GPT-4}, \textit{GPT-4o-mini}). 
During translation, we encountered two major challenges: 
\begin{enumerate}[leftmargin=*, nolistsep]
    \item \textbf{\textit{Proper nouns with embedded adjectives often led to mistranslations}}. For example, `Future Shop', a company name, was mistranslated by some MT models into Hindi as `\foreignlanguage{hindi}{भविष्य की दुकान}', meaning `shop of the future'. Also, `Spectre' was translated by some models to `\foreignlanguage{hindi}{भूत-प्रेत}' (`ghosts') in Hindi, distorting the originally intended meaning. 
    \item \textbf{\textit{Models struggle with abbreviations}}. For example, \textit{IndicTrans2} rendered `Kingston Rd' as `\foreignlanguage{hindi}{किंग्स्टन आरडी}', failing to transliterate `Rd' (which is an abbreviation for `Road') correctly, whereas \textit{GPT-4o-mini} translated it correctly as `\foreignlanguage{hindi}{किंग्स्टन रोड}', preserving semantic accuracy.
\end{enumerate}
We selected \textit{GPT-4o-mini} as our translation model for its superior handling of named entities, and preservation of semantic intent.

\xhdr{Evaluation of translation quality} 
We conduct an extensive evaluation of the translations by the author team that included  \textit{native speakers} of HI, BN and UR, and an intermediate-level speaker for DE. 
The manual evaluation was over 50 randomly sampled QA pairs from each of the mTREx categories for all four target languages (600 samples in total) along three dimensions: semantic fidelity, named-entity correctness, and fluency / naturalness. 
As reported in Table~\ref{tab:trans-manual-eval}, our evaluation shows consistently high quality of mTREx across languages. On average, more than 90\% of translations were correct. Most of the errors observed were typographical or script variations in how entity names are written in particular languages, rather than semantic mistranslations or factual errors.

\begin{table}[t]
\centering
\small
\begin{tabular}{|c|c|c|c|}
\hline
\multirow{2}{*}{\bf Lang.} & \multicolumn{3}{c|} {\bf \% of correct translation} \\
\cline{2-4}
& {\bf Capitals} & {\bf Country} & {\bf Official Language} \\
\hline
\en-to-\de & 90 & 88 & 96 \\
\en-to-\hi & 88 & 90 & 80 \\
\en-to-\bn & 94 & 92 & 90 \\
\en-to-\ur & 96 & 86 & 92 \\
\hline
\avg & 92 & 89 & 90 \\
\hline
\end{tabular}
\caption{Assessment of translation quality in mTREx.}
\label{tab:trans-manual-eval}
\end{table}


\subsection{The Global-MMLU (G-MMLU) dataset}
The G-MMLU  dataset~\citep{G-MMLU_paper} is a multilingual extension of MMLU~\cite{hendrycks2021measuringmassivemultitasklanguage}. It offers a broad set of domain-diverse multiple-choice questions spanning a large number of disciplines.
In our study, we focus on two disciplines (\textit{STEM} and \textit{Humanities}) and four languages, namely, English, German, Hindi, and Bengali. Urdu is not covered in G-MMLU.
We uniformly sample 2,500 examples from each category while trying to retain a balance between subcategories. In the end, we obtained 2,510 samples from \textit{STEM} and 2,506 from \textit{Humanities}.


%% file: 04_HD_methods.tex
\section{Hallucination Detection Methods}
\label{sec:hallu-methods-main}

We focus on three hallucination detection (HD) methods that represent two popular families of HD methods: 
(i)~\textit{methods that leverage model internal artifacts during generation}, and 
(ii)~\textit{sampling-based blackbox detectors utilizing only model responses}. 
We briefly describe the HD methods below; detailed descriptions are in Appendix~\ref{sec:hallu-appn}. 

\subsection{Model Artifacts Method (\mam)}

These techniques utilize a model's internal artifacts to identify signals of hallucination~\cite{snyder_early_2024,azaria-mitchell-2023-internal}.
We consider two types of artifacts: \textit{self-attention scores} and \textit{fully-connected activations}.
Following \cite{snyder_early_2024}, we consider the hidden states from the final layer at the first generated token, as they showed no gains from alternative layers or token positions. We also perform ablation studies with averaging models' artifacts across multiple generated tokens (up to the first 10 generated tokens) from the final layer. The performances remain largely similar in both setups, as discussed in Appendix~\ref{subsec:appn-hallu-methods}.

\xhdr{Classifier for hallucination detection} After extracting the hidden states, we train a single-layer neural network with a hidden dimension of 256 to classify whether a response was factually correct or hallucinated. 
For each dataset category, we train and evaluate the classifier on a random 80/20 split in three settings: (i) \textit{same language setup} (train and test over the same language data), (ii) \textit{\en-to-target cross-lingual setup} (train on \en and test over target language data), and (iii) \textit{multilingual setup} (train on combined multilingual data including \en and test over each target langauge data).

\subsection{SelfCheckGPT Method (\sgm)}

This sampling-based  method~\citep{manakul-etal-2023-selfcheckgpt} utilizes the LLM itself to determine the factuality of a generated response by measuring the factual consistency across multiple responses sampled for the same question. 
The intuition is that factual responses tend to remain consistent across multiple generations, whereas hallucinated responses are likely to contradict one another. 

For a given question, a set of  responses is generated with high temperature, and a `base response' is generated by setting the temperature to zero to obtain the deterministic output from the model.
Next, the method computes the average negative log-likelihood (NLL) of the base response,
to evaluate how much the base response is varied with respect to the distribution defined by the sampled responses. 
The NLL is then transformed into the \textit{final SelfCheck n-gram score}  $1.0 - \exp(-avg_{NLL})$, ranging between 0 and 1. Higher scores correspond to lower NLL and thus higher consistency.
For each question, the computed SelfCheck n-gram score serves as the detector's output, predicting the likelihood of the base response being factual or hallucinated. 

\subsection{Semantic Entropy Method (\sem)}


\sem is also a sampling-based method~\citep{semantic_entropy_paper} 
where the key idea is that when an LLM is uncertain, it is more likely to generate responses that diverge significantly in their semantic content, leading to higher entropy. 
Conversely, when a model is confident, it generates semantically consistent responses that are expected to cluster around a single or a few closely related meanings, resulting in lower \textit{semantic entropy}.

For a given question, a set of diverse responses is generated with high temperature, and a `base response' is generated with zero temperature to serve as the reference for determining correctness.
To assess the semantic equivalence of sampled responses, we utilize the language-agnostic \textit{LaBSE} model~\citep{labse_paper} to encode each response into a fixed-dimensional semantic embedding. 
The sampled responses are grouped into semantic clusters if the mutual cosine similarity exceeds a threshold ($\tau = 0.75$, chosen based on the internal evaluation by the native speakers). 
The probability of each cluster $C_k$ is computed by summing the probabilities of the individual responses belonging to the cluster. If $C_k$ contains $I_k$ responses, then the aggregated probability is given by:
$P(C_k) = \sum_{i \in I_k} P(\text{response}_i)$.
The \textit{semantic entropy} is then computed over the distribution of $P(C_k)$.
Finally, the entropy values are compared against binary ground-truth labels. 

\subsection{Defining Hallucinations}
\label{subsec:define-hallu}

We consider a response to be a hallucination if it does not match the reference answer~\citep{snyder_early_2024, ji2023survey}. 
However, the nature of the expected model responses and evaluation criteria differ across the datasets. 

\xhdr{mTREx} The questions from mTREx require short-form factual answers. But, due to the potential verbosity and variation in LLM responses, exact string match is \textit{not} a reliable metric for correctness \cite{adlakha-etal-2024-evaluating}. For instance, for the question \textit{`What is the official language of Italy?'}, the reference answer is \textit{`Italian'}. However, responses such as \textit{`Italian.'} or \textit{`Italian is the official language of Italy.'} are both correct. 
Hence, we adopt the same heuristic as suggested in~\cite{snyder_early_2024,liang2023holistic,adlakha-etal-2024-evaluating}, where a generated response $A$ is marked correct if the reference answer $R$ is a substring of $A$. We perform all comparisons in lowercase.

We manually evaluated this heuristic over a set of 50 model responses in all five languages (750 samples in total) and observed close alignment with human judgment -- \textbf{on average, this heuristic was found to be correct in more than 95\% cases for \en and more than 88\% cases for non-\en languages}. Details of this evaluation are in Appendix~\ref{sec:hallu-heuristic-eval}.

\xhdr{G-MMLU} This dataset consists of multiple-choice questions where the model has to select one correct option from four. So, if the model's prediction does not match the correct option, the response is labeled as hallucinated.

%% file: 05_experiments.tex
\section{Experimental Setup} 
\label{sec:exp-setup}

We now describe our models, prompts and evaluation metrics.

\subsection{Models}
\label{subsec:llms}

We considered four popular instruction-tuned LLMs, representing a range of sizes, to analyse the effect of model size on hallucinations across both low and high-resource languages --
Mistral-7B-Instruct,
LLaMA-8B-Instruct,
Mistral-24B-Instruct 
and LLaMA-70B-Instruct. 
More details about the LLM versions, hyperparameters and infrastructure are given in Appendix~\ref{appendix:expt-settings}.

\subsection{Prompts}

We observed that LLMs often generate responses in English, even when the question is in a different language. To mitigate this issue we designed dedicated, language-specific prompt templates 
guiding the models to answer in the target language only. 

\xhdr{Prompts used for mTREx} To generate responses from LLMs for mTREx, we adopted a structured prompt template consisting of a language-specific general instruction (\textit{system prompt}) followed by a question template (\textit{user prompt}) with a designated answer placeholder (\textit{assistant prompt}), explicitly asking for a short answer in the same language.
For the \textit{Country} relation, the prompt in English is shown in Figure~\ref{fig:mtrex-prompt}.
Details of prompts for other languages are in Appendix~\ref{appendix:prompts_trex}.

\begin{figure}[t]
\centering
\begin{tcolorbox}[colback=gray!15, colframe=black, boxrule=1pt, width=0.49\textwidth, fontupper=\normalfont\small, boxsep=1pt, left=2pt, right=2pt, top=1pt, bottom=1pt, halign=justify] 
\noindent \textbf{System Prompt:} Suppose your job is to answer given questions in English. When you are asked a question in English, please give a brief answer in English only. Do not write anything else except the answer. 
\\ [0.3em]
\textbf{User Prompt:} Question: In which country is the \texttt{<X>} located? 
\\ [0.3em]
\textbf{Assistant prompt:} Answer (in English):
\end{tcolorbox}
\vspace{-1em}

\caption{Example prompt for mTREx (Country relationship) in English.}
\label{fig:mtrex-prompt}
\end{figure}


\xhdr{Prompts used for G-MMLU}
For the experiments with G-MMLU, we designed the prompt that incorporates few-shot learning with the two smallest language-specific examples from the same category.
Here, the prompt follows a structured format consisting of an instruction (\textit{system prompt}), a small set of two QA pairs, and a test question (\textit{user prompt}). The model is expected to output the correct option (in A/B/C/D format) followed by a brief explanation. The English prompt is provided in Figures~\ref{fig:eng-gmmlu-prompt}. 
Full prompt templates for each language are detailed in Appendix~\ref{appendix:prompts_gmmlu}. 

\begin{figure}[t]
\centering
\begin{tcolorbox}[colback=gray!15, colframe=black, boxrule=1pt, width=0.49\textwidth, fontupper=\normalfont\small, boxsep=1pt, left=2pt, right=2pt, top=1pt, bottom=1pt, halign=justify] 
\noindent \textbf{System Prompt:} You are a helpful assistant trained to answer objective questions from <subject-category>. Each question comes with 4 options (A, B, C, and D). Provide your answer in the format of a single letter (A, B, C, or D) followed by an explanation in 20 words. Use the given examples to guide your answers. The examples do not have an explanation, but your response should have. 
\\ [0.3em]
\textbf{User Prompt:} Q1: <example-question-1> 
\\ [0.2em]
\textbf{Assistant Prompt:} Answer: <answer-example-question-1> 
\\ [0.3em]
\textbf{User Prompt:} Q2: <example-question-2> 
\\ [0.2em]
\textbf{Assistant Prompt:} Answer: <answer-example-question-2> 
\\ [0.3em]
\textbf{User Prompt:} Question: <actual-test-question-with-multiple-choices> 
\\ [0.3em]
\textbf{Assistant Prompt:} Answer:
\end{tcolorbox}
\caption{Prompt for G-MMLU in English.}
\label{fig:eng-gmmlu-prompt}
\end{figure}


\subsection{Evaluation Metrics}
\label{subsec:eval-metric}

\xhdr{Task performance} We report percentage accuracy for the LLMs, where correct / hallucination is decided as described in \secref{subsec:define-hallu}. 

\xhdr{Hallucination detection performance}
To evaluate hallucination detectors, we choose the \textit{Area Under the Receiver Operating Characteristic} (\textbf{AUROC}) curve due to its robustness against class imbalance and its threshold-independent nature. 
Since the ratio of correct vs. hallucinated responses is often highly imbalanced (Table~\ref{table:acc}), usage of binary classification accuracy could be misleading.
AUROC ranges from 0 to 1, where 1 indicates perfect classification, 0.5 corresponds to random guessing, and values below 0.5 suggest worse-than-random performance.
For easier readability, we reported AUROC scores multiplied by 100.

\xhdr{Task Performance to Hallucination Ratio} To investigate the alignment between disparities in task performance and Hallucination Detector's (HD) performance across languages, we introduce a novel metric: \textit{Task Performance to Hallucination Ratio (\metric)}. 
It quantifies how the difference in the model's task accuracy for a given language $L$ compares to the difference in its HD's performance for the same language relative to the corresponding English (\en) baseline, thus serving as a \textit{useful metric for examining the low-resource effect in multilingual hallucination detection}. 
Formally, for a specific LLM and a specific HD method, the \metric for a target language $L$ is defined as:

{\small
\begin{equation*}
\text{\metric}(L) = 
\log_{10}\!\left(
\frac{|\text{Accuracy}(\text{EN}) - \text{Accuracy}(L)|}
{|\text{AUROC}_{\text{HD}}(\text{EN}) - \text{AUROC}_{\text{HD}}(L)|}
\right)
\end{equation*}
}

\noindent \metric provides interpretable signals:
\begin{itemize}[leftmargin=*, nolistsep]
\item \metric $\approx 0$: The disparity in HD's performance aligns with the disparity in task accuracy, i.e., both task performance and HD performance change equally w.r.t. the English baseline.
\item \metric $= +k$ ($k \in \mathbb{R}^{+}$): The difference in task accuracy is $10^{k}$ times larger than the difference in HD's performance. 
In particular, \metric $> 1$ implies the degradation in LLM's task accuracy is more than $10$ times the degradation in HD's performance.
\item \metric $= -k$ ($k \in \mathbb{R}^{+}$): Difference in HD's performance is $10^{k}$ times larger than the difference in LLM's task accuracy. 
\end{itemize}
In edge cases, \metric is assigned special values, namely UN and NA. UN arises when the task accuracy delta is zero, making the ratio 0, and thus rendering the logarithm undefined; in such cases, no meaningful conclusion can be drawn regarding the HD's behavior. In contrast, NA occurs when HD's AUROC delta is zero for a language compared to the \en baseline, which directly indicates strong detector robustness, i.e., HD's performance for the other language is equivalent to that in \en.


\input{final_tables/task_acc_main}

\input{final_tables/mam_fc_main}

\section{Results and Observations}
\label{sec:results-main}

We now present our observations on task performance and hallucination detection performance.

\input{final_tables/tphr_log10_table}

\subsection{Task Performance}
\label{sec:task-acc}

Table~\ref{table:acc_delta_percentage} presents the percentage of increment or decrement of task accuracy in answering factual questions for \mistralS, \llamaS, \mistralM, and \llamaM relative to their respective \en baselines. The exact accuracy values are shown in Table~\ref{table:acc} of Appendix~\ref{sec:appn-results}. 
\textbf{We observe a substantial drop in performance for low-resource languages such as \bn and \ur}. On average, accuracies decrease by more than 62\% for \bn, 53\% for \ur, and 49\% for \hi relative to the \en baseline, in the mTREx categories. For \de, the performance degradations are comparatively much lower, mostly around 10\%--20\%.
The performance drop is consistently more pronounced in mTREx than in G-MMLU. 
\textit{These observations highlight a severe performance disparity in LLMs when evaluated beyond high-resource languages and motivate the need for multilingual hallucination mitigation strategies}.


\begin{figure}[t]
\centering
    \includegraphics[width=0.9\linewidth, height=5cm]{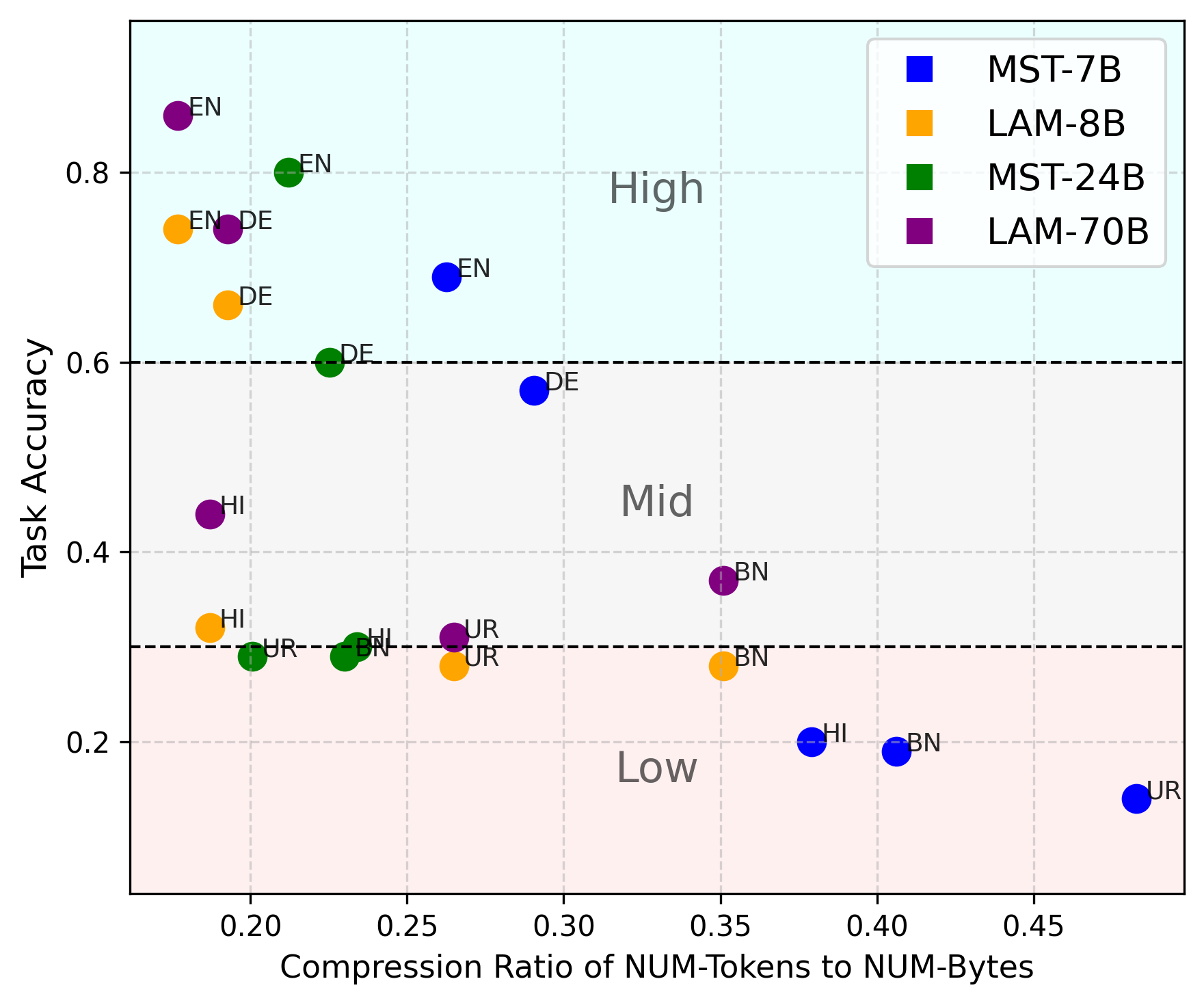}
    \vspace{-0.5em}
    \caption{Task accuracy vs. token compression ratio for the Capitals dataset. Low-resource languages have higher compression ratios, showing inefficient tokenization, along with lower task performance.}
\label{fig:token-ratio-capitals}
\vspace{-5mm}
\end{figure}

\xhdr{Tokenization capabilities across languages}
Inspired by \citet{zhou2022richer,schwobel-etal-2023-geographical,neitemeier2025hierarchical}, we investigate how tokenization efficiency relates to task performance. Specifically, we analyze the \textit{token compression ratio}, defined as the ratio of the number of tokens produced by the model's tokenizer to the number of bytes in the input string. Figure~\ref{fig:token-ratio-capitals} illustrates the relation between this ratio and the task accuracy over the Capitals dataset, across all LLMs and languages. Similar figures for other categories are presented in Appendix~\ref{subsec:appn-token-plots}. We find \textbf{an inverse relation between token compression ratio and model accuracy}. In particular, \textit{low-resource languages such as \bn and \ur, show high compression ratios -- indicating inefficient tokenization -- while simultaneously yielding very low task accuracy}. In contrast, for resource-rich languages like \en and \de, models show more optimized tokenization and higher performance. This analysis supports the finding that \textit{current LLM tokenizers are poorly adapted to non-English scripts, and tokenization inefficiency may be a key factor to reduced performance in multilingual scenarios.}

\subsection{Hallucination Detectors Performance}

Next, we analyze the performances of HDs across languages by comparing the AUROC scores in non-\en languages against their respective \en baselines. Tables~\ref{tab:snyder_fullyconnected_delta_percentage} present the percentage of increment or decrement in AUROC for \mam~{\texttt{(fully connected activations)}. 
Results for other HD methods: \mam~\texttt{(self-attention)}, \sem, and \sgm, along with ablation and statistical studies for examining the robustness of HDs are reported in detail in Appendix~\ref{subsec:appn-hallu-methods}.
We observe a consistent trend across datasets: \textbf{although the task accuracy for resource-poor languages (\bn and \ur) drops significantly compared to \en} (discussed in \secref{sec:task-acc}), \textbf{the degradation in HD's performance is often much smaller}. For example, in Capitals and Country, the drop in AUROC for \bn and \ur is limited to around 5--10\% for most LLMs, whereas the degradation in task performance is as high as 45--90\% for these low-resource languages. In some cases, the HD performance even increases.
For instance, for \mistralM and \llamaM, around 5\% gains are observed in HD in \ur, \bn, \hi on the Official Language dataset. 
Overall, the findings highlight that HDs remain relatively robust across languages, especially compared to the significant performance drop in task accuracy. 

\xhdr{Insights from \metric values} 
Table~\ref{table:ratio_acc_auc_tphr_log10} 
shows that \metric values are notably high ($>1$) in low-resource languages like \bn and \ur, particularly for the mTREx datasets. 
For instance, in \mam (fully connected activations), \mistralM shows a high \metric value of 1.71 in \bn for Capitals, while for Country, \mistralS shows \metric = 1.53 in \ur. 
Note that \metric $>1$ indicates the degradation in the model's task accuracy is more than $10$ times the degradation in HD's performance w.r.t. \en baseline, i.e., 
\textbf{although the task accuracy for these languages drops drastically, the HD's performance remains much more stable}. 
In comparison, \de tend to exhibit lower \metric values for most cases, implying a more balanced degradation in both task and HD performance. 
Thus, the \metric metric provides a quantitative lens to interpret these disparities and \textit{validates the stability of HDs in multilingual settings, particularly for low-resource languages where task accuracy suffers the most}.

\input{final_tables/mam_attn_cross_table_main}

\xhdr{Entropy analyses between generators and detectors} To further investigate why detectors' performance remains stable across languages even when generators' performance (i.e., LLMs' task accuracy) degrades, we analyze the \textit{entropy of softmax distributions} for both generators and detectors. Low-resource languages are typically tokenized into a larger number of tokens, which are also seen far less frequently during pretraining. Consequently, at each generation step, the model must choose from a larger and less familiar token set, yielding a more spread-out softmax distribution in low-resource languages. Whereas high-resource languages exhibit more concentrated softmax distributions. On the other hand, detectors make a binary decision (hallu vs. non-hallu) with the same training setup across languages, leading to more comparable softmax distributions irrespective of language. Building on this intuition, our analysis also reveals that \textit{generators exhibit much higher entropy in low-resource settings} (e.g., \mistralM on \texttt{Capitals}: 0.325 in \en vs. 1.762 in \ur), \textit{while detector entropies remain largely stable across languages} (e.g., \mam \texttt{(fully connected activations)} on \texttt{Country} with \mistralM: 0.3 in \en vs. 0.336 in \ur). \textit{This contrast suggests that low-resource effects primarily impact the high-dimensional generation space, whereas detection operates over a simpler, language-agnostic binary decision space (hallu vs.\ non-hallu), leading to more stable detectors' performance.} Detailed analyses and results are provided in Appendix~\ref{subsec:appn-entropy-analyses}.

\xhdr{Comparison among HD methods} Another key finding in our study is that both \mam methods (fully connected activations and self-attention)  outperformed sampling-based blackbox methods \sem and \sgm across all languages, demonstrating that LLMs' internal signals -- captured through artifacts -- remain informative for detecting hallucination even in a multilingual scenario.

\xhdr{Cross-lingual and multilingual analyses for \mam detectors} So far, we evaluated the \mam classifiers in the same-language settings. We now conduct cross-lingual and multilingual experiments to observe language transfer effects.

\noindent \underline{\textit{Cross-lingual setting:}}
In this setup, classifiers are trained on 80\% of \en data and evaluated on 20\% of each non-\en language separately, across all datasets and LLMs. We \textbf{observe substantial performance drops when transferring across languages}. For instance, on the Capitals dataset for \mam~\texttt{(self-attention)}, average AUROC across all LLMs decreases by more than 19\% in \de, 29\% in \hi, 26\% in \bn, and 34\% in \ur (see Table~\ref{table:snyder_attention} \& Table~\ref{table:mam-self-attn-cross-lingual}).

\noindent \underline{\textit{Multilingual setting:}} 
Here, classifiers are trained on 80\% of the combined multilingual data (all languages including \en) and evaluated on 20\% of each language individually. In contrast to the cross-lingual case, \textbf{performance in this multilingual setting remains close to the same-language baseline}. For instance, for \mam~\texttt{(self-attention)} over Capitals, average AUROC across all LLMs is nearly unchanged: 74 vs. 72 for \de, 70 vs. 68 for \hi, 71 vs. 71 for \bn, and 70 vs. 67 for \ur (see Table~\ref{table:snyder_attention} \& Table~\ref{table:mam-self-attn-multi-lingual}).


\noindent Table~\ref{table:mam-self-attn-cross-lingual_delta_percentage_main} \& Table~\ref{table:mam-self-attn-multi-lingual_delta_percentage_main} present the percentage of increment or decrement in AUROC w.r.t. the \en baseline for \mam (\texttt{self-attention}) for cross-lingual and multilingual setups, respectively.
Detailed cross-lingual and multilingual results are discussed in Appendix~\ref{subsec:appn-cross-multi-mam}. \textit{These experiments demonstrate that zero-shot cross-lingual transfer is challenging, but multilingual training with in-language supervision mitigates the performance gap.}

}

\input{tables/only_percentage_difference}
\input{tables/accuracy_final}
\input{tables/snyder_attention_final}
\input{tables/snyder_fully_connected_final}
\input{tables/selfcheckgpt_final}
\input{tables/semanticentropy_final}
\input{tables/only_acc_auroc_values}

%% file: final_tables/task_acc_main.tex
\begin{table*}[!htb]
\centering
\small
\resizebox{\textwidth}{!}{
\begin{tabular}{l|ccccc|ccccc|ccccc|cccc|cccc|}
\toprule
\multirow{2}{*}{\bf Models} & \multicolumn{5}{|c|}{\bf mTREx -- Capitals} & \multicolumn{5}{|c|}{\bf mTREx -- Country} & \multicolumn{5}{|c|}{\bf mTREx -- Official Language} & \multicolumn{4}{|c|}{\bf G-MMLU -- STEM} & \multicolumn{4}{|c|}{\bf G-MMLU -- Humanities} \\
 & \en & \de & \hi & \bn & \ur & \en & \de & \hi & \bn & \ur & \en & \de & \hi & \bn & \ur & \en & \de & \hi & \bn & \en & \de & \hi & \bn \\
\midrule
\mistralS & 69 & \colorbox{red!20}{\scriptsize ↓17} & \colorbox{red!50}{\scriptsize ↓71} & \colorbox{red!50}{\scriptsize ↓72} & \colorbox{red!50}{\scriptsize ↓80} & 73 & \colorbox{red!30}{\scriptsize ↓32} & \colorbox{red!50}{\scriptsize ↓90} & \colorbox{red!50}{\scriptsize ↓93} & \colorbox{red!50}{\scriptsize ↓93} & 82 & \colorbox{red!20}{\scriptsize ↓13} & \colorbox{red!50}{\scriptsize ↓85} & \colorbox{red!50}{\scriptsize ↓87} & \colorbox{red!50}{\scriptsize ↓83} & 43 & \colorbox{red!20}{\scriptsize ↓16} & \colorbox{red!40}{\scriptsize ↓44} & \colorbox{red!40}{\scriptsize ↓47} & 53 & \colorbox{red!20}{\scriptsize ↓19} & \colorbox{red!40}{\scriptsize ↓45} & \colorbox{red!40}{\scriptsize ↓51} \\
\llamaS & 74 & \colorbox{red!20}{\scriptsize ↓11} & \colorbox{red!40}{\scriptsize ↓57} & \colorbox{red!50}{\scriptsize ↓62} & \colorbox{red!50}{\scriptsize ↓62} 
& 77 & \colorbox{red!30}{\scriptsize ↓36} & \colorbox{red!40}{\scriptsize ↓52} & \colorbox{red!50}{\scriptsize ↓60} & \colorbox{red!40}{\scriptsize ↓58} 
& 70 & {\scriptsize ↓0} & \colorbox{red!30}{\scriptsize ↓37} & \colorbox{red!40}{\scriptsize ↓51} & \colorbox{red!40}{\scriptsize ↓43} & 56 & \colorbox{red!20}{\scriptsize ↓12} & \colorbox{red!30}{\scriptsize ↓25} & \colorbox{red!30}{\scriptsize ↓38} & 61 & \colorbox{red!20}{\scriptsize ↓16} & \colorbox{red!30}{\scriptsize ↓28} & \colorbox{red!30}{\scriptsize ↓39} \\
\mistralM & 80 & \colorbox{red!30}{\scriptsize ↓25} & \colorbox{red!50}{\scriptsize ↓62} & \colorbox{red!50}{\scriptsize ↓64} & \colorbox{red!50}{\scriptsize ↓64} & 77 & \colorbox{red!30}{\scriptsize ↓36} & \colorbox{red!40}{\scriptsize ↓51} & \colorbox{red!50}{\scriptsize ↓60} & \colorbox{red!50}{\scriptsize ↓62} & 85 & \colorbox{red!20}{\scriptsize ↓15} & \colorbox{red!40}{\scriptsize ↓46} & \colorbox{red!40}{\scriptsize ↓55} & \colorbox{red!40}{\scriptsize ↓48} & 70 & \colorbox{red!10}{\scriptsize ↓09} & \colorbox{red!30}{\scriptsize ↓30} & \colorbox{red!30}{\scriptsize ↓36} & 72 & \colorbox{red!20}{\scriptsize ↓17} & \colorbox{red!30}{\scriptsize ↓38} & \colorbox{red!40}{\scriptsize ↓44} \\
\llamaM & 86 & \colorbox{red!20}{\scriptsize ↓14} & \colorbox{red!40}{\scriptsize ↓49} & \colorbox{red!40}{\scriptsize ↓57} & \colorbox{red!50}{\scriptsize ↓64} & 84 & \colorbox{red!30}{\scriptsize ↓32} & \colorbox{red!30}{\scriptsize ↓33} & \colorbox{red!40}{\scriptsize ↓46} & \colorbox{red!40}{\scriptsize ↓49} & 87 & \colorbox{red!10}{\scriptsize ↓07} & \colorbox{red!30}{\scriptsize ↓29} & \colorbox{red!40}{\scriptsize ↓54} & \colorbox{red!30}{\scriptsize ↓39} & 74 & \colorbox{red!10}{\scriptsize ↓07} & \colorbox{red!30}{\scriptsize ↓27} & \colorbox{red!20}{\scriptsize ↓14} & 80 & \colorbox{red!20}{\scriptsize ↓12} & \colorbox{red!20}{\scriptsize ↓18} & \colorbox{red!30}{\scriptsize ↓21} \\
\midrule
\avg & 77 & \colorbox{red!20}{\scriptsize ↓17} & \colorbox{red!40}{\scriptsize ↓58} & \colorbox{red!50}{\scriptsize ↓64} & \colorbox{red!50}{\scriptsize ↓66} & 78 & \colorbox{red!30}{\scriptsize ↓35} & \colorbox{red!40}{\scriptsize ↓56} & \colorbox{red!50}{\scriptsize ↓64} & \colorbox{red!50}{\scriptsize ↓65} & 81 & \colorbox{red!10}{\scriptsize ↓09} & \colorbox{red!40}{\scriptsize ↓49} & \colorbox{red!50}{\scriptsize ↓62} & \colorbox{red!40}{\scriptsize ↓53} & 61 & \colorbox{red!20}{\scriptsize ↓11} & \colorbox{red!30}{\scriptsize ↓31} & \colorbox{red!30}{\scriptsize ↓31} & 66 & \colorbox{red!20}{\scriptsize ↓15} & \colorbox{red!30}{\scriptsize ↓30} & \colorbox{red!30}{\scriptsize ↓36} \\
\bottomrule
\end{tabular}
}
\caption{The left column for each dataset shows the task accuracy for English (\en). The following columns show $\%$ increase \colorbox{green!17}{\scriptsize ↑} or decrease \colorbox{red!17}{\scriptsize ↓} in task accuracy for the corresponding language w.r.t. \en.
Darker color shades indicate larger  differences.
The task performance decreases for all languages when compared with \en.
}
\label{table:acc_delta_percentage}
\end{table*}

%% file: final_tables/mam_fc_main.tex
\begin{table*}[!htb]
\centering
\small
\resizebox{\textwidth}{!}{
\begin{tabular}{l|ccccc|ccccc|ccccc|cccc|cccc|}
\toprule
\multirow{2}{*}{\bf Models} & \multicolumn{5}{|c|}{\bf mTREx -- Capitals} & \multicolumn{5}{|c|}{\bf mTREx -- Country} & \multicolumn{5}{|c|}{\bf mTREx -- Official Language} & \multicolumn{4}{|c|}{\bf G-MMLU -- STEM} & \multicolumn{4}{|c|}{\bf G-MMLU -- Humanities} \\
 & \en & \de & \hi & \bn & \ur & \en & \de & \hi & \bn & \ur & \en & \de & \hi & \bn & \ur & \en & \de & \hi & \bn & \en & \de & \hi & \bn \\
\midrule
\mistralS & 82 & \colorbox{red!20}{\scriptsize ↓12} & \colorbox{red!10}{\scriptsize ↓09} & \colorbox{red!20}{\scriptsize ↓10} & \colorbox{red!20}{\scriptsize ↓13} & 79 & \colorbox{green!10}{\scriptsize ↑08} & \colorbox{red!20}{\scriptsize ↓16} & \colorbox{green!10}{\scriptsize ↑05} & \colorbox{red!10}{\scriptsize ↓03} & 81 & \colorbox{green!10}{\scriptsize ↑04} & \colorbox{red!10}{\scriptsize ↓07} & \colorbox{red!10}{\scriptsize ↓05} & \colorbox{green!10}{\scriptsize ↑06} & 76 & \colorbox{red!20}{\scriptsize ↓12} & \colorbox{red!30}{\scriptsize ↓25} & \colorbox{red!30}{\scriptsize ↓32} & 73 & \colorbox{red!10}{\scriptsize ↓08} & \colorbox{red!30}{\scriptsize ↓22} & \colorbox{red!30}{\scriptsize ↓30} \\
\llamaS & 79 & \colorbox{red!10}{\scriptsize ↓06} & \colorbox{red!10}{\scriptsize ↓09} & \colorbox{red!10}{\scriptsize ↓06} & \colorbox{red!20}{\scriptsize ↓11} & 83 & \colorbox{green!10}{\scriptsize ↑05} & \colorbox{red!10}{\scriptsize ↓07} & \colorbox{green!10}{\scriptsize ↑01} & \colorbox{red!10}{\scriptsize ↓07} & 88 & \colorbox{red!10}{\scriptsize ↓03} & \colorbox{red!10}{\scriptsize ↓05} & \colorbox{red!10}{\scriptsize ↓08} & \colorbox{red!10}{\scriptsize ↓05} & 78 & \colorbox{red!10}{\scriptsize ↓05} & \colorbox{red!20}{\scriptsize ↓15} & \colorbox{red!20}{\scriptsize ↓15} & 78 & \colorbox{red!10}{\scriptsize ↓05} & \colorbox{red!20}{\scriptsize ↓19} & \colorbox{red!20}{\scriptsize ↓19} \\
\mistralM & 76 & \colorbox{green!10}{\scriptsize ↑01} & \colorbox{red!10}{\scriptsize ↓05} & \colorbox{green!10}{\scriptsize ↑01} & \colorbox{red!10}{\scriptsize ↓08} & 86 & \colorbox{green!10}{\scriptsize ↑09} & \colorbox{red!10}{\scriptsize ↓05} & \colorbox{red!10}{\scriptsize ↓01} & \colorbox{red!10}{\scriptsize ↓05} & 82 & \colorbox{green!10}{\scriptsize ↑09} & \colorbox{green!10}{\scriptsize ↑05} & \colorbox{green!10}{\scriptsize ↑05} & \colorbox{green!10}{\scriptsize ↑05} & 82 & \colorbox{red!10}{\scriptsize ↓04} & \colorbox{red!20}{\scriptsize ↓11} & \colorbox{red!20}{\scriptsize ↓16} & 78 & \colorbox{green!10}{\scriptsize ↑04} & \colorbox{red!20}{\scriptsize ↓17} & \colorbox{red!20}{\scriptsize ↓18} \\
\llamaM & 80 & \colorbox{red!10}{\scriptsize ↓04} & \colorbox{red!20}{\scriptsize ↓19} & \colorbox{red!20}{\scriptsize ↓14} & \colorbox{red!20}{\scriptsize ↓11} & 87 & \colorbox{green!10}{\scriptsize ↑08} & \colorbox{red!20}{\scriptsize ↓10} & \colorbox{red!20}{\scriptsize ↓10} & \colorbox{red!10}{\scriptsize ↓08} & 82 & \colorbox{green!10}{\scriptsize ↑07} & \colorbox{green!10}{\scriptsize ↑07} & {\scriptsize ↓0} & \colorbox{green!10}{\scriptsize ↑05} & 84 & \colorbox{red!10}{\scriptsize ↓05} & \colorbox{red!10}{\scriptsize ↓01} & \colorbox{red!20}{\scriptsize ↓11} & 86 & \colorbox{red!10}{\scriptsize ↓08} & \colorbox{red!10}{\scriptsize ↓06} & \colorbox{red!20}{\scriptsize ↓16} \\
\midrule
\avg & 79 & \colorbox{red!10}{\scriptsize ↓05} & \colorbox{red!20}{\scriptsize ↓10} & \colorbox{red!10}{\scriptsize ↓06} & \colorbox{red!20}{\scriptsize ↓11} & 84 & \colorbox{green!10}{\scriptsize ↑07} & \colorbox{red!20}{\scriptsize ↓10} & \colorbox{red!10}{\scriptsize ↓02} & \colorbox{red!10}{\scriptsize ↓06} 
& 83 & \colorbox{green!10}{\scriptsize ↑04} & {\scriptsize ↓0} & \colorbox{red!10}{\scriptsize ↓01} & \colorbox{green!10}{\scriptsize ↑04} & 80 & \colorbox{red!10}{\scriptsize ↓06} & \colorbox{red!20}{\scriptsize ↓12} & \colorbox{red!20}{\scriptsize ↓18} & 79 & \colorbox{red!10}{\scriptsize ↓05} & \colorbox{red!20}{\scriptsize ↓16} & \colorbox{red!30}{\scriptsize ↓22} \\
\bottomrule
\end{tabular}
}
\caption{Increase \colorbox{green!17}{\scriptsize ↑} or decrease \colorbox{red!17}{\scriptsize ↓} in hallucination detector AUROC scores (shown in percentages) w.r.t. the \en baseline. The table show the results for the \textbf{\mam \texttt{(fully connected activations)}} method. Results of other hallucination detection methods are reported in Appendix~\ref{subsec:appn-hallu-methods}. The HD performance often drops when compared with \en but the drops are usually smaller than corresponding drops in the task accuracy (Table~\ref{table:acc}).}
\label{tab:snyder_fullyconnected_delta_percentage}
\vspace{-5mm}
\end{table*}

%% file: final_tables/tphr_log10_table.tex
\begin{table*}[t]
\centering
\small
\resizebox{\textwidth}{!}{
\begin{tabular}{|l|cccc|cccc|cccc|ccc|ccc|}
\toprule
\multirow{2}{*}{\bf Models} & \multicolumn{4}{|c|}{\bf mTREx -- Capitals} & \multicolumn{4}{|c|}{\bf mTREx -- Country} & \multicolumn{4}{|c|}{\bf mTREx -- Official Language} & \multicolumn{3}{|c|}{\bf G-MMLU -- STEM} & \multicolumn{3}{|c|}{\bf G-MMLU -- Humanities} \\
 &  \de & \hi & \bn & \ur &  \de & \hi & \bn & \ur &  \de & \hi & \bn & \ur &  \de & \hi & \bn &  \de & \hi & \bn \\
\midrule
\multicolumn{19}{|c|}{\textbf{\mam (fully connected activations)}} \\
\hline
\mistralS & \colorbox{MidnightBlue!10}{0.079} & \colorbox{MidnightBlue!10}{0.845} & \colorbox{MidnightBlue!10}{0.796} & \colorbox{MidnightBlue!10}{0.699} & \colorbox{MidnightBlue!10}{0.584} & \colorbox{MidnightBlue!10}{0.706} & \colorbox{MidnightBlue!30}{1.230} & \colorbox{MidnightBlue!30}{1.531} & \colorbox{MidnightBlue!10}{0.564} & \colorbox{MidnightBlue!30}{1.067} & \colorbox{MidnightBlue!30}{1.249} & \colorbox{MidnightBlue!30}{1.134} & \colorbox{MidnightBlue!10}{-0.109} & \colorbox{MidnightBlue!10}{0.000} & \colorbox{MidnightBlue!10}{-0.079} & \colorbox{MidnightBlue!10}{0.222} & \colorbox{MidnightBlue!10}{0.176} & \colorbox{MidnightBlue!10}{0.089} \\
\llamaS & \colorbox{MidnightBlue!10}{0.204} & \colorbox{MidnightBlue!10}{0.778} & \colorbox{MidnightBlue!10}{0.964} & \colorbox{MidnightBlue!10}{0.709} & \colorbox{MidnightBlue!10}{0.845} & \colorbox{MidnightBlue!10}{0.824} & \colorbox{MidnightBlue!30}{1.663} & \colorbox{MidnightBlue!10}{0.875} & UN & \colorbox{MidnightBlue!10}{0.813} & \colorbox{MidnightBlue!10}{0.711} & \colorbox{MidnightBlue!10}{0.875} & \colorbox{MidnightBlue!10}{0.243} & \colorbox{MidnightBlue!10}{0.067} & \colorbox{MidnightBlue!10}{0.243} & \colorbox{MidnightBlue!10}{0.398} & \colorbox{MidnightBlue!10}{0.054} & \colorbox{MidnightBlue!10}{0.204} \\
\mistralM & \colorbox{MidnightBlue!30}{1.301} & \colorbox{MidnightBlue!30}{1.097} & \colorbox{MidnightBlue!30}{1.708} & \colorbox{MidnightBlue!10}{0.929} & \colorbox{MidnightBlue!10}{0.544} & \colorbox{MidnightBlue!10}{0.989} & \colorbox{MidnightBlue!30}{1.663} & \colorbox{MidnightBlue!30}{1.079} & \colorbox{MidnightBlue!10}{0.269} & \colorbox{MidnightBlue!10}{0.989} & \colorbox{MidnightBlue!30}{1.070} & \colorbox{MidnightBlue!30}{1.011} & \colorbox{MidnightBlue!10}{0.301} & \colorbox{MidnightBlue!10}{0.368} & \colorbox{MidnightBlue!10}{0.284} & \colorbox{MidnightBlue!10}{0.602} & \colorbox{MidnightBlue!10}{0.317} & \colorbox{MidnightBlue!10}{0.359} \\
\llamaM & \colorbox{MidnightBlue!10}{0.602} & \colorbox{MidnightBlue!10}{0.447} & \colorbox{MidnightBlue!10}{0.649} & \colorbox{MidnightBlue!10}{0.786} & \colorbox{MidnightBlue!10}{0.586} & \colorbox{MidnightBlue!10}{0.493} & \colorbox{MidnightBlue!10}{0.637} & \colorbox{MidnightBlue!10}{0.768} & \colorbox{MidnightBlue!10}{0.000} & \colorbox{MidnightBlue!10}{0.620} & NA & \colorbox{MidnightBlue!10}{0.929} & \colorbox{MidnightBlue!10}{0.097} & \colorbox{MidnightBlue!30}{1.301} & \colorbox{MidnightBlue!10}{0.046} & \colorbox{MidnightBlue!10}{0.155} & \colorbox{MidnightBlue!10}{0.447} & \colorbox{MidnightBlue!10}{0.084} \\
\midrule
\avg & \colorbox{MidnightBlue!10}{0.512} & \colorbox{MidnightBlue!10}{0.750} & \colorbox{MidnightBlue!10}{0.991} & \colorbox{MidnightBlue!10}{0.753} & \colorbox{MidnightBlue!10}{0.653} & \colorbox{MidnightBlue!10}{0.740} & \colorbox{MidnightBlue!30}{1.398} & \colorbox{MidnightBlue!30}{1.009} & \colorbox{MidnightBlue!10}{0.368} & NA & \colorbox{MidnightBlue!30}{1.699} & \colorbox{MidnightBlue!30}{1.156} & \colorbox{MidnightBlue!10}{0.146} & \colorbox{MidnightBlue!10}{0.279} & \colorbox{MidnightBlue!10}{0.133} & \colorbox{MidnightBlue!10}{0.398} & \colorbox{MidnightBlue!10}{0.187} & \colorbox{MidnightBlue!10}{0.150} \\
\midrule
\multicolumn{19}{|c|}{\textbf{\mam (self-attention)}} \\
\hline
\mistralS & \colorbox{MidnightBlue!10}{0.125} & \colorbox{MidnightBlue!10}{0.991} & \colorbox{MidnightBlue!30}{1.000} & \colorbox{MidnightBlue!10}{0.837} & \colorbox{MidnightBlue!10}{0.760} & \colorbox{MidnightBlue!30}{1.217} & \colorbox{MidnightBlue!10}{0.987} & \colorbox{MidnightBlue!30}{1.134} & \colorbox{MidnightBlue!10}{0.342} & \colorbox{MidnightBlue!30}{1.544} & \colorbox{MidnightBlue!30}{1.249} & \colorbox{MidnightBlue!30}{1.230} & \colorbox{MidnightBlue!10}{0.000} & \colorbox{MidnightBlue!10}{0.000} & \colorbox{MidnightBlue!10}{-0.041} & \colorbox{MidnightBlue!10}{0.222} & \colorbox{MidnightBlue!10}{0.150} & \colorbox{MidnightBlue!10}{0.130} \\
\llamaS & \colorbox{MidnightBlue!10}{0.204} & \colorbox{MidnightBlue!10}{0.582} & \colorbox{MidnightBlue!10}{0.663} & \colorbox{MidnightBlue!10}{0.621} & \colorbox{MidnightBlue!30}{1.447} & \colorbox{MidnightBlue!10}{0.648} & \colorbox{MidnightBlue!30}{1.186} & \colorbox{MidnightBlue!10}{0.750} & UN & \colorbox{MidnightBlue!10}{0.716} & \colorbox{MidnightBlue!10}{0.711} & \colorbox{MidnightBlue!30}{1.000} & \colorbox{MidnightBlue!10}{0.368} & \colorbox{MidnightBlue!10}{0.105} & \colorbox{MidnightBlue!10}{0.281} & \colorbox{MidnightBlue!10}{0.301} & \colorbox{MidnightBlue!10}{0.117} & \colorbox{MidnightBlue!10}{0.234} \\
\mistralM & \colorbox{MidnightBlue!30}{1.000} & \colorbox{MidnightBlue!10}{0.854} & \colorbox{MidnightBlue!30}{1.009} & \colorbox{MidnightBlue!10}{0.804} & \colorbox{MidnightBlue!10}{0.447} & \colorbox{MidnightBlue!30}{1.114} & \colorbox{MidnightBlue!30}{1.362} & \colorbox{MidnightBlue!30}{1.681} & \colorbox{MidnightBlue!10}{0.637} & \colorbox{MidnightBlue!30}{1.290} & \colorbox{MidnightBlue!30}{1.672} & NA & \colorbox{MidnightBlue!10}{0.477} & \colorbox{MidnightBlue!30}{1.021} & \colorbox{MidnightBlue!10}{0.319} & \colorbox{MidnightBlue!10}{0.301} & \colorbox{MidnightBlue!10}{0.653} & \colorbox{MidnightBlue!10}{0.602} \\
\llamaM & \colorbox{MidnightBlue!10}{0.234} & \colorbox{MidnightBlue!10}{0.419} & \colorbox{MidnightBlue!10}{0.576} & \colorbox{MidnightBlue!10}{0.661} & \colorbox{MidnightBlue!10}{0.586} & \colorbox{MidnightBlue!10}{0.845} & \colorbox{MidnightBlue!10}{0.746} & \colorbox{MidnightBlue!30}{1.011} & \colorbox{MidnightBlue!10}{-0.067} & \colorbox{MidnightBlue!10}{0.699} & \colorbox{MidnightBlue!30}{1.672} & \colorbox{MidnightBlue!10}{0.686} & \colorbox{MidnightBlue!10}{0.222} & NA & \colorbox{MidnightBlue!10}{0.097} & \colorbox{MidnightBlue!10}{0.097} & \colorbox{MidnightBlue!10}{0.368} & \colorbox{MidnightBlue!10}{0.084} \\
\midrule
\avg & \colorbox{MidnightBlue!10}{0.336} & \colorbox{MidnightBlue!10}{0.653} & \colorbox{MidnightBlue!10}{0.736} & \colorbox{MidnightBlue!10}{0.708} & \colorbox{MidnightBlue!10}{0.732} & \colorbox{MidnightBlue!10}{0.944} & NA & \colorbox{MidnightBlue!30}{1.009} & \colorbox{MidnightBlue!10}{0.544} & NA & \colorbox{MidnightBlue!30}{1.097} & \colorbox{MidnightBlue!30}{1.332} & \colorbox{MidnightBlue!10}{0.243} & \colorbox{MidnightBlue!10}{0.376} & \colorbox{MidnightBlue!10}{0.165} & \colorbox{MidnightBlue!10}{0.523} & \colorbox{MidnightBlue!10}{0.301} & \colorbox{MidnightBlue!10}{0.234} \\
\midrule
\multicolumn{19}{|c|}{\textbf{SelfCheckGPT}} \\
\hline
\mistralS & \colorbox{MidnightBlue!10}{0.778} & \colorbox{MidnightBlue!10}{0.991} & \colorbox{MidnightBlue!30}{1.000} & \colorbox{MidnightBlue!10}{0.786} & \colorbox{MidnightBlue!10}{0.760} & \colorbox{MidnightBlue!10}{0.865} & \colorbox{MidnightBlue!10}{0.577} & \colorbox{MidnightBlue!10}{0.833} & \colorbox{MidnightBlue!10}{0.439} & \colorbox{MidnightBlue!30}{1.368} & \colorbox{MidnightBlue!30}{1.249} & \colorbox{MidnightBlue!10}{0.833} & \colorbox{MidnightBlue!10}{0.544} & \colorbox{MidnightBlue!10}{0.677} & \colorbox{MidnightBlue!10}{0.699} & \colorbox{MidnightBlue!30}{1.000} & \colorbox{MidnightBlue!10}{0.681} & \colorbox{MidnightBlue!10}{0.653} \\
\llamaS & \colorbox{MidnightBlue!10}{0.903} & \colorbox{MidnightBlue!30}{1.146} & \colorbox{MidnightBlue!30}{1.186} & \colorbox{MidnightBlue!30}{1.362} & \colorbox{MidnightBlue!30}{1.447} & \colorbox{MidnightBlue!10}{0.561} & \colorbox{MidnightBlue!10}{0.362} & \colorbox{MidnightBlue!10}{0.539} & UN & \colorbox{MidnightBlue!10}{0.336} & \colorbox{MidnightBlue!10}{0.711} & \colorbox{MidnightBlue!10}{0.477} & \colorbox{MidnightBlue!10}{0.845} & \colorbox{MidnightBlue!10}{0.368} & \colorbox{MidnightBlue!10}{0.419} & \colorbox{MidnightBlue!10}{0.398} & \colorbox{MidnightBlue!10}{0.385} & \colorbox{MidnightBlue!10}{0.477} \\
\mistralM & \colorbox{MidnightBlue!10}{0.347} & \colorbox{MidnightBlue!10}{0.585} & \colorbox{MidnightBlue!10}{0.708} & \colorbox{MidnightBlue!10}{0.753} & \colorbox{MidnightBlue!10}{0.669} & \colorbox{MidnightBlue!10}{0.591} & \colorbox{MidnightBlue!30}{1.061} & \colorbox{MidnightBlue!10}{0.426} & \colorbox{MidnightBlue!10}{0.000} & \colorbox{MidnightBlue!10}{0.688} & \colorbox{MidnightBlue!10}{0.769} & \colorbox{MidnightBlue!30}{1.136} & NA & \colorbox{MidnightBlue!10}{0.368} & \colorbox{MidnightBlue!10}{0.444} & \colorbox{MidnightBlue!30}{1.079} & NA & \colorbox{MidnightBlue!10}{0.903} \\
\llamaM & \colorbox{MidnightBlue!10}{0.301} & \colorbox{MidnightBlue!10}{0.544} & \colorbox{MidnightBlue!10}{0.649} & \colorbox{MidnightBlue!30}{1.041} & \colorbox{MidnightBlue!10}{0.317} & \colorbox{MidnightBlue!10}{0.125} & \colorbox{MidnightBlue!10}{0.688} & \colorbox{MidnightBlue!10}{0.768} & \colorbox{MidnightBlue!10}{-0.176} & \colorbox{MidnightBlue!30}{1.097} & \colorbox{MidnightBlue!10}{0.827} & \colorbox{MidnightBlue!30}{1.054} & \colorbox{MidnightBlue!10}{0.398} & \colorbox{MidnightBlue!30}{1.301} & \colorbox{MidnightBlue!30}{1.000} & \colorbox{MidnightBlue!30}{1.000} & \colorbox{MidnightBlue!10}{0.845} & \colorbox{MidnightBlue!30}{1.230} \\
\midrule
\avg & \colorbox{MidnightBlue!10}{0.512} & \colorbox{MidnightBlue!10}{0.699} & \colorbox{MidnightBlue!10}{0.787} & \colorbox{MidnightBlue!10}{0.862} & \colorbox{MidnightBlue!10}{0.829} & \colorbox{MidnightBlue!10}{0.530} & \colorbox{MidnightBlue!10}{0.620} & \colorbox{MidnightBlue!10}{0.804} & \colorbox{MidnightBlue!10}{0.146} & \colorbox{MidnightBlue!30}{1.301} & \colorbox{MidnightBlue!30}{1.699} & \colorbox{MidnightBlue!30}{1.156} & \colorbox{MidnightBlue!10}{0.845} & \colorbox{MidnightBlue!10}{0.580} & \colorbox{MidnightBlue!10}{0.580} & \colorbox{MidnightBlue!10}{0.699} & \colorbox{MidnightBlue!10}{0.699} & \colorbox{MidnightBlue!10}{0.681} \\
\midrule
\multicolumn{19}{|c|}{\textbf{Semantic Entropy}} \\
\hline
\mistralS & \colorbox{MidnightBlue!10}{0.778} & \colorbox{MidnightBlue!10}{0.611} & \colorbox{MidnightBlue!10}{0.745} & \colorbox{MidnightBlue!10}{0.594} & \colorbox{MidnightBlue!10}{0.517} & \colorbox{MidnightBlue!30}{1.121} & \colorbox{MidnightBlue!10}{0.878} & \colorbox{MidnightBlue!10}{0.753} & \colorbox{MidnightBlue!10}{0.564} & \colorbox{MidnightBlue!30}{1.067} & \colorbox{MidnightBlue!10}{0.810} & \colorbox{MidnightBlue!10}{0.686} & \colorbox{MidnightBlue!10}{0.544} & \colorbox{MidnightBlue!10}{0.434} & \colorbox{MidnightBlue!10}{0.398} & \colorbox{MidnightBlue!10}{0.523} & \colorbox{MidnightBlue!10}{0.426} & \colorbox{MidnightBlue!10}{0.431} \\
\llamaS & \colorbox{MidnightBlue!10}{0.125} & \colorbox{MidnightBlue!10}{0.544} & \colorbox{MidnightBlue!10}{0.517} & \colorbox{MidnightBlue!10}{0.459} & \colorbox{MidnightBlue!10}{0.333} & \colorbox{MidnightBlue!10}{0.648} & \colorbox{MidnightBlue!10}{0.885} & \colorbox{MidnightBlue!10}{0.653} & UN & \colorbox{MidnightBlue!10}{0.938} & \colorbox{MidnightBlue!30}{1.556} & \colorbox{MidnightBlue!30}{1.000} & \colorbox{MidnightBlue!10}{0.544} & \colorbox{MidnightBlue!10}{0.192} & \colorbox{MidnightBlue!10}{0.208} & \colorbox{MidnightBlue!30}{1.000} & \colorbox{MidnightBlue!10}{0.385} & \colorbox{MidnightBlue!10}{0.380} \\
\mistralM & \colorbox{MidnightBlue!10}{0.301} & \colorbox{MidnightBlue!10}{0.585} & \colorbox{MidnightBlue!10}{0.666} & \colorbox{MidnightBlue!10}{0.561} & \colorbox{MidnightBlue!10}{0.669} & \colorbox{MidnightBlue!10}{0.813} & \colorbox{MidnightBlue!10}{0.709} & \colorbox{MidnightBlue!10}{0.640} & \colorbox{MidnightBlue!10}{0.415} & \colorbox{MidnightBlue!10}{0.591} & \colorbox{MidnightBlue!10}{0.973} & \colorbox{MidnightBlue!10}{0.613} & \colorbox{MidnightBlue!10}{0.079} & \colorbox{MidnightBlue!10}{0.720} & \colorbox{MidnightBlue!10}{0.921} & \colorbox{MidnightBlue!10}{0.380} & \colorbox{MidnightBlue!30}{1.431} & \colorbox{MidnightBlue!30}{1.505} \\
\llamaM & \colorbox{MidnightBlue!10}{0.234} & \colorbox{MidnightBlue!10}{0.669} & \colorbox{MidnightBlue!30}{1.088} & \colorbox{MidnightBlue!30}{1.439} & \colorbox{MidnightBlue!10}{0.477} & \colorbox{MidnightBlue!10}{0.493} & \colorbox{MidnightBlue!30}{1.591} & \colorbox{MidnightBlue!10}{0.835} & \colorbox{MidnightBlue!10}{0.301} & \colorbox{MidnightBlue!10}{0.921} & \colorbox{MidnightBlue!30}{1.672} & \colorbox{MidnightBlue!10}{0.490} & NA & \colorbox{MidnightBlue!30}{1.301} & \colorbox{MidnightBlue!10}{0.699} & NA & \colorbox{MidnightBlue!30}{1.146} & NA \\
\midrule
\avg & \colorbox{MidnightBlue!10}{0.269} & \colorbox{MidnightBlue!10}{0.574} & \colorbox{MidnightBlue!10}{0.690} & \colorbox{MidnightBlue!10}{0.628} & \colorbox{MidnightBlue!10}{0.477} & \colorbox{MidnightBlue!10}{0.798} & \colorbox{MidnightBlue!10}{0.854} & \colorbox{MidnightBlue!10}{0.862} & \colorbox{MidnightBlue!10}{0.544} & \colorbox{MidnightBlue!30}{1.602} & \colorbox{MidnightBlue!30}{1.398} & \colorbox{MidnightBlue!30}{1.633} & \colorbox{MidnightBlue!10}{0.845} & \colorbox{MidnightBlue!10}{0.580} & \colorbox{MidnightBlue!10}{0.501} & \colorbox{MidnightBlue!30}{1.000} & \colorbox{MidnightBlue!10}{0.699} & \colorbox{MidnightBlue!10}{0.681} \\
\bottomrule
\end{tabular}
}
\caption{\small \metric scores ($\log_{10}$ of ratio of accuracy delta to AUROC delta w.r.t \en) for \textbf{\mam (fully connected activations)}, \textbf{\mam (self-attention)}, \textbf{SelfCheckGPT}, and \textbf{Semantic Entropy} methods. Cells are marked as NA when the AUROC delta is zero, and as UN when the task accuracy delta is zero. Darker color shades indicate a higher \metric value, signifying that although the task accuracy for these languages drops drastically, the HD's performance remains much more stable.}
\label{table:ratio_acc_auc_tphr_log10}
\vspace{-4mm}
\end{table*}

%% file: final_tables/mam_attn_cross_table_main.tex
\begin{table*}[!htb]
\centering
\small
\resizebox{\textwidth}{!}{
\begin{tabular}{l|ccccc|ccccc|ccccc|cccc|cccc|}
\toprule
\multirow{2}{*}{\bf Models} & \multicolumn{5}{|c|}{\bf mTREx -- Capitals} & \multicolumn{5}{|c|}{\bf mTREx -- Country} & \multicolumn{5}{|c|}{\bf mTREx -- Official Language} & \multicolumn{4}{|c|}{\bf G-MMLU -- STEM} & \multicolumn{4}{|c|}{\bf G-MMLU -- Humanities} \\
 & \en & \de & \hi & \bn & \ur & \en & \de & \hi & \bn & \ur & \en & \de & \hi & \bn & \ur & \en & \de & \hi & \bn & \en & \de & \hi & \bn \\
\midrule
\mistralS & 79 & \colorbox{red!30}{\scriptsize ↓28} & \colorbox{red!30}{\scriptsize ↓32} & \colorbox{red!30}{\scriptsize ↓35} & \colorbox{red!40}{\scriptsize ↓48} & 77 & \colorbox{red!20}{\scriptsize ↓18} & \colorbox{red!30}{\scriptsize ↓23} & \colorbox{red!30}{\scriptsize ↓35} & \colorbox{red!30}{\scriptsize ↓22} & 78 & \colorbox{red!20}{\scriptsize ↓19} & \colorbox{red!30}{\scriptsize ↓24} & \colorbox{red!30}{\scriptsize ↓37} & \colorbox{red!40}{\scriptsize ↓41} & 73 & \colorbox{red!20}{\scriptsize ↓14} & \colorbox{red!30}{\scriptsize ↓27} & \colorbox{red!30}{\scriptsize ↓26} & 72 & \colorbox{red!10}{\scriptsize ↓08} & \colorbox{red!30}{\scriptsize ↓25} & \colorbox{red!30}{\scriptsize ↓25} \\
\llamaS & 79 & \colorbox{red!20}{\scriptsize ↓15} & \colorbox{red!30}{\scriptsize ↓33} & \colorbox{red!30}{\scriptsize ↓29} & \colorbox{red!30}{\scriptsize ↓28} & 86 & \colorbox{red!30}{\scriptsize ↓31} & \colorbox{red!30}{\scriptsize ↓27} & \colorbox{red!30}{\scriptsize ↓24} & \colorbox{red!30}{\scriptsize ↓37} & 88 & \colorbox{red!30}{\scriptsize ↓22} & \colorbox{red!40}{\scriptsize ↓47} & \colorbox{red!30}{\scriptsize ↓25} & \colorbox{red!40}{\scriptsize ↓44} & 77 & \colorbox{red!10}{\scriptsize ↓06} & \colorbox{red!20}{\scriptsize ↓13} & \colorbox{red!20}{\scriptsize ↓16} & 78 & \colorbox{red!20}{\scriptsize ↓17} & \colorbox{red!30}{\scriptsize ↓21} & \colorbox{red!20}{\scriptsize ↓19} \\
\mistralM & 78 & \colorbox{red!30}{\scriptsize ↓21} & \colorbox{red!30}{\scriptsize ↓29} & \colorbox{red!30}{\scriptsize ↓24} & \colorbox{red!30}{\scriptsize ↓28} & 83 & \colorbox{red!30}{\scriptsize ↓27} & \colorbox{red!30}{\scriptsize ↓24} & \colorbox{red!30}{\scriptsize ↓24} & \colorbox{red!30}{\scriptsize ↓33} & 86 & \colorbox{red!20}{\scriptsize ↓12} & \colorbox{red!30}{\scriptsize ↓36} & \colorbox{red!30}{\scriptsize ↓24} & \colorbox{red!30}{\scriptsize ↓31} & 74 & \colorbox{red!10}{\scriptsize ↓09} & \colorbox{red!20}{\scriptsize ↓16} & \colorbox{red!30}{\scriptsize ↓24} & 70 & \colorbox{green!10}{\scriptsize ↑03} & \colorbox{red!30}{\scriptsize ↓21} & \colorbox{red!30}{\scriptsize ↓20} \\
\llamaM & 82 & \colorbox{red!30}{\scriptsize ↓21} & \colorbox{red!30}{\scriptsize ↓34} & \colorbox{red!30}{\scriptsize ↓30} & \colorbox{red!30}{\scriptsize ↓33} & 86 & \colorbox{red!30}{\scriptsize ↓21} & \colorbox{red!30}{\scriptsize ↓35} & \colorbox{red!20}{\scriptsize ↓17} & \colorbox{red!30}{\scriptsize ↓34} & 82 & \colorbox{red!30}{\scriptsize ↓23} & \colorbox{red!30}{\scriptsize ↓33} & \colorbox{red!30}{\scriptsize ↓26} & \colorbox{red!30}{\scriptsize ↓34} & 83 & \colorbox{red!10}{\scriptsize ↓05} & \colorbox{red!20}{\scriptsize ↓16} & \colorbox{red!20}{\scriptsize ↓12} & 86 & \colorbox{red!20}{\scriptsize ↓12} & \colorbox{red!20}{\scriptsize ↓19} & \colorbox{red!30}{\scriptsize ↓21} \\
\midrule
\avg & 80 & \colorbox{red!30}{\scriptsize ↓22} & \colorbox{red!30}{\scriptsize ↓32} & \colorbox{red!30}{\scriptsize ↓30} & \colorbox{red!30}{\scriptsize ↓35} & 83 & \colorbox{red!30}{\scriptsize ↓25} & \colorbox{red!30}{\scriptsize ↓28} & \colorbox{red!30}{\scriptsize ↓25} & \colorbox{red!30}{\scriptsize ↓33} & 84 & \colorbox{red!30}{\scriptsize ↓20} & \colorbox{red!30}{\scriptsize ↓36} & \colorbox{red!30}{\scriptsize ↓29} & \colorbox{red!30}{\scriptsize ↓38} & 77 & \colorbox{red!10}{\scriptsize ↓09} & \colorbox{red!20}{\scriptsize ↓19} & \colorbox{red!20}{\scriptsize ↓19} & 76 & \colorbox{red!10}{\scriptsize ↓09} & \colorbox{red!30}{\scriptsize ↓21} & \colorbox{red!30}{\scriptsize ↓21} \\
\bottomrule
\end{tabular}
}
\caption{\small Percentage of increment \colorbox{green!17}{\scriptsize ↑} or decrement \colorbox{red!17}{\scriptsize ↓} in AUROC scores for \mistralS, \llamaS, \mistralM, and \llamaM across languages for \textbf{\mam (self-attention)} method w.r.t. the corresponding \en baseline for the model and dataset in the cross-lingual setting (train over \en data and test over target language data).}
\label{table:mam-self-attn-cross-lingual_delta_percentage_main}
\vspace{-2mm}
\end{table*}

\begin{table*}[!htb]
\centering
\small
\resizebox{\textwidth}{!}{
\begin{tabular}{l|ccccc|ccccc|ccccc|cccc|cccc|}
\toprule
\multirow{2}{*}{\bf Models} & \multicolumn{5}{|c|}{\bf mTREx -- Capitals} & \multicolumn{5}{|c|}{\bf mTREx -- Country} & \multicolumn{5}{|c|}{\bf mTREx -- Official Language} & \multicolumn{4}{|c|}{\bf G-MMLU -- STEM} & \multicolumn{4}{|c|}{\bf G-MMLU -- Humanities} \\
 & \en & \de & \hi & \bn & \ur & \en & \de & \hi & \bn & \ur & \en & \de & \hi & \bn & \ur & \en & \de & \hi & \bn & \en & \de & \hi & \bn \\
\midrule
\mistralS & 78 & \colorbox{red!20}{\scriptsize ↓12} & \colorbox{red!10}{\scriptsize ↓08} & \colorbox{red!10}{\scriptsize ↓04} & \colorbox{red!30}{\scriptsize ↓22} & 79 & \colorbox{red!10}{\scriptsize ↓01} & \colorbox{red!10}{\scriptsize ↓06} & {\scriptsize ↓0} & \colorbox{red!20}{\scriptsize ↓11} & 74 & \colorbox{green!10}{\scriptsize ↑09} & \colorbox{red!10}{\scriptsize ↓04} & \colorbox{red!10}{\scriptsize ↓05} & \colorbox{green!10}{\scriptsize ↑05} & 74 & \colorbox{red!20}{\scriptsize ↓11} & \colorbox{red!30}{\scriptsize ↓20} & \colorbox{red!30}{\scriptsize ↓28} & 68 & \colorbox{red!10}{\scriptsize ↓03} & \colorbox{red!20}{\scriptsize ↓19} & \colorbox{red!30}{\scriptsize ↓24} \\
\llamaS & 80 & \colorbox{red!20}{\scriptsize ↓11} & \colorbox{red!20}{\scriptsize ↓19} & \colorbox{red!20}{\scriptsize ↓12} & \colorbox{red!20}{\scriptsize ↓16} & 83 & \colorbox{green!10}{\scriptsize ↑01} & \colorbox{red!10}{\scriptsize ↓06} & {\scriptsize ↓0} & \colorbox{red!10}{\scriptsize ↓07} & 88 & \colorbox{red!10}{\scriptsize ↓05} & \colorbox{red!10}{\scriptsize ↓06} & \colorbox{red!20}{\scriptsize ↓10} & \colorbox{red!10}{\scriptsize ↓07} & 77 & \colorbox{red!10}{\scriptsize ↓04} & \colorbox{red!20}{\scriptsize ↓10} & \colorbox{red!20}{\scriptsize ↓10} & 76 & \colorbox{red!10}{\scriptsize ↓03} & \colorbox{red!20}{\scriptsize ↓14} & \colorbox{red!20}{\scriptsize ↓17} \\
\mistralM & 72 & \colorbox{green!10}{\scriptsize ↑03} & \colorbox{red!10}{\scriptsize ↓07} & \colorbox{green!10}{\scriptsize ↑01} & \colorbox{red!10}{\scriptsize ↓04} & 82 & \colorbox{green!20}{\scriptsize ↑12} & \colorbox{red!10}{\scriptsize ↓05} & \colorbox{green!10}{\scriptsize ↑02} & \colorbox{red!10}{\scriptsize ↓02} & 84 & \colorbox{green!10}{\scriptsize ↑05} & \colorbox{red!10}{\scriptsize ↓01} & {\scriptsize ↓0} & {\scriptsize ↓0} & 77 & \colorbox{red!10}{\scriptsize ↓06} & \colorbox{red!10}{\scriptsize ↓09} & \colorbox{red!20}{\scriptsize ↓19} & 71 & \colorbox{green!10}{\scriptsize ↑08} & \colorbox{red!20}{\scriptsize ↓13} & \colorbox{red!20}{\scriptsize ↓13} \\
\llamaM & 82 & \colorbox{red!10}{\scriptsize ↓09} & \colorbox{red!20}{\scriptsize ↓17} & \colorbox{red!20}{\scriptsize ↓16} & \colorbox{red!20}{\scriptsize ↓11} & 86 & \colorbox{green!10}{\scriptsize ↑07} & \colorbox{red!10}{\scriptsize ↓05} & \colorbox{red!20}{\scriptsize ↓10} & \colorbox{red!10}{\scriptsize ↓05} & 78 & \colorbox{green!20}{\scriptsize ↑12} & \colorbox{green!10}{\scriptsize ↑09} & {\scriptsize ↓0} & \colorbox{green!20}{\scriptsize ↑10} & 85 & \colorbox{red!10}{\scriptsize ↓05} & {\scriptsize ↓0} & \colorbox{red!20}{\scriptsize ↓12} & 85 & \colorbox{red!10}{\scriptsize ↓08} & \colorbox{red!10}{\scriptsize ↓06} & \colorbox{red!20}{\scriptsize ↓15} \\
\midrule
\avg & 78 & \colorbox{red!10}{\scriptsize ↓08} & \colorbox{red!20}{\scriptsize ↓13} & \colorbox{red!10}{\scriptsize ↓09} & \colorbox{red!20}{\scriptsize ↓14} & 82 & \colorbox{green!10}{\scriptsize ↑05} & \colorbox{red!10}{\scriptsize ↓05} & \colorbox{red!10}{\scriptsize ↓02} & \colorbox{red!10}{\scriptsize ↓06} & 80 & \colorbox{green!10}{\scriptsize ↑05} & 80 & \colorbox{red!10}{\scriptsize ↓04} & \colorbox{green!10}{\scriptsize ↑02} & 77 & \colorbox{red!10}{\scriptsize ↓05} & \colorbox{red!10}{\scriptsize ↓09} & \colorbox{red!20}{\scriptsize ↓17} & 75 & \colorbox{red!10}{\scriptsize ↓03} & \colorbox{red!20}{\scriptsize ↓13} & \colorbox{red!20}{\scriptsize ↓17} \\
\bottomrule
\end{tabular}
}
\caption{\small Percentage of increment \colorbox{green!17}{\scriptsize ↑} or decrement \colorbox{red!17}{\scriptsize ↓} in AUROC scores for \mistralS, \llamaS, \mistralM, and \llamaM across languages for \textbf{\mam (self-attention)} method w.r.t. the corresponding \en baseline for the model and dataset in the multilingual setting (train over data across all languages and test over target language data).}
\label{table:mam-self-attn-multi-lingual_delta_percentage_main}
\vspace{-4mm}
\end{table*}

%% file: 06_conclusion.tex
\section{Conclusion}

We present a comprehensive study of LLM performance and hallucination detection in a multilingual QA setting, especially with resource-poor languages. 
Our findings reveal that although there is a consistent drop in task accuracy of LLMs for low-resource languages, the performance of hallucination detectors (HDs) remains relatively stable across languages.
To quantify this phenomenon, we introduce a novel metric (\metric) which reveals that HDs suffer significantly less from the low-resource effect than the underlying LLMs.
We hypothesize that this robustness may be since hallucination detection is an easier binary classification problem than generation that requires predicting the correct token sequence from a large vocabulary at each step.
Our findings also show that HDs that leverage models' internal artifacts (while potentially benefiting from explicit supervision over these internal states) serve as more stable indicators of hallucination across languages than blackbox approaches that rely solely on generated responses.

%% file: 07_limitations_ethics.tex
\section*{Acknowledgments}

This research was partly conducted during Debtanu Datta's visit to Ruhr-Universität Bochum, Germany. The visit was supported by the UAR Research Center for Trustworthy Data Science and Security (RC Trust), Germany. Debtanu Datta is also supported by the Prime Minister's Research Fellowship (PMRF) from the Government of India.

\section*{Limitations}

While our study provides valuable insights regarding hallucination detectors' performance in low-resource languages, several limitations remain. We focus only on two specific types of question-answering tasks, where model responses contain a single true answer. Extending our analyses on tasks such as dialogue generation---where multiple hallucinations can be observed in a single generation---can be considered as a direction for future work. Extending the analysis to a broader range of QA tasks, domains, and generation types is also an important future direction. 

Also, hallucinations can arise from diverse underlying sources, such as reasoning failures, lack of knowledge, or the generation of fabricated information, and these factors may play distinct roles in model behavior. However, the scope of this study is limited to a comparative evaluation of hallucination detector performance across languages, rather than an analysis or classification of the root causes of hallucinations. We therefore leave a systematic investigation of hallucination sources as an interesting direction for future work.

In this study, we evaluate five languages (\en, \de, \hi, \bn, and \ur) that use four distinct scripts (Latin, Devanagari, Perso-Arabic, Bengali) and represent a range of resource levels. 
This diversity suggests that our framework should extend to other languages, but of course, a dedicated analysis is needed. 
Incorporating more languages would further generalize our findings. 
It can be noted that the detectors we study are largely language-agnostic and domain-agnostic, so they can be extended beyond the five languages and factual-QA tasks. Specifically, MAM detectors require access to model artifacts (self-attention, activations, etc.) and therefore could extend readily to any language that an LLM supports. Sampling-based detectors (SEM and SGM) are black-box and thus easily applied to new languages and domains. 




%% file: 08_appendix.tex
\section{Cleaning, Predicate Selection and Filtering Criteria for TREx}
\label{appendix:trex_filtering}
First, we cleaned the TREx dataset by removing entries where the subject or object is missing and by excluding entries where the subject or object is a pronoun (e.g., `he', `she', `it', `we', `they'), as these lack the specificity needed for factual relationships. 
Next, several predicates from the original TREx dataset are excluded based on qualitative and quantitative considerations. 
The \textit{Continent} predicate is discarded due to class imbalance, with `Antarctica' appearing as the answer for more than half of the instances. The \textit{Discoverer or Inventor} predicate exhibited ambiguity in subject references, creating difficulty in ensuring consistent interpretation. The \textit{Head of Government} predicate was outdated in many entries, reducing its relevance for current factual evaluation of LLMs. The \textit{Award Received} predicate included numerous acronyms and abbreviations that the translation models failed to translate accurately, and are prone to being accepted universally without proper understanding in the case of multilingual settings. Additionally, several other predicates -- such as \textit{Basic Form of Government}, \textit{Currency}, \textit{Location of Discovery}, and \textit{Symptoms and Signs} -- are excluded due to insufficient data coverage (fewer than 500 instances). By filtering predicates with high semantic clarity, sufficient scale, and multilingual relevance, we aim to ensure a robust and fair benchmark for hallucination detection across languages.

\section{LLM versions, Hyperparameters \& Infrastructure}
\label{appendix:expt-settings}

\noindent \textbf{LLM versions:} In this study, we use the following LLMs whose sizes (number of parameters) vary from 7 billion to 70 billion:
\begin{enumerate}
    \item Mistral-7B-Instruct (\textbf{\mistralS})\footnote{\url{https://huggingface.co/mistralai/Mistral-7B-Instruct-v0.3}}
    \item LLaMA-8B-Instruct (\textbf{\llamaS})\footnote{\url{https://huggingface.co/meta-llama/Llama-3.1-8B-Instruct}}
    \item Mistral-24B-Instruct (\textbf{\mistralM})\footnote{\url{https://huggingface.co/mistralai/Mistral-Small-24B-Instruct-2501}}
    \item LLaMA-70B-Instruct (\textbf{\llamaM})\footnote{\url{https://huggingface.co/meta-llama/Llama-3.3-70B-Instruct}}
\end{enumerate}

\noindent \textbf{Hyperparameters \& Infrastructure:} All experiments were conducted on a server equipped with 8x NVIDIA H200 GPUs (shared with several other researchers). 
For all three HD methods, the base LLM responses are generated by sampling the most likely token according to the Softmax probability, i.e., the greedy decoding with \textit{temperature} as 0, continuing until an <end of text> token is reached or the output length reaches 50 tokens. 
For \sem and \sgm methods, we generate 20 sample responses for an input with a high temperature of 1.0 to compare with the base response at zero temperature. 
The classifiers for the Model Artifacts method are trained using the Adam optimizer with a learning rate of $10^{-4}$, weight decay of $10^{-2}$, a batch size of 128, and for 1000 iterations. 

\section{Prompts}

\subsection{Prompt Templates for \mtrex}
\label{appendix:prompts_trex}

In this section, we describe the exact prompt templates used for all five languages and three relation types (Capitals, Country, and Official Language) for the \mtrex Dataset. For each language, the prompt comprises a general instruction (\textit{system prompt}) followed by a language-specific question (\textit{user prompt}) with a designated answer placeholder (\textit{assistant prompt}). Figures \ref{fig:eng-mtrex-prompt-appn}, \ref{fig:ger-mtrex-prompt-appn}, \ref{fig:hin-mtrex-prompt-appn}, \ref{fig:ben-mtrex-prompt-appn}, and \ref{fig:urd-mtrex-prompt-appn} present the detailed prompts for the \mtrex dataset in English, German, Hindi, Bengali, and Urdu, respectively.

\begin{figure}[!htb]
\centering
\begin{tcolorbox}[colback=gray!15, colframe=black, boxrule=1pt, width=0.49\textwidth, fontupper=\normalfont\small, boxsep=1pt, left=2pt, right=2pt, top=1pt, bottom=1pt, halign=justify] 
\noindent \textbf{System Prompt:} Suppose your job is to answer given questions in English. When you are asked a question in English, please give a brief answer in English only. Do not write anything else except the answer. 
\\ [0.3em]
\textbf{User Prompt (Capitals):} Question: What is the capital of \texttt{<X>}? 
\\ [0.3em]
\textbf{User Prompt (Country):} Question: In which country is the \texttt{<X>} located? 
\\ [0.3em]
\textbf{User Prompt (Official Language):} Question: What is the official language of the \texttt{<X>}? 
\\ [0.3em]
\textbf{Assistant prompt:} Answer (in English):
\end{tcolorbox}
\caption{Prompt for \mtrex in English.}
\label{fig:eng-mtrex-prompt-appn}

\end{figure}

\begin{figure}[!htb]
\centering
\begin{tcolorbox}[colback=gray!15, colframe=black, boxrule=1pt, width=0.49\textwidth, fontupper=\normalfont\small, boxsep=1pt, left=2pt, right=2pt, top=1pt, bottom=1pt, halign=justify] 
\noindent \textbf{System Prompt:} Angenommen, Ihre Aufgabe ist es, Fragen auf Deutsch zu beantworten. Wenn Ihnen eine Frage auf Deutsch gestellt wird, geben Sie bitte eine kurze Antwort auf Deutsch. Nichts außer der Antwort. 
\\ [0.3em]
\textbf{User Prompt (Capitals):} Frage: Was ist die Hauptstadt von \texttt{<X>}? 
\\ [0.3em]
\textbf{User Prompt (Country):} Frage: In welchem Land liegt die \texttt{<X>}? 
\\ [0.3em]
\textbf{User Prompt (Official Language):} Frage: Was ist die Amtssprache von \texttt{<X>}? 
\\ [0.3em]
\textbf{Assistant prompt:} Antwort (auf Deutsch):
\end{tcolorbox}
\caption{Prompt for \mtrex in German.}
\label{fig:ger-mtrex-prompt-appn}
\end{figure}

\begin{figure}[!htb]
\centering
\begin{tcolorbox}[colback=gray!15, colframe=black, boxrule=1pt, width=0.49\textwidth, fontupper=\normalfont\small, boxsep=1pt, left=2pt, right=2pt, top=1pt, bottom=1pt, halign=justify] 

\noindent \textbf{System Prompt:} \foreignlanguage{hindi}{मान लीजिए कि आपका काम हिंदी में सवालों के जवाब देना है। जब आपसे हिंदी में कोई सवाल पूछा जाए, तो कृपया उसका संक्षिप्त जवाब हिंदी में ही दें। जवाब के अलावा कुछ और न लिखें।} 
\\ [0.3em]
\textbf{User Prompt (Capitals):} \foreignlanguage{hindi}{प्रश्न: \texttt{<X>} की राजधानी क्या है?} 
\\ [0.3em]
\textbf{User Prompt (Country):} \foreignlanguage{hindi}{प्रश्न: \texttt{<X>} किस देश में स्थित है?} 
\\ [0.3em]
\textbf{User Prompt (Official Language):} \foreignlanguage{hindi}{प्रश्न: \texttt{<X>} की आधिकारिक भाषा क्या है?} 
\\ [0.3em]
\textbf{Assistant prompt:} \foreignlanguage{hindi}{उत्तर (हिंदी में):}
\end{tcolorbox}
\caption{Prompt for \mtrex in Hindi.}
\label{fig:hin-mtrex-prompt-appn}
\end{figure}

\begin{figure}[!htb]
\centering
\begin{tcolorbox}[colback=gray!15, colframe=black, boxrule=1pt, width=0.49\textwidth, fontupper=\normalfont\small, boxsep=1pt, left=2pt, right=2pt, top=1pt, bottom=1pt, halign=justify] 
\noindent \textbf{System Prompt:} \foreignlanguage{bengali}{ধরুন আপনার কাজ হল বাংলায় প্রশ্নের উত্তর দেওয়া। যখন আপনাকে বাংলায় প্রশ্ন করা হবে, অনুগ্রহ করে শুধুমাত্র বাংলায় একটি সংক্ষিপ্ত উত্তর দিন। উত্তর ছাড়া অন্য কিছু লিখবেন না।} 
\\ [0.3em]
\textbf{User Prompt (Capitals):} \foreignlanguage{bengali}{প্রশ্ন: \texttt{<X>} এর রাজধানী কি?} 
\\ [0.3em]
\textbf{User Prompt (Country):} \foreignlanguage{bengali}{প্রশ্ন: কোন দেশে \texttt{<X>} অবস্থিত?} 
\\ [0.3em]
\textbf{User Prompt (Official Language):} \foreignlanguage{bengali}{প্রশ্ন: \texttt{<X>} এর সরকারী ভাষা কী?} 
\\ [0.3em]
\textbf{Assistant prompt:} \foreignlanguage{bengali}{উত্তর (বাংলায়):}
\end{tcolorbox}
\caption{Prompt for \mtrex in Bengali.}
\label{fig:ben-mtrex-prompt-appn}
\end{figure}

\begin{figure}[!htb]
\centering
\begin{tcolorbox}[colback=gray!15, colframe=black, boxrule=1pt, width=0.49\textwidth, fontupper=\normalfont\small, boxsep=1pt, left=2pt, right=2pt, top=1pt, bottom=1pt, halign=justify] 

\noindent \textbf{System Prompt:} \foreignlanguage{urdu}{فرض کریں آپ کا کام اردو میں سوالوں کے جواب دینا ہے۔ جب آپ سے اردو میں کوئی سوال پوچھا جائے تو برائے مہربانی مختصر جواب صرف اردو میں دیں۔ جواب کے علاوہ کچھ نہ لکھیں۔ } 
\\ [0.3em]
\textbf{User Prompt (Capitals):} \foreignlanguage{urdu}{سوال: \texttt{<X>} کا دارالحکومت کیا ہے؟ } 
\\ [0.3em]
\textbf{User Prompt (Country):} \foreignlanguage{urdu}{سوال: \texttt{<X>} کس ملک میں واقع ہے؟ }
\\ [0.3em]
\textbf{User Prompt (Official Language):} \foreignlanguage{urdu}{سوال: \texttt{<X>} کی سرکاری زبان کیا ہے؟ } 
\\ [0.3em]
\textbf{Assistant prompt:} \foreignlanguage{urdu}{جواب (اردو میں):}
\end{tcolorbox}
\caption{Prompt for \mtrex in Urdu.}
\label{fig:urd-mtrex-prompt-appn}
\vspace{-2mm}
\end{figure}

\subsection{Prompt Templates for \gmmlu}
\label{appendix:prompts_gmmlu}

We describe below the prompt templates used for all languages (English, German, Hindi, and Bengali) for the G-MMLU dataset. Each prompt includes a general instruction (\textit{system prompt}) followed by two language-specific in-context examples and the actual multiple-choice question (\textit{user prompt}). Figures \ref{fig:eng-gmmlu-prompt-appn}, \ref{fig:ger-gmmlu-prompt-appn}, \ref{fig:hin-gmmlu-prompt-appn}, and \ref{fig:ben-gmmlu-prompt-appn} provide the full prompt templates for English, German, Hindi, and Bengali, respectively.

\begin{figure}[!htb]
\centering
\begin{tcolorbox}[colback=gray!15, colframe=black, boxrule=1pt, width=0.49\textwidth, fontupper=\normalfont\small, boxsep=1pt, left=2pt, right=2pt, top=1pt, bottom=1pt, halign=justify] 
\noindent \textbf{System Prompt:} You are a helpful assistant trained to answer objective questions from <subject-category>. Each question comes with 4 options (A, B, C, and D). Provide your answer in the format of a single letter (A, B, C, or D) followed by an explanation in 20 words. Use the given examples to guide your answers. The examples do not have an explanation, but your response should have. 
\\ [0.3em]
\textbf{User Prompt:} Q1: <example-question-1> 
\\ [0.2em]
\textbf{Assistant Prompt:} Answer: <answer-example-question-1> 
\\ [0.3em]
\textbf{User Prompt:} Q2: <example-question-2> 
\\ [0.2em]
\textbf{Assistant Prompt:} Answer: <answer-example-question-2> 
\\ [0.3em]
\textbf{User Prompt:} Question: <actual-test-question-with-multiple-choices> 
\\ [0.3em]
\textbf{Assistant Prompt:} Answer:
\end{tcolorbox}
\caption{Prompt for \gmmlu in English.}
\label{fig:eng-gmmlu-prompt-appn}
\end{figure}

\begin{figure}[!htb]
\centering
\begin{tcolorbox}[colback=gray!15, colframe=black, boxrule=1pt, width=0.49\textwidth, fontupper=\normalfont\small, boxsep=1pt, left=2pt, right=2pt, top=1pt, bottom=1pt, halign=justify] 
\noindent \textbf{System Prompt:} Sie sind ein hilfreicher Assistent, der darauf trainiert ist, objektive Fragen von <subject-category> auf Deutsch zu beantworten. Jede Frage hat vier Antwortmöglichkeiten (A, B, C, und D). Geben Sie Ihre Antwort in Form eines einzelnen Buchstabens (A, B, C, oder D) an, gefolgt von einer Erklärung in 20 Wörtern. Orientieren Sie sich bei Ihren Antworten an den Beispielen. Die Beispiele enthalten keine Erklärungen, Ihre Antwort sollte jedoch welche enthalten.
\\ [0.3em]
\textbf{User Prompt:} Q1: <example-question-1> 
\\ [0.2em]
\textbf{Assistant Prompt:} Answer: <answer-example-question-1> 
\\ [0.3em]
\textbf{User Prompt:} Q2: <example-question-2> 
\\ [0.2em]
\textbf{Assistant Prompt:} Answer: <answer-example-question-2> 
\\ [0.3em]
\textbf{User Prompt:} Question: <actual-test-question-with-multiple-choices> 
\\ [0.3em]
\textbf{Assistant Prompt:} Answer:
\end{tcolorbox}
\caption{Prompt for \gmmlu in German.}
\label{fig:ger-gmmlu-prompt-appn}
\end{figure}

\begin{figure}[!htb]
\centering
\begin{tcolorbox}[colback=gray!15, colframe=black, boxrule=1pt, width=0.49\textwidth, fontupper=\normalfont\small, boxsep=1pt, left=2pt, right=2pt, top=1pt, bottom=1pt, halign=justify] 
\noindent \textbf{System Prompt:} \foreignlanguage{hindi}{आप} <subject-category> \foreignlanguage{hindi}{से हिन्दी में वस्तुनिष्ठ प्रश्नों का उत्तर देने के लिए प्रशिक्षित एक सहायक सहायक हैं. प्रत्येक प्रश्न में} 4 \foreignlanguage{hindi}{विकल्प} (A, B, C, \foreignlanguage{hindi}{और} D) \foreignlanguage{hindi}{दिए गए हैं। अपना उत्तर एक अक्षर प्रारूप में} (A, B, C, \foreignlanguage{hindi}{या} D) \foreignlanguage{hindi}{दर्ज करें और उसके बाद} 20 \foreignlanguage{hindi}{शब्दों में स्पष्टीकरण दें। अपने उत्तरों को निर्देशित करने के लिए दिए गए उदाहरणों का उपयोग करें। उदाहरणों में स्पष्टीकरण नहीं है, लेकिन आपके उत्तर में स्पष्टीकरण होना चाहिए।}
\\ [0.3em]
\textbf{User Prompt:} Q1: <example-question-1> 
\\ [0.2em]
\textbf{Assistant Prompt:} Answer: <answer-example-question-1> 
\\ [0.3em]
\textbf{User Prompt:} Q2: <example-question-2> 
\\ [0.2em]
\textbf{Assistant Prompt:} Answer: <answer-example-question-2> 
\\ [0.3em]
\textbf{User Prompt:} Question: <actual-test-question-with-multiple-choices> 
\\ [0.3em]
\textbf{Assistant Prompt:} Answer:
\end{tcolorbox}
\caption{Prompt for \gmmlu in Hindi.}
\label{fig:hin-gmmlu-prompt-appn}
\end{figure}

\begin{figure}[!htb]
\centering
\begin{tcolorbox}[colback=gray!15, colframe=black, boxrule=1pt, width=0.49\textwidth, fontupper=\normalfont\small, boxsep=1pt, left=2pt, right=2pt, top=1pt, bottom=1pt, halign=justify] 
\noindent \textbf{System Prompt:} \foreignlanguage{bengali}{আপনি} <subject-category> \foreignlanguage{bengali}{থেকে বাংলায় বস্তুনিষ্ঠ প্রশ্নের উত্তর দেওয়ার জন্য প্রশিক্ষিত একজন সহায়ক সহকারী। প্রতিটি প্রশ্নের জন্য ৪টি বিকল্প} (A, B, C, \foreignlanguage{bengali}{এবং} D) \foreignlanguage{bengali}{দেওয়া আছে। আপনার উত্তর একটি অক্ষরের} (A, B, C, \foreignlanguage{bengali}{অথবা} D) \foreignlanguage{bengali}{আকারে দিন এবং তারপর ২০ শব্দে ব্যাখ্যা দিন। আপনার উত্তরের দিকনির্দেশনা দিতে প্রদত্ত উদাহরণগুলি ব্যবহার করুন। উদাহরণগুলিতে ব্যাখ্যা নেই কিন্তু আপনার উত্তরে ব্যাখ্যা থাকতে হবে।}
\\ [0.3em]
\textbf{User Prompt:} Q1: <example-question-1> 
\\ [0.2em]
\textbf{Assistant Prompt:} Answer: <answer-example-question-1> 
\\ [0.3em]
\textbf{User Prompt:} Q2: <example-question-2> 
\\ [0.2em]
\textbf{Assistant Prompt:} Answer: <answer-example-question-2> 
\\ [0.3em]
\textbf{User Prompt:} Question: <actual-test-question-with-multiple-choices> 
\\ [0.3em]
\textbf{Assistant Prompt:} Answer:
\end{tcolorbox}
\caption{Prompt for \gmmlu in Bengali.}
\label{fig:ben-gmmlu-prompt-appn}
\end{figure}

\section{Human Evaluation of the  Heuristic for Identifying Hallucinations}
\label{sec:hallu-heuristic-eval}

In this study, we adopt a substring-based heuristic to identify factual hallucinations in LLM responses, which has been applied in several prior works in the generative question answering task~\citep{snyder_early_2024, liang2023holistic, adlakha-etal-2024-evaluating}. 
According to this heuristic, an LLM-generated response $A$ is marked correct if the reference answer $R$ (for the same question) is a substring of $A$; otherwise, $A$ is considered a hallucination. All comparisons are performed in lowercase.

To quantify the reliability of this heuristic, we conducted a detailed human evaluation. 
We manually evaluated responses from the 
\llamaM on 50 samples from each mTREx category across all five languages: \en, \de, \hi, \bn, and \ur (750 samples in total). A human annotator was shown the question, the LLM-generated response $A$, the reference answer $R$, and then asked to judge if $A$ is correct or a hallucination. The human judgment was matched with the decision according to the above-mentioned heuristic. 

Results (percentage of cases where the human judgement matched with the decision according to the heuristic) across all languages are presented in Table~\ref{tab:heuristic-manual-eval}. The results show that the heuristic aligns closely with human judgment: \textbf{for \en, it is correct for more than 95\% of cases on average}, with errors typically arising in scenarios where the gold answer is part of the question itself, as shown in Figure~\ref{fig:heuristic-eval-examples-en}.

\noindent \textbf{For non-\en languages, the heuristic achieves more than 88\% accuracy on average}.  
As shown in Figure~\ref{fig:heuristic-eval-examples-non-en}, most of the errors in these non-\en languages are False Positives (the LLM is actually correct, but the heuristic marks the answer as wrong), primarily due to typographical or script variations in how the same entity name can be written in a particular language. Overall, this analysis \textbf{demonstrates high agreement between the substring-match heuristic and human evaluation}. 

\begin{figure}[!htb]
\centering
\begin{tcolorbox}[colback=gray!15, colframe=black, boxrule=1pt, fontupper=\normalfont\small, boxsep=1pt, left=2pt, right=2pt, top=1pt, bottom=1pt, halign=justify, enhanced]
\textbf{Question (\en):} What is the capital of Houghton County? 
\\
\textbf{Ref. Ans. (\en):} Houghton, MI <OR> Houghton <OR> Houghton, Michigan
\\
\textbf{Model Res.:} \textcolor{Red}{Houghton County is a county in the U.S. state of Michigan, and Hancock is its county seat.}
\\
\textbf{Heuristic Evaluation:} Correct (Not a hallucination)
\\
\textbf{Human Evaluation:} Wrong (Hallucination)
\end{tcolorbox}
\vspace{-1em}
\caption{Example in English (\en) language where hallucination heuristic evaluation is incorrect. Here, the LLM response is wrong, but the heuristic marks the response as correct.}
\label{fig:heuristic-eval-examples-en}
\end{figure}

\begin{table}[!h]
\centering
\scriptsize
\begin{tabular}{|c|c|c|c|}
\hline
\multirow{2}{*}{\bf Lang.} & \multicolumn{3}{c|} {\bf \% of correct detection using heuristic} \\
\cline{2-4}
& {\bf Capitals} & {\bf Country} & {\bf Official Language} \\
\hline
\en & 90 & 96 & 100 \\
\de & 80 & 90 & 94 \\
\hi & 80 & 96 & 92 \\
\bn & 82 & 96 & 78 \\
\ur & 80 & 94 & 96 \\
\hline
\end{tabular}
\caption{Assessment of substring-based heuristic for hallucination detection.}
\label{tab:heuristic-manual-eval}
\end{table}

\begin{figure}[t]
\centering
\begin{tcolorbox}[colback=gray!15, colframe=black, boxrule=1pt, fontupper=\normalfont\small, boxsep=1pt, left=2pt, right=2pt, top=1pt, bottom=1pt, halign=justify, enhanced]
\textbf{Question (\de):} Was ist die Hauptstadt von Schaki-Khanat?
\\
\textbf{Ref. Ans. (\de):} Schach <OR> Scheqi, Aserbaidschan <OR> Scheki
\\
\textbf{Model Res.:} \textcolor{ForestGreen}{Nukha (heute Şəki)}
\\
\textbf{Heuristic Evaluation:} Wrong (Hallucination) 
\\
\textbf{Human Evaluation:} Correct (Not a hallucination)
\tcbline
\textbf{Question (\hi):} {\fontsize{8}{8}\selectfont \foreignlanguage{hindi}{फेयट काउंटी, इलिनॉयस की राजधानी क्या है?}}
\\
\textbf{Ref. Ans. (\hi):} {\fontsize{8}{8}\selectfont \foreignlanguage{hindi}{वंडालिया}}
\\
\textbf{Model Res.:} {\fontsize{8}{8}\selectfont \textcolor{ForestGreen}{\foreignlanguage{hindi}{फेयट काउंटी, इलिनॉयस की राजधानी वांडलिया है।}}}
\\
\textbf{Heuristic Evaluation:} Wrong (Hallucination) 
\\
\textbf{Human Evaluation:} Correct (Not a hallucination)
\tcbline
\textbf{Question (\bn):} {\fontsize{8}{8}\selectfont \foreignlanguage{bengali}{সেন্ট জর্জ প্যারিশ, সেন্ট ভিনসেন্ট এবং গ্রেনাডাইনস এর রাজধানী কি?}}
\\
\textbf{Ref. Ans. (\bn):} {\fontsize{8}{8}\selectfont \foreignlanguage{bengali}{কিংস্টাউন}}
\\
\textbf{Model Res.:} {\fontsize{8}{8}\selectfont \textcolor{ForestGreen}{\foreignlanguage{bengali}{কিংসটাউন}}}
\\
\textbf{Heuristic Evaluation:} Wrong (Hallucination) 
\\
\textbf{Human Evaluation:} Correct (Not a hallucination)
\tcbline
\textbf{Question (\ur):} \foreignlanguage{urdu}{ماناواتو کا دارالحکومت کیا ہے؟}
\\
\textbf{Ref. Ans. (\ur):} \foreignlanguage{urdu}{پالمسٹون نارتھ}
\\
\textbf{Model Ans.:} \textcolor{ForestGreen}{\foreignlanguage{urdu}{ماناواتو کا دارالحکومت پالمرسٹن نارتھ ہے۔}}
\\
\textbf{Heuristic Evaluation:} Wrong (Hallucination) 
\\
\textbf{Human Evaluation:} Correct (Not a hallucination)
\end{tcolorbox}
\vspace{-1em}
\caption{Examples in non-\en languages, where hallucination heuristic evaluation is incorrect. Here, the LLM responses are actually correct, but the heuristic marks the responses as wrong.}
\label{fig:heuristic-eval-examples-non-en}
\end{figure}

\section{Hallucination Detection Methods}
\label{sec:hallu-appn}

In our study, we focus on three Hallucination Detection (HD) methods that represent popular families of HD methods: (i) \textit{methods utilizing model internal artifacts during generation}, (ii) \textit{sampling-based black-box detectors that leverage only responses generated by the model}.

\subsection{Model Artifacts Method (\mam)}

Inspired by the work of \citet{snyder_early_2024}, \textit{we investigate whether the model's internal artifacts associated with the generation can provide signals on hallucination across languages}.
This method is model-agnostic, enabling the probing of factual reliability directly from the model's internal states. 

\xhdr{Artifacts for Detecting Hallucinations}
We focus on the following key artifacts from the final decoder layer (say $L$) of the Transformer architecture, specifically after processing the input prompt and generating the \textit{first token} of the response. 

\begin{itemize}[leftmargin=*, nolistsep]
    \item \textbf{Self-Attention Scores ($\mathbf{S}_\ell$):} These represent the outputs of the final linear projection within the multi-head self-attention module at layer $\ell~\in \{1, \ldots, L\}$, encoding contextual dependencies among tokens. We extract the self-attention scores for the final decoder layer $L$ associated with the interaction between the final token of the input prompt (say, $I_N$ for $N$ tokens in input) and the first generated response token (say, $O_1$). This is denoted as $\mathbf{S}_L(I_N, O_1)$.
    \item \textbf{Fully connected activations ($\mathbf{FC}_\ell$):} These are the fully connected hidden representation at layer $\ell$. We similarly extract $\mathbf{FC}_L(I_N, O_1)$ to represent the activations linking the final input token $I_N$ and the first answer token $O_1$.
\end{itemize}



\xhdr{Classifier for hallucination detection} After extracting these features, we train a single-layer neural network with a hidden dimension of 256 to classify whether a response was factually correct or hallucinated. For each dataset category, we train and evaluate the classifier on a random 80/20 split. 

\subsection{SelfCheckGPT Method (\sgm)}

Along with analyzing models' internal artifacts, we have also benchmarked sampling-based blackbox methods such as \textbf{SelfCheckGPT} method (\sgm)~\citep{manakul-etal-2023-selfcheckgpt}. 
It leverages the LLM itself to determine the factuality of a generated response by measuring the factual consistency across multiple responses sampled for the same question. 
The intuition is that factual responses tend to remain consistent across multiple stochastic generations, whereas hallucinated responses are likely to diverge and contradict one another. 
We adopt the n-gram based variant of \sgm to assess lexical consistency across sampled responses. The key steps are outlined below:

\xhit{(i) Stochastic Sampling} For a given input, a set of diverse responses has been generated with high temperature to encourage variability in the outputs. In contrast, the base response is generated by setting the temperature to zero to obtain the deterministic output from the model.

\xhit{(ii) Consistency Scoring} To evaluate how much the base response is varied with respect to the distribution defined by the sampled responses, the average negative log-likelihood (NLL) of the base response has been calculated. Which is then transformed into the final SelfCheck n-gram score. The specific transformation used in our implementation, following common practice, is $1.0 - \exp(-avg_{NLL})$. This maps the NLL (which ranges from 0 to $\infty$) to a score between 0 and 1, where higher scores correspond to lower NLL and thus higher consistency.

Finally, for each question, the computed SelfCheck n-gram score serves as the detector's output, predicting the likelihood of the base response being factual or hallucinated. 

\subsection{Semantic Entropy Method (\sem)}

In contrast to the previous n-gram-based lexical approach, we evaluate a semantic approach, which is also a sampling-based method, called \textbf{Semantic Entropy} (\sem) \citep{semantic_entropy_paper}. It assesses hallucination by estimating \textit{semantic uncertainty} across sampled responses generated by the model.
The key idea is that \textit{when an LLM is uncertain, it is more likely to generate responses that diverge significantly in their semantic content}, leading to higher entropy. 
Conversely, when a model is confident, it generates semantically consistent responses. This diversity is quantified as \textit{semantic entropy}. 
The confident and factual responses are expected to cluster around a single or a few closely related meanings, resulting in lower \textit{semantic entropy}. The method proceeds in the following steps:

\xhit{(i) Stochastic Sampling} For a given question, sample responses have been generated with high temperature to encourage output diversity. 
For each response, the sequence-level log-likelihood has been computed by averaging the token-level log-probabilities. 
The base response has been generated at zero temperature to serve as the reference for determining correctness.

\xhit{(ii) Semantic Clustering} To assess the semantic equivalence of sampled responses, we utilize the language-agnostic \textit{LaBSE} model~\citep{labse_paper}, which encodes each response into a fixed-dimensional semantic embedding. 
Next, the sampled responses are grouped into semantic clusters if the mutual cosine similarity exceeds a threshold ($\tau = 0.75$)\footnote{The threshold has been chosen based on the internal evaluation by the native speakers.}. This enables the clustering of responses that convey the same underlying meaning, even if phrased differently.

\xhit{(iii) Cluster Probability and Entropy} The probability of each cluster $C_k$ is computed by summing the probabilities of the individual responses that belong to it. If $C_k$ contains $I_k$ responses, then the aggregated probability is given by:
\begin{equation*}
P(C_k) = \sum_{i \in I_k} P(\text{response}_i)
\end{equation*}
The \textit{semantic entropy} is then computed over the distribution of $P(C_k)$.
Finally, the entropy values are compared against binary ground-truth labels. 

\section{Additional Results}
\label{sec:appn-results}

\subsection{Task Accuracy Details}
\label{subsec:appn-task-acc}

Table~\ref{table:acc} presents the task accuracies of \mistralS, \llamaS, \mistralM, and \llamaM across all datasets in answering factual questions for all five languages: English (\en), German (\de), Hindi (\hi), Bengali (\bn), and Urdu (\ur).

\input{final_tables/task_acc_appn}

\subsection{Results of Hallucination Detection Methods}
\label{subsec:appn-hallu-methods}

Tables~\ref{table:snyder_attention}, \ref{table:snyder_fullyconnected}, \ref{table:selfcheckgpt}, and \ref{table:semanticentropy} present exact AUROC scores of \textbf{\mam \texttt{(self-attention)}}, \textbf{\mam \texttt{(fully connected activations)}}, \textbf{\sgm}, and \textbf{\sem} methods for \mistralS, \llamaS, \mistralM, and \llamaM across datasets and languages, respectively. Tables~\ref{table:snyder_attention_delta_percentage}, \ref{table:selfcheckgpt_delta_percentage}, and \ref{table:semanticentropy_delta_percentage} present the percentage of increment or decrement in AUROC scores for \mistralS, \llamaS, \mistralM, and \llamaM across languages for \textbf{\mam \texttt{(self-attention)}}, \textbf{\sgm}, and \textbf{\sem} methods w.r.t. the corresponding \en baseline for the model and dataset, respectively. 

\xhdr{Ablation Studies for Validating Robustness of Hallucination Detectors} 
\new{To further validate the robustness of hallucination detectors (HDs), we conducted the following detailed ablation studies covering both families of detector methods:}
\begin{itemize}[leftmargin=*, nolistsep]
    \item \new{For \textbf{\mam detectors} (both \textbf{\texttt{self-attention}} and \textbf{\texttt{fully connected activations}} variants by considering artifacts from the final layer at the first generated token), we trained the classifiers 20 times in the same language setup using different random seeds to create diverse train-test splits. Across all four LLMs and five languages, we \textbf{observe that the standard errors of AUROC scores across runs remain very small (only around 0.02)}, demonstrating stable detector performance. Detailed results for \mam (\texttt{self-attention} and \texttt{fully connected activations}) variants are reported in Tables~\ref{tab:self-attn-ablation} and \ref{tab:fc-ablation}, respectively.}

    \item Additionally, we further evaluate \textbf{\mam detectors} (both \textbf{\texttt{self-attention}} and \textbf{\texttt{fully connected activations}} by averaging artifacts across multiple generated tokens (up to the first 10 tokens) from the final layer. Results for the \mam (\texttt{self-attention} and \texttt{fully connected activations}) variants are reported in Tables~\ref{table:mam_attention_avg_tkns} and \ref{table:mam_fc_activation_avg_tkns}, respectively. We observe that, \textit{on average, the performance remains largely similar to the previous setup, where only the `first generated token' was considered}. For instance, \mam (\texttt{self-attention}) over the Capitals dataset in \de with the \mistralM model yields 77 vs. 76 AUROC (see Tables~\ref{table:mam_attention_avg_tkns} and \ref{table:snyder_attention}). Also, in several cases, performance slightly improves upon considering multiple generated tokens, e.g., for \mam (\texttt{self-attention}) over the Capitals in \hi with \llamaS: 72 vs. 68 AUROC (see Tables~\ref{table:mam_attention_avg_tkns} and \ref{table:snyder_attention}).

    Furthermore, as reported in Tables~\ref{table:mam_attention_avg_tkns_cross_lingual}--\ref{table:mam_fc_activation_avg_tkns_multi_lingual_delta_percentage}, we have also conducted the cross-lingual and multilingual analyses (discussed in Appendix~\ref{subsec:appn-cross-multi-mam}) with this new `multiple generated token' variant of \mam detectors, and it exhibits the same qualitative trends as the `first generated token' setup: (i) EN-to-target cross-lingual transfer remains challenging, and (ii) multilingual training mitigates this gap. Tables~\ref{table:mam_attention_avg_tkns_cross_lingual} \& \ref{table:mam_fc_activation_avg_tkns_cross_lingual} present exact AUROC scores, and Tables~\ref{table:mam_attention_avg_tkns_cross_lingual_delta_percentage} \& \ref{table:mam_fc_activation_avg_tkns_cross_lingual_delta_percentage} present the percentage of increment or decrement in AUROC w.r.t. the \en baseline in the cross-lingual setting. For the multilingual setting, Tables~\ref{table:mam_attention_avg_tkns_multi_lingual} \& \ref{table:mam_fc_activation_avg_tkns_multi_lingual} show exact AUROC scores, and Tables~\ref{table:mam_attention_avg_tkns_multi_lingual_delta_percentage} \& \ref{table:mam_fc_activation_avg_tkns_multi_lingual_delta_percentage} show the percentage of increment or decrement in AUROC w.r.t. the \en baseline.

    \item For \textbf{sampling-based detectors} -- Semantic Entropy (\sem) and SelfCheckGPT (\sgm) -- we performed ablation experiments by varying the number of sampled responses $N \in \{10, 15, 20\}$ at high temperature settings. Experiments were conducted across all five datasets for \mistralS, \llamaS, and \mistralM models in all five languages. \textbf{Results show that absolute AUROC values change only minimally as $N$ varies, and the standard errors in AUROC remain extremely small (less than 0.01) across all models and languages}. This further strengthens the stability of our findings. Detailed results for \sgm and \sem detectors are reported in Tables~\ref{tab:sgm-ablation} and \ref{tab:sem-ablation}, respectively.
\end{itemize}

\subsection{Entropy Analyses Between Generators and Detectors}
\label{subsec:appn-entropy-analyses}

To further investigate the stability of detectors' performance across languages compared to the performance of generators (i.e., LLMs' task accuracy), we analyze the \textit{entropy of the softmax distributions} produced by both LLM generators and hallucination detectors across high- and low-resource languages.

It is true that inefficient tokenization of LLMs for low-resource languages primarily affects the generation task at each step. More specifically, low-resource languages usually get tokenized into a larger number of tokens than high-resource languages, and these tokens are observed far less frequently during pretraining due to the low-resource nature of the language. Consequently, \textit{at each generation step, the model must select from a larger and less familiar token set, leading to a more spread out softmax distribution in low-resource languages}, that is, the token probabilities are closer to each other. In contrast, high-resource languages tend to exhibit more concentrated softmax distributions.

On the other hand, the hallucination detectors have to pick between the same set of options (exactly two: hallucinated vs. non-hallucinated) regardless of the language, and are trained on the same amount of data, potentially leading to more comparable softmax distributions across languages.

To quantify this phenomenon, we compute the entropy (a measure of concentration, or lack thereof) of the softmax distribution for the first generated token for all LLMs across all five languages. As reported in Table~\ref{table:generator_entropy}, generator entropies are consistently low for high-resource languages but substantially higher for low-resource ones, indicating greater uncertainty during generation in low-resource settings. For instance, on the \texttt{Capitals} dataset for \mistralM: 0.325 for \en vs. 1.762 for \ur.

In contrast, the detector entropies (as reported in Tables~\ref{table:mam_attention_entropy} \& \ref{table:mam_fc_entropy} for \mam \texttt{(self-attention)} and \mam \texttt{(fully connected activations)}, respectively) remain consistent across languages, for both high- and low-resource languages. For instance, \mam \texttt{(fully connected activations)} on \texttt{Country} dataset with \mistralM: 0.3 for \en vs. 0.336 for \ur. \textit{These results demonstrate that generator entropies vary significantly between high- and low-resource languages, whereas detector entropies do not}, as they operate on a fundamentally simpler prediction space (hallucination vs. non-hallucination).

\subsection{Cross-lingual and Multilingual Analyses for \mam Detectors}
\label{subsec:appn-cross-multi-mam}
\new{
We conduct additional experiments to analyze the behavior of \mam detectors in cross-lingual and multilingual settings. These experiments complement the same-language setup reported in the main paper (where training and testing had been done over the same language data) and provide deeper insights into the cross-lingual transfer effects. 
}

\xhdr{Cross-lingual Setting} \new{
In the cross-lingual setting, we train \mam detectors on 80\% of the \en data and test them on 20\% of the target-language data, across all dataset categories and for all four LLMs. \textbf{We observe significant performance drops in this cross-lingual setting compared to the same-language setup.} For instance, on \texttt{mTREx-Capitals} using \textbf{\mam \texttt{(self-attention)}}, average AUROC across all LLMs decreases from 74 $\rightarrow$ 62 in \de, 70 $\rightarrow$ 54 in \hi, 71 $\rightarrow$ 56 in \bn, and 70 $\rightarrow$ 52 in \ur. Similar drops are observed across other datasets and both \mam variants (\texttt{self-attention} and \texttt{fully connected activations}). Detailed results across all datasets are reported in Tables~\ref{table:mam-self-attn-cross-lingual}, \ref{table:mam-self-attn-cross-lingual_delta_percentage}, \ref{table:mam-fc-cross-lingual}, and \ref{table:mam-fc-cross-lingual_delta_percentage}, where Tables~\ref{table:mam-self-attn-cross-lingual} and \ref{table:mam-fc-cross-lingual} present exact AUROC scores and Tables~\ref{table:mam-self-attn-cross-lingual_delta_percentage} and \ref{table:mam-fc-cross-lingual_delta_percentage} present the percentage of increment or decrement in AUROC w.r.t. the \en baseline.
}

\xhdr{Multilingual Setting} \new{
In this setup, \mam detectors had been trained on 80\% of the data across all five languages (EN, DE, HI, BN, UR) and tested on 20\% of each language separately. \textbf{Unlike the cross-lingual setting, here the performance remains close to the same-language setup, showing that multilingual supervision mitigates the cross-lingual transfer gap.} For example, on \texttt{mTREx-Capitals} with \textbf{\mam \texttt{(self-attention)}}, average AUROC across all LLMs is nearly unchanged: 74 vs. 72 in \de, 70 vs. 68 in \hi, 71 vs. 71 in \bn, and 70 vs. 67 in \ur. Detailed results across all datasets and LLMs are provided in Tables~\ref{table:mam-self-attn-multi-lingual}, \ref{table:mam-self-attn-multi-lingual_delta_percentage}, \ref{table:mam-fc-multi-lingual}, and \ref{table:mam-fc-multi-lingual_delta_percentage}, where Tables~\ref{table:mam-self-attn-multi-lingual} and \ref{table:mam-fc-multi-lingual} present exact AUROC scores and Tables~\ref{table:mam-self-attn-multi-lingual_delta_percentage} and \ref{table:mam-fc-multi-lingual_delta_percentage} present the percentage of increment or decrement in AUROC w.r.t. the \en baseline.
}

These experiments together show two key findings: (i) \textbf{zero-shot \en-to-target cross-lingual transfer for hallucination detection is challenging}, and (ii) \textbf{combined multilingual training mitigates this gap}. A brief theoretical intuition helps explain these effects. Consider the very simple setting where the last layer last token embeddings for each class (correct vs. hallucinated) follow a 1D Gaussian distribution: In EN, hallucinated responses follow $N(-5,1)$, that is, a mean of $5$ and a variance of $1$, and non-hallucinated responses follow $N(5,1)$. Clearly, both can be separated with almost 100\% accuracy. Now, consider another language, say \ur, where hallucinated responses follow $N(10,1)$ and non-hallucinated responses $N(-10,1)$. These are also well separated, but the \en-trained threshold would give almost zero accuracy in the case of \ur. The key point is that while the correct and hallucinated responses in both languages are almost perfectly separable, they are not separable via the same classifier and \textbf{need a language-specific classifier}. Hence, zero-shot cross-lingual transfer struggles, but modest in-language supervision or pooled multilingual training restores performance. 


\subsection{Accuracy of Classifiers for \mam method}
\label{subsec:appn-cls-acc}

Tables \ref{table:appendix_snyder_attention_delta} and \ref{table:appendix_snyder_fullyconnected_delta} present the binary classification accuracy of the Model Artifacts method across all four LLMs, covering all the dataset categories in mTREx and G-MMLU datasets and all languages.

\subsection{Interpretation of $10^{\metric}$ Values}
\label{appendix:tphr_magnitude}

\new{\textit{To illustrate the magnitude of the differences} in the model's task accuracy compared to the differences in HD's performance across languages with respect to the \en baseline, we also present the values of $10^{\metric}$ in Table~\ref{table:ratio_acc_auc_tphr_original}. 
For a given language $L$, $10^{\metric(L)}$ directly represents the \textit{ratio of disparities}, that is, how many times larger the change in task accuracy is compared to the change in HD's performance. Here,}
\new{
\begin{itemize}[leftmargin=*, nolistsep]
\item $10^{\metric}$ $\approx 1$ indicates the disparities in task accuracy and HD's performance are of equal magnitude relative to the \en baseline. In other words, both the task performance and the HD performance change equally with respect to the \en baseline.
\item $10^{\metric}$ $\gg 1$ means the difference in task accuracy is much higher than the difference in HD's performance.
\item $10^{\metric}$ $\approx 0$ represents the difference in HD's performance is significantly greater than the difference in LLM's task accuracy.
\end{itemize}
}

\new{As reported in the Table~\ref{table:ratio_acc_auc_tphr_original}, we observe for low-resource languages like \bn and \ur, the $10^{\metric}$ values are very high, even many times in the range of 30 to 50, particularly for the mTREX datasets.}

\input{final_tables/hd_auroc_appn}

\input{final_tables/hd_auroc_percentage_appn}

\input{final_tables/ablation_exp_tables}

\input{final_tables/entropies_tables}

\input{final_tables/cross_lingual_tables}

\input{final_tables/multi_lingual_tables}


\input{tables/appendix_snyder_attention_final}

\input{tables/appendix_snyder_fully_connected_final}

\input{final_tables/tphr_tables_main}

\subsection{Comparison of Task Accuracy with Tokenization capability}
\label{subsec:appn-token-plots}

\begin{figure*}[!htb]
\centering
    \subfloat[For mTREx -- Capitals]{\includegraphics[width=0.33\linewidth, height=5.5cm]{figures/mTREx_accuracy_vs_compression_per_category/capitals_mTREx_accuracy_vs_token_compression.png}}\hfill
    \subfloat[For mTREx -- Country]{\includegraphics[width=0.33\linewidth, height=5.5cm]{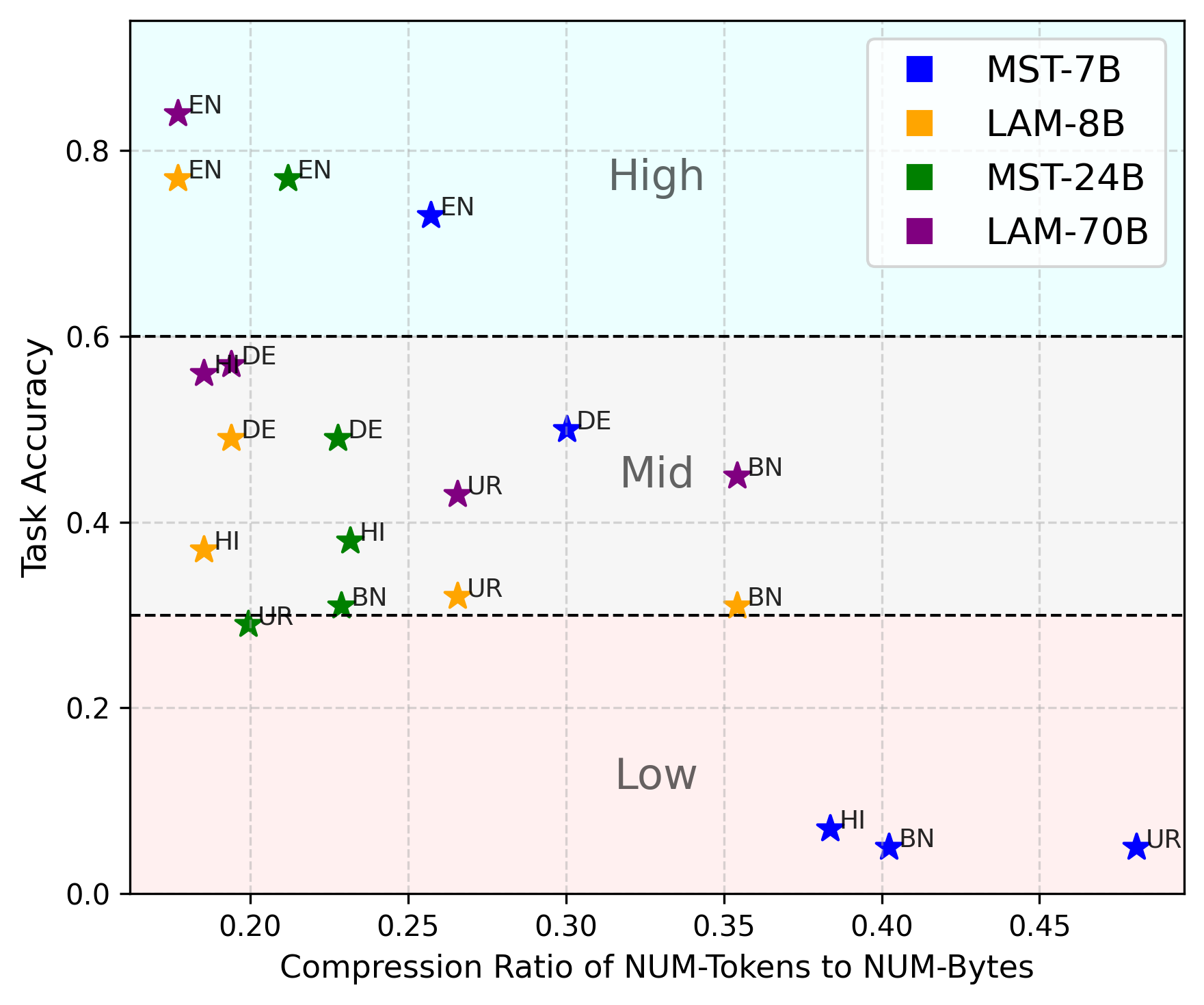}}\hfill
    \subfloat[For mTREx -- Official Languages]{\includegraphics[width=0.33\linewidth, height=5.5cm]{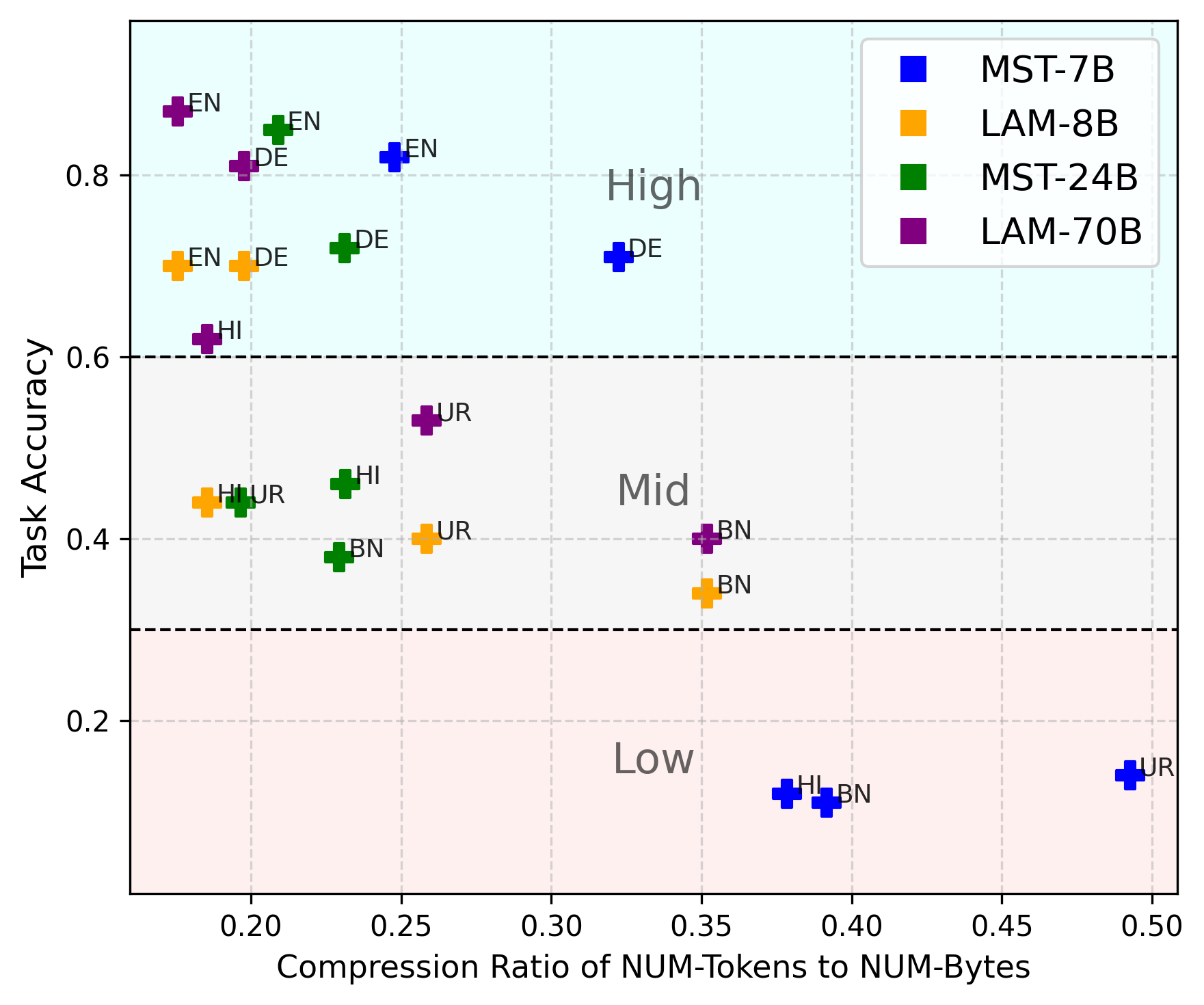}}
\caption{Comparison of task accuracy with token compression ratio (number of tokens produced by the model's tokenizer divided by the number of bytes in the input prompt) for mTREx across all LLMs and languages.}
\label{fig:acc_vs_token_comp_mtrex}
\end{figure*}

\begin{figure*}[!htb]
\centering
    \subfloat[For G-MMLU -- STEM]{\includegraphics[width=0.40\linewidth, height=6cm]{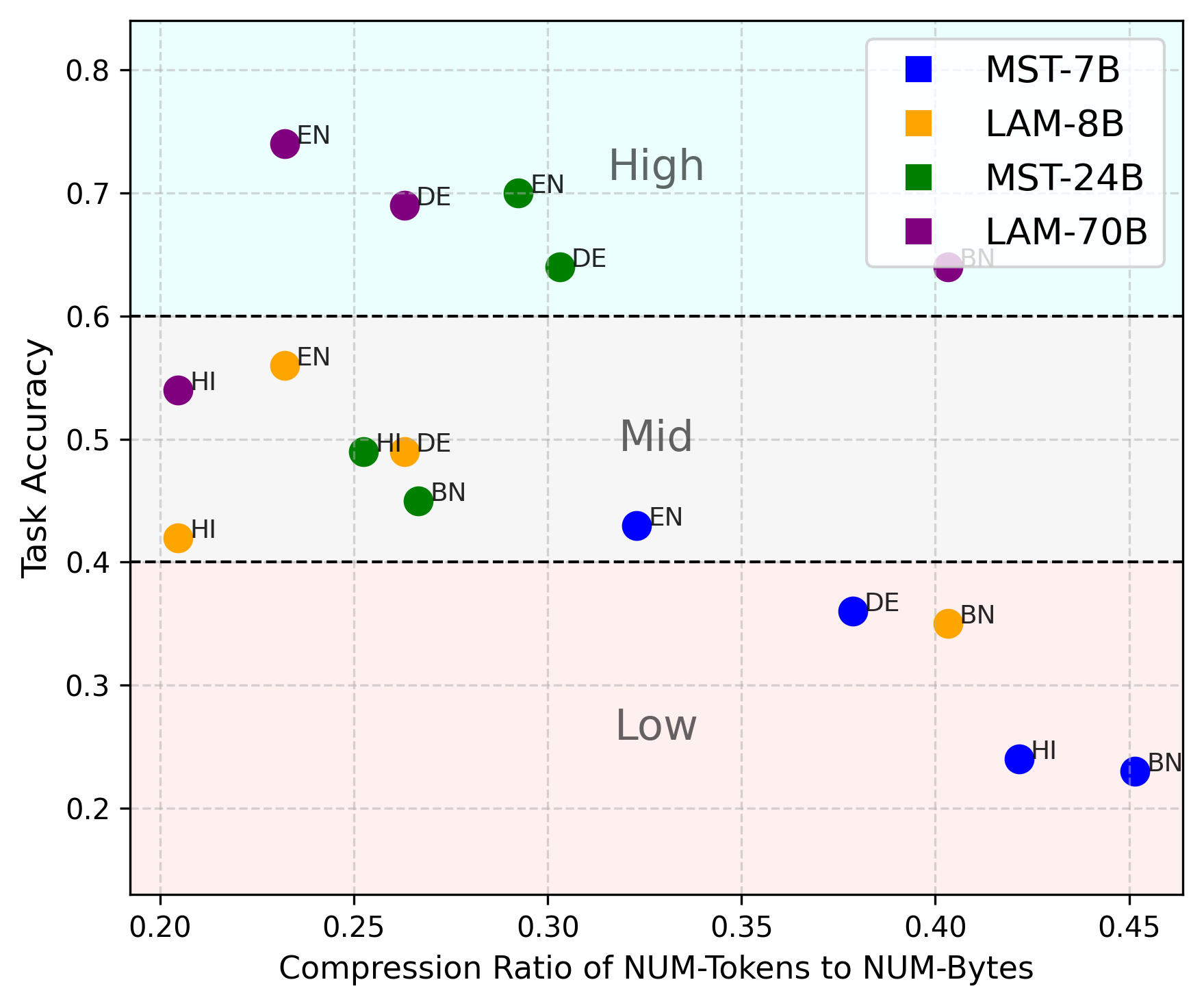}}
    \hspace{3mm}
    \subfloat[For G-MMLU -- Humanities]{\includegraphics[width=0.40\linewidth, height=6cm]{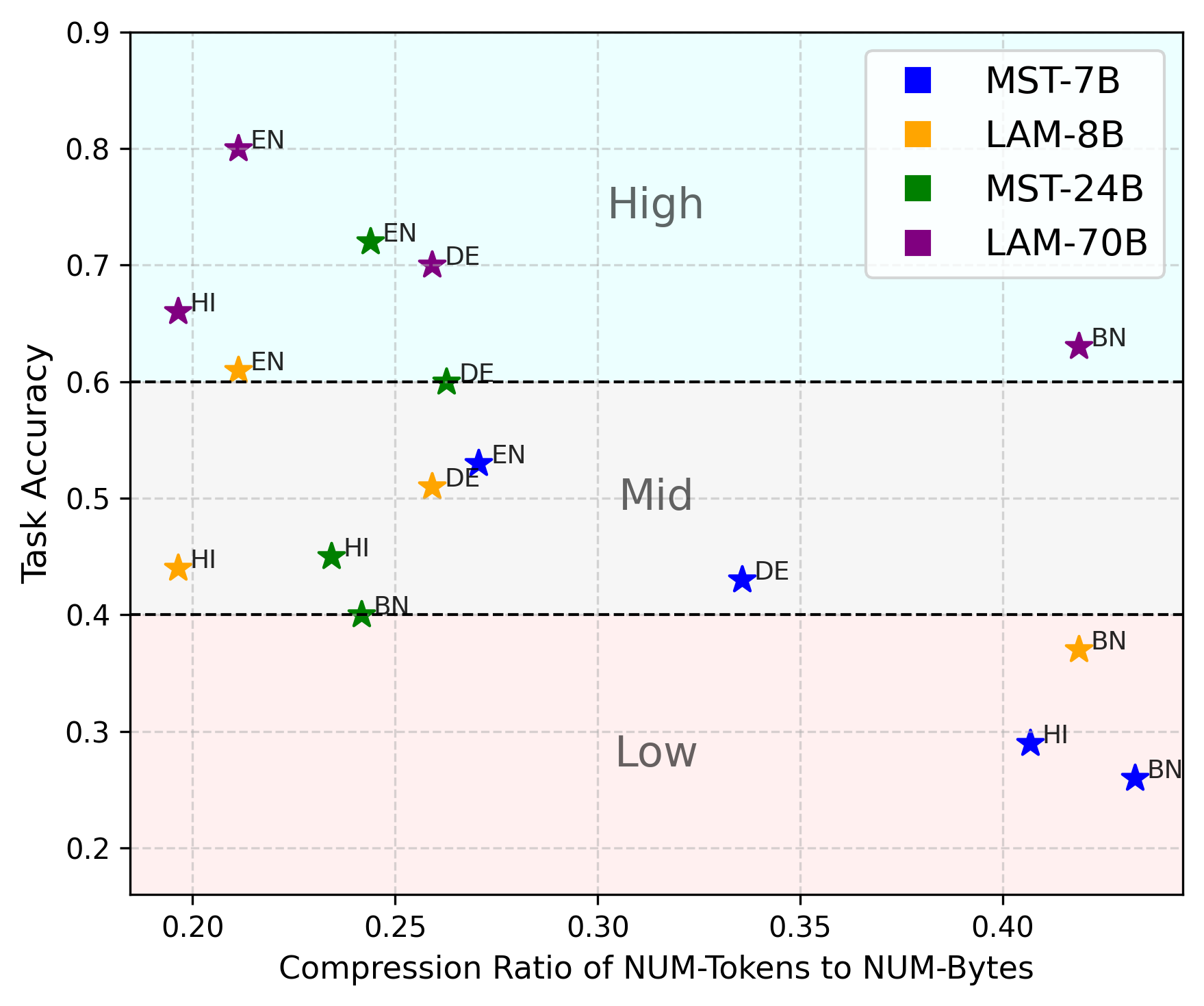}}
\caption{Comparison of task accuracy with token compression ratio (number of tokens produced by the model's tokenizer divided by the number of bytes in the input prompt) for G-MMLU across all LLMs and languages.}
\label{fig:acc_vs_token_comp_gmmlu}
\end{figure*}

To gain deeper insights into the language capabilities of LLMs, we examine the relationship between tokenization efficiency and task performance. Specifically, we analyze the token compression ratio, defined as the number of tokens produced by the model's tokenizer divided by the number of bytes in the input string. Figures~\ref{fig:acc_vs_token_comp_mtrex}, \ref{fig:acc_vs_token_comp_gmmlu} illustrate how this ratio correlates with task accuracy across all LLMs and languages for mTREx and G-MMLU, respectively.

%% file: final_tables/task_acc_appn.tex
\begin{table*}[!htb]
\centering
\small
\resizebox{\textwidth}{!}{
\begin{tabular}{l|ccccc|ccccc|ccccc|cccc|cccc|}
\toprule
\multirow{2}{*}{\bf Models} & \multicolumn{5}{|c|}{\bf mTREx -- Capitals} & \multicolumn{5}{|c|}{\bf mTREx -- Country} & \multicolumn{5}{|c|}{\bf mTREx -- Official Language} & \multicolumn{4}{|c|}{\bf G-MMLU -- STEM} & \multicolumn{4}{|c|}{\bf G-MMLU -- Humanities} 
\\
 & \en & \de & \hi & \bn & \ur
 & \en & \de & \hi & \bn & \ur
 & \en & \de & \hi & \bn & \ur
 & \en & \de & \hi & \bn 
 & \en & \de & \hi & \bn 
\\
\midrule
\mistralS
 & 69 & 57 & 20 & 19 & 14
 & 73 & 50 & 07 & 05 & 05
 & 82 & 71 & 12 & 11 & 14
 & 43 & 36 & 24 & 23
 & 53 & 43 & 29 & 26
\\
\llamaS
 & 74 & 66 & 32 & 28 & 28
 & 77 & 49 & 37 & 31 & 32
 & 70 & 70 & 44 & 34 & 40
 & 56 & 49 & 42 & 35
 & 61 & 51 & 44 & 37
\\
\mistralM
 & 80 & 60 & 30 & 29 & 29
 & 77 & 49 & 38 & 31 & 29
 & 85 & 72 & 46 & 38 & 44
 & 70 & 64 & 49 & 45
 & 72 & 60 & 45 & 40
\\
\llamaM
 & 86 & 74 & 44 & 37 & 31
 & 84 & 57 & 56 & 45 & 43
 & 87 & 81 & 62 & 40 & 53
 & 74 & 69 & 54 & 64
 & 80 & 70 & 66 & 63
\\
\midrule
\avg
 & 77 & 64 & 32 & 28 & 26
 & 78 & 51 & 34 & 28 & 27
 & 81 & 74 & 41 & 31 & 38
 & 61 & 54 & 42 & 42
 & 66 & 56 & 46 & 42
\\
\bottomrule
\end{tabular}
}
\caption{\small Task accuracy of \mistralS, \llamaS, \mistralM, and \llamaM across datasets in answering factual questions in English (\en), German (\de), Hindi (\hi), Bengali (\bn), and Urdu (\ur) languages.}
\label{table:acc}
\end{table*}

%% file: final_tables/hd_auroc_appn.tex
\begin{table*}[!htb]
\centering
\small
\resizebox{\textwidth}{!}{
\begin{tabular}{l|ccccc|ccccc|ccccc|cccc|cccc|}
\toprule
\multirow{2}{*}{\bf Models} & \multicolumn{5}{|c|}{\bf mTREx -- Capitals} & \multicolumn{5}{|c|}{\bf mTREx -- Country} & \multicolumn{5}{|c|}{\bf mTREx -- Official Language} & \multicolumn{4}{|c|}{\bf G-MMLU -- STEM} & \multicolumn{4}{|c|}{\bf G-MMLU -- Humanities} 
\\
 & \en & \de & \hi & \bn & \ur
 & \en & \de & \hi & \bn & \ur
 & \en & \de & \hi & \bn & \ur
 & \en & \de & \hi & \bn 
 & \en & \de & \hi & \bn 
\\
\midrule
\mistralS
 & 79 & 70 & 74 & 74 & 71
 & 77 & 81 & 73 & 84 & 72
 & 78 & 83 & 80 & 74 & 82
 & 73 & 66 & 54 & 51
 & 72 & 66 & 55 & 52
\\
\llamaS
 & 79 & 74 & 68 & 69 & 68
 & 86 & 87 & 77 & 83 & 78
 & 88 & 84 & 83 & 81 & 85
 & 77 & 74 & 66 & 66
 & 78 & 73 & 65 & 64
\\
\mistralM
 & 78 & 76 & 71 & 73 & 70
 & 83 & 93 & 80 & 85 & 82
 & 86 & 89 & 84 & 85 & 86
 & 74 & 72 & 72 & 62
 & 70 & 76 & 64 & 62
\\
\llamaM
 & 82 & 75 & 66 & 69 & 70
 & 86 & 93 & 82 & 79 & 82
 & 82 & 89 & 87 & 81 & 89
 & 83 & 80 & 83 & 75
 & 86 & 78 & 80 & 72
\\
\midrule
\avg
 & 80 & 74 & 70 & 71 & 70
 & 83 & 88 & 78 & 83 & 78
 & 84 & 86 & 84 & 80 & 86
 & 77 & 73 & 69 & 64
 & 76 & 73 & 66 & 62
\\
\bottomrule
\end{tabular}
}
\caption{\small AUROC scores of \textbf{\mam (self-attention)} method for \mistralS, \llamaS, \mistralM, and \llamaM across datasets and languages.}
\label{table:snyder_attention}
\end{table*}

\begin{table*}[!htb]
\centering
\small
\resizebox{\textwidth}{!}{
\begin{tabular}{l|ccccc|ccccc|ccccc|cccc|cccc|}
\toprule
\multirow{2}{*}{\bf Models} & \multicolumn{5}{|c|}{\bf mTREx -- Capitals} & \multicolumn{5}{|c|}{\bf mTREx -- Country} & \multicolumn{5}{|c|}{\bf mTREx -- Official Language} & \multicolumn{4}{|c|}{\bf G-MMLU -- STEM} & \multicolumn{4}{|c|}{\bf G-MMLU -- Humanities} 
\\
 & \en & \de & \hi & \bn & \ur
 & \en & \de & \hi & \bn & \ur
 & \en & \de & \hi & \bn & \ur
 & \en & \de & \hi & \bn 
 & \en & \de & \hi & \bn 
\\
\midrule
\mistralS
 & 82 & 72 & 75 & 74 & 71
 & 79 & 85 & 66 & 83 & 77
 & 81 & 84 & 75 & 77 & 86
 & 76 & 67 & 57 & 52
 & 73 & 67 & 57 & 51
\\
\llamaS
 & 79 & 74 & 72 & 74 & 70
 & 83 & 87 & 77 & 84 & 77
 & 88 & 85 & 84 & 81 & 84
 & 78 & 74 & 66 & 66
 & 78 & 74 & 63 & 63
\\
\mistralM
 & 76 & 77 & 72 & 77 & 70
 & 86 & 94 & 82 & 85 & 82
 & 82 & 89 & 86 & 86 & 86
 & 82 & 79 & 73 & 69
 & 78 & 81 & 65 & 64
\\
\llamaM
 & 80 & 77 & 65 & 69 & 71
 & 87 & 94 & 78 & 78 & 80
 & 82 & 88 & 88 & 82 & 86
 & 84 & 80 & 83 & 75
 & 86 & 79 & 81 & 72
\\
\midrule
\avg
 & 79 & 75 & 71 & 74 & 70
 & 84 & 90 & 76 & 82 & 79
 & 83 & 86 & 83 & 82 & 86
 & 80 & 75 & 70 & 66
 & 79 & 75 & 66 & 62
\\
\bottomrule
\end{tabular}
}
\caption{\small AUROC scores of \textbf{\mam (fully connected activations)} method for \mistralS, \llamaS, \mistralM, and \llamaM across datasets and languages.}
\label{table:snyder_fullyconnected}
\end{table*}

\begin{table*}[!htb]
\centering
\small
\resizebox{\textwidth}{!}{
\begin{tabular}{l|ccccc|ccccc|ccccc|cccc|cccc|}
\toprule
\multirow{2}{*}{\bf Models} & \multicolumn{5}{|c|}{\bf mTREx -- Capitals} & \multicolumn{5}{|c|}{\bf mTREx -- Country} & \multicolumn{5}{|c|}{\bf mTREx -- Official Language} & \multicolumn{4}{|c|}{\bf G-MMLU -- STEM} & \multicolumn{4}{|c|}{\bf G-MMLU -- Humanities} 
\\
 & \en & \de & \hi & \bn & \ur
 & \en & \de & \hi & \bn & \ur
 & \en & \de & \hi & \bn & \ur
 & \en & \de & \hi & \bn 
 & \en & \de & \hi & \bn 
\\
\midrule
\mistralS
 & 71 & 69 & 66 & 66 & 62
 & 76 & 72 & 67 & 58 & 66
 & 68 & 72 & 65 & 64 & 58
 & 57 & 59 & 53 & 53
 & 58 & 59 & 53 & 52
\\
\llamaS
 & 71 & 72 & 68 & 68 & 69
 & 78 & 77 & 67 & 58 & 65
 & 75 & 69 & 63 & 68 & 65
 & 63 & 64 & 57 & 55
 & 64 & 68 & 57 & 56
\\
\mistralM
 & 78 & 69 & 65 & 68 & 69
 & 81 & 87 & 71 & 77 & 63
 & 61 & 74 & 69 & 69 & 64
 & 63 & 63 & 54 & 54
 & 57 & 58 & 57 & 53
\\
\llamaM
 & 74 & 68 & 62 & 63 & 69
 & 69 & 82 & 48 & 61 & 76
 & 56 & 65 & 54 & 63 & 59
 & 54 & 52 & 53 & 55
 & 53 & 54 & 51 & 52
\\
\midrule
\avg
 & 74 & 70 & 65 & 66 & 67
 & 76 & 80 & 63 & 64 & 68
 & 65 & 70 & 63 & 66 & 62
 & 59 & 60 & 54 & 54
 & 58 & 60 & 54 & 53
\\
\bottomrule
\end{tabular}
}
\caption{\small AUROC scores of \textbf{SelfCheckGPT} method (\textbf{\sgm}) for \mistralS, \llamaS, \mistralM, and \llamaM across datasets and languages.}
\label{table:selfcheckgpt}
\end{table*}

\begin{table*}[!htb]
\centering
\small
\resizebox{\textwidth}{!}{
\begin{tabular}{l|ccccc|ccccc|ccccc|cccc|cccc|}
\toprule
\multirow{2}{*}{\bf Models} & \multicolumn{5}{|c|}{\bf mTREx -- Capitals} & \multicolumn{5}{|c|}{\bf mTREx -- Country} & \multicolumn{5}{|c|}{\bf mTREx -- Official Language} & \multicolumn{4}{|c|}{\bf G-MMLU -- STEM} & \multicolumn{4}{|c|}{\bf G-MMLU -- Humanities} 
\\
 & \en & \de & \hi & \bn & \ur
 & \en & \de & \hi & \bn & \ur
 & \en & \de & \hi & \bn & \ur
 & \en & \de & \hi & \bn 
 & \en & \de & \hi & \bn 
\\
\midrule
\mistralS
& 67 & 65 & 55 & 58 & 53
& 76 & 69 & 71 & 67 & 64
& 71 & 68 & 65 & 60 & 57
& 61 & 63 & 54 & 53 
& 64 & 61 & 55 & 54
\\
\llamaS
& 74 & 68 & 62 & 60 & 58
& 82 & 69 & 73 & 76 & 72
& 73 & 62 & 70 & 74 & 70
& 68 & 66 & 59 & 55 
& 67 & 68 & 60 & 57
\\
\mistralM
& 71 & 61 & 58 & 60 & 57
& 79 & 73 & 73 & 70 & 68
& 63 & 68 & 73 & 68 & 73
& 59 & 64 & 55 & 56
& 58 & 63 & 57 & 57
\\
\llamaM
& 71 & 64 & 62 & 67 & 69
& 70 & 61 & 61 & 69 & 76
& 64 & 67 & 67 & 63 & 75
& 60 & 60 & 61 & 58 
& 55 & 55 & 56 & 55
\\
\midrule
\avg
 & 71 & 64 & 59 & 61 & 59
 & 77 & 68 & 70 & 70 & 70
 & 68 & 66 & 69 & 66 & 69
 & 62 & 63 & 57 & 56
 & 61 & 62 & 57 & 56
\\
\bottomrule
\end{tabular}
}
\caption{\small AUROC scores of \textbf{Semantic Entropy} method (\textbf{\sem}) for \mistralS, \llamaS, \mistralM, and \llamaM across datasets and languages.}
\label{table:semanticentropy}
\end{table*}

%% file: final_tables/hd_auroc_percentage_appn.tex
\begin{table*}[!htb]
\centering
\small
\resizebox{\textwidth}{!}{
\begin{tabular}{l|ccccc|ccccc|ccccc|cccc|cccc|}
\toprule
\multirow{2}{*}{\bf Models} & \multicolumn{5}{|c|}{\bf mTREx -- Capitals} & \multicolumn{5}{|c|}{\bf mTREx -- Country} & \multicolumn{5}{|c|}{\bf mTREx -- Official Language} & \multicolumn{4}{|c|}{\bf G-MMLU -- STEM} & \multicolumn{4}{|c|}{\bf G-MMLU -- Humanities} \\
 & \en & \de & \hi & \bn & \ur & \en & \de & \hi & \bn & \ur & \en & \de & \hi & \bn & \ur & \en & \de & \hi & \bn & \en & \de & \hi & \bn \\
\midrule
\mistralS & 79 & \colorbox{red!20}{\scriptsize ↓11} & \colorbox{red!10}{\scriptsize ↓06} & \colorbox{red!10}{\scriptsize ↓06} & \colorbox{red!20}{\scriptsize ↓10} & 77 & \colorbox{green!10}{\scriptsize ↑05} & \colorbox{red!10}{\scriptsize ↓05} & \colorbox{green!10}{\scriptsize ↑09} & \colorbox{red!10}{\scriptsize ↓06} & 78 & \colorbox{green!10}{\scriptsize ↑06} & \colorbox{green!10}{\scriptsize ↑03} & \colorbox{red!10}{\scriptsize ↓05} & \colorbox{green!10}{\scriptsize ↑05} & 73 & \colorbox{red!20}{\scriptsize ↓10} & \colorbox{red!30}{\scriptsize ↓26} & \colorbox{red!30}{\scriptsize ↓30} & 72 & \colorbox{red!10}{\scriptsize ↓08} & \colorbox{red!30}{\scriptsize ↓24} & \colorbox{red!30}{\scriptsize ↓28} \\
\llamaS & 79 & \colorbox{red!10}{\scriptsize ↓06} & \colorbox{red!20}{\scriptsize ↓14} & \colorbox{red!20}{\scriptsize ↓13} & \colorbox{red!20}{\scriptsize ↓14} & 86 & \colorbox{green!10}{\scriptsize ↑01} & \colorbox{red!20}{\scriptsize ↓10} & \colorbox{red!10}{\scriptsize ↓03} & \colorbox{red!10}{\scriptsize ↓09} & 88 & \colorbox{red!10}{\scriptsize ↓05} & \colorbox{red!10}{\scriptsize ↓06} & \colorbox{red!10}{\scriptsize ↓08} & \colorbox{red!10}{\scriptsize ↓03} & 77 & \colorbox{red!10}{\scriptsize ↓04} & \colorbox{red!20}{\scriptsize ↓14} & \colorbox{red!20}{\scriptsize ↓14} & 78 & \colorbox{red!10}{\scriptsize ↓06} & \colorbox{red!20}{\scriptsize ↓17} & \colorbox{red!20}{\scriptsize ↓18} \\
\bottomrule
\end{tabular}
}
\caption{\small Percentage of increment \colorbox{green!17}{\scriptsize ↑} or decrement \colorbox{red!17}{\scriptsize ↓} in AUROC scores for \mistralS, \llamaS, \mistralM, and \llamaM across languages for \textbf{\mam (self-attention)} method w.r.t. the corresponding \en baseline for the model and dataset. Darker color shades indicate
a larger magnitude of percentage difference.}
\label{table:snyder_attention_delta_percentage}
\end{table*}

\begin{table*}[!htb]
\centering
\small
\resizebox{\textwidth}{!}{
\begin{tabular}{l|ccccc|ccccc|ccccc|cccc|cccc|}
\toprule
\multirow{2}{*}{\bf Models} & \multicolumn{5}{|c|}{\bf mTREx -- Capitals} & \multicolumn{5}{|c|}{\bf mTREx -- Country} & \multicolumn{5}{|c|}{\bf mTREx -- Official Language} & \multicolumn{4}{|c|}{\bf G-MMLU -- STEM} & \multicolumn{4}{|c|}{\bf G-MMLU -- Humanities} \\
 & \en & \de & \hi & \bn & \ur & \en & \de & \hi & \bn & \ur & \en & \de & \hi & \bn & \ur & \en & \de & \hi & \bn & \en & \de & \hi & \bn \\
\midrule
\mistralS & 71 & \colorbox{red!10}{\scriptsize ↓03} & \colorbox{red!10}{\scriptsize ↓07} & \colorbox{red!10}{\scriptsize ↓07} & \colorbox{red!20}{\scriptsize ↓13} & 76 & \colorbox{red!10}{\scriptsize ↓05} & \colorbox{red!20}{\scriptsize ↓12} & \colorbox{red!30}{\scriptsize ↓24} & \colorbox{red!20}{\scriptsize ↓13} & 68 & \colorbox{green!10}{\scriptsize ↑06} & \colorbox{red!10}{\scriptsize ↓04} & \colorbox{red!10}{\scriptsize ↓06} & \colorbox{red!20}{\scriptsize ↓15} & 57 & \colorbox{green!10}{\scriptsize ↑04} & \colorbox{red!10}{\scriptsize ↓07} & \colorbox{red!10}{\scriptsize ↓07} & 58 & \colorbox{green!10}{\scriptsize ↑02} & \colorbox{red!10}{\scriptsize ↓09} & \colorbox{red!20}{\scriptsize ↓10} \\
\llamaS & 71 & \colorbox{green!10}{\scriptsize ↑01} & \colorbox{red!10}{\scriptsize ↓04} & \colorbox{red!10}{\scriptsize ↓04} & \colorbox{red!10}{\scriptsize ↓03} & 78 & \colorbox{red!10}{\scriptsize ↓01} & \colorbox{red!20}{\scriptsize ↓14} & \colorbox{red!30}{\scriptsize ↓26} & \colorbox{red!20}{\scriptsize ↓17} & 75 & \colorbox{red!10}{\scriptsize ↓08} & \colorbox{red!20}{\scriptsize ↓16} & \colorbox{red!10}{\scriptsize ↓09} & \colorbox{red!20}{\scriptsize ↓13} & 63 & \colorbox{green!10}{\scriptsize ↑02} & \colorbox{red!20}{\scriptsize ↓10} & \colorbox{red!20}{\scriptsize ↓13} & 64 & \colorbox{green!10}{\scriptsize ↑06} & \colorbox{red!20}{\scriptsize ↓11} & \colorbox{red!20}{\scriptsize ↓12} \\
\bottomrule
\end{tabular}
}
\caption{\small Percentage of increment \colorbox{green!17}{\scriptsize ↑} or decrement \colorbox{red!17}{\scriptsize ↓} in AUROC scores for \mistralS, \llamaS, \mistralM, and \llamaM across languages for \textbf{SelfCheckGPT} method (\textbf{\sgm}) w.r.t. the corresponding \en baseline for the model and dataset.}
\label{table:selfcheckgpt_delta_percentage}
\end{table*}

\begin{table*}[!htb]
\centering
\small
\resizebox{\textwidth}{!}{
\begin{tabular}{l|ccccc|ccccc|ccccc|cccc|cccc|}
\toprule
\multirow{2}{*}{\bf Models} & \multicolumn{5}{|c|}{\bf mTREx -- Capitals} & \multicolumn{5}{|c|}{\bf mTREx -- Country} & \multicolumn{5}{|c|}{\bf mTREx -- Official Language} & \multicolumn{4}{|c|}{\bf G-MMLU -- STEM} & \multicolumn{4}{|c|}{\bf G-MMLU -- Humanities} \\
 & \en & \de & \hi & \bn & \ur & \en & \de & \hi & \bn & \ur & \en & \de & \hi & \bn & \ur & \en & \de & \hi & \bn & \en & \de & \hi & \bn \\
\midrule
\mistralS & 67 & \colorbox{red!10}{\scriptsize ↓03} & \colorbox{red!20}{\scriptsize ↓18} & \colorbox{red!20}{\scriptsize ↓13} & \colorbox{red!30}{\scriptsize ↓21} & 76 & \colorbox{red!10}{\scriptsize ↓09} & \colorbox{red!10}{\scriptsize ↓07} & \colorbox{red!20}{\scriptsize ↓12} & \colorbox{red!20}{\scriptsize ↓16} & 71 & \colorbox{red!10}{\scriptsize ↓04} & \colorbox{red!10}{\scriptsize ↓08} & \colorbox{red!20}{\scriptsize ↓15} & \colorbox{red!30}{\scriptsize ↓20} & 61 & \colorbox{green!10}{\scriptsize ↑03} & \colorbox{red!20}{\scriptsize ↓11} & \colorbox{red!20}{\scriptsize ↓13} & 64 & \colorbox{red!10}{\scriptsize ↓05} & \colorbox{red!20}{\scriptsize ↓14} & \colorbox{red!20}{\scriptsize ↓16} \\
\llamaS & 74 & \colorbox{red!10}{\scriptsize ↓08} & \colorbox{red!20}{\scriptsize ↓16} & \colorbox{red!20}{\scriptsize ↓19} & \colorbox{red!30}{\scriptsize ↓22} & 82 & \colorbox{red!20}{\scriptsize ↓16} & \colorbox{red!20}{\scriptsize ↓11} & \colorbox{red!10}{\scriptsize ↓07} & \colorbox{red!20}{\scriptsize ↓12} & 73 & \colorbox{red!20}{\scriptsize ↓15} & \colorbox{red!10}{\scriptsize ↓04} & \colorbox{green!10}{\scriptsize ↑01} & \colorbox{red!10}{\scriptsize ↓04} & 68 & \colorbox{red!10}{\scriptsize ↓03} & \colorbox{red!20}{\scriptsize ↓13} & \colorbox{red!20}{\scriptsize ↓19} & 67 & \colorbox{green!10}{\scriptsize ↑01} & \colorbox{red!20}{\scriptsize ↓10} & \colorbox{red!20}{\scriptsize ↓15} \\
\bottomrule
\end{tabular}
}
\caption{\small Percentage of increment \colorbox{green!17}{\scriptsize ↑} or decrement \colorbox{red!17}{\scriptsize ↓} in AUROC scores for \mistralS, \llamaS, \mistralM, and \llamaM across languages for \textbf{Semantic Entropy} method \textbf{(\sem)} w.r.t. the corresponding \en baseline for the model and dataset.}
\label{table:semanticentropy_delta_percentage}
\end{table*}

\begin{table*}[!htb]
\centering
\scriptsize
\resizebox{\textwidth}{!}{
\begin{tabular}{l|ccccc|ccccc|ccccc|}
\toprule
\multirow{2}{*}{\bf Models} & \multicolumn{5}{|c|}{\bf mTREx -- Capitals} & \multicolumn{5}{|c|}{\bf mTREx -- Country} & \multicolumn{5}{|c|}{\bf mTREx -- Official Language}
\\
 & \en & \de & \hi & \bn & \ur
 & \en & \de & \hi & \bn & \ur
 & \en & \de & \hi & \bn & \ur
\\
\midrule
\mistralS
 & 79 & 74 & 77 & 75 & 75
 & 82 & 86 & 76 & 88 & 83
 & 83 & 84 & 77 & 75 & 80
\\
\llamaS
 & 86 & 74 & 72 & 76 & 75
 & 84 & 88 & 83 & 85 & 85
 & 88 & 83 & 85 & 83 & 86
\\
\mistralM
 & 81 & 77 & 74 & 78 & 72
 & 81 & 94 & 82 & 86 & 86
 & 84 & 90 & 85 & 88 & 87 
\\
\midrule
\avg
 & 82 & 75 & 74 & 76 & 74
 & 82 & 89 & 80 & 86 & 84
 & 85 & 85 & 82 & 82 & 84
\\
\bottomrule
\end{tabular}
}
\caption{\small AUROC scores of \textbf{\mam (self-attention)} method [when averaging artifacts across multiple generated tokens] for \mistralS, \llamaS, and \mistralM over \mtrex datasets across languages.}
\label{table:mam_attention_avg_tkns}
\end{table*}

\begin{table*}[!htb]
\centering
\scriptsize
\resizebox{\textwidth}{!}{
\begin{tabular}{l|ccccc|ccccc|ccccc|}
\toprule
\multirow{2}{*}{\bf Models} & \multicolumn{5}{|c|}{\bf mTREx -- Capitals} & \multicolumn{5}{|c|}{\bf mTREx -- Country} & \multicolumn{5}{|c|}{\bf mTREx -- Official Language}
\\
 & \en & \de & \hi & \bn & \ur
 & \en & \de & \hi & \bn & \ur
 & \en & \de & \hi & \bn & \ur
\\
\midrule
\mistralS
 & 82 & 77 & 78 & 75 & 76
 & 85 & 90 & 76 & 89 & 89
 & 84 & 85 & 80 & 77 & 83
\\
\llamaS
 & 82 & 76 & 74 & 77 & 76
 & 83 & 88 & 84 & 87 & 85
 & 89 & 85 & 86 & 85 & 87
\\
\mistralM
 & 83 & 79 & 76 & 80 & 73
 & 85 & 95 & 83 & 89 & 88
 & 86 & 90 & 88 & 86 & 89
\\
\midrule
\avg
 & 82 & 77 & 75 & 77 & 74
 & 84 & 90 & 81 & 88 & 87
 & 86 & 86 & 84 & 82 & 86
\\
\bottomrule
\end{tabular}
}
\caption{\small AUROC scores of \textbf{\mam (fully connected activations)} method [when averaging artifacts across multiple generated tokens] for \mistralS, \llamaS, and \mistralM over \mtrex datasets across languages.}
\label{table:mam_fc_activation_avg_tkns}
\end{table*}

%% file: final_tables/ablation_exp_tables.tex
\begin{table*}[ht]
\centering
\small
\resizebox{\textwidth}{!}{
\begin{tabular}{|l|l|c|c|c|c|c|}
\hline
\textbf{Dataset} & \textbf{Model} & \textbf{\en} & \textbf{\de} & \textbf{\hi} & \textbf{\bn} & \textbf{\ur} \\
\hline
\multirow{4}{*}{mTREx -- Capitals} & \mistralS & 0.765 ± 0.019 & 0.717 ± 0.022 & 0.763 ± 0.022 & 0.722 ± 0.024 & 0.718 ± 0.036 \\
 & \llamaS & 0.767 ± 0.019 & 0.739 ± 0.021 & 0.690 ± 0.022 & 0.711 ± 0.026 & 0.677 ± 0.022 \\
 & \mistralM & 0.760 ± 0.024 & 0.777 ± 0.018 & 0.684 ± 0.031 & 0.713 ± 0.022 & 0.683 ± 0.023 \\
 & \llamaM & 0.785 ± 0.021 & 0.739 ± 0.021 & 0.669 ± 0.020 & 0.679 ± 0.018 & 0.717 ± 0.022 \\
\hline
\multirow{4}{*}{mTREx -- Country} & \mistralS & 0.768 ± 0.015 & 0.803 ± 0.021 & 0.718 ± 0.035 & 0.809 ± 0.066 & 0.750 ± 0.049 \\
 & \llamaS & 0.854 ± 0.020 & 0.879 ± 0.012 & 0.763 ± 0.020 & 0.830 ± 0.021 & 0.767 ± 0.017 \\
 & \mistralM & 0.849 ± 0.016 & 0.931 ± 0.006 & 0.793 ± 0.019 & 0.845 ± 0.015 & 0.806 ± 0.015 \\
\hline
\end{tabular}
}
\caption{\small AUROC scores (mean ± standard error) [without multiplying by 100] for \textbf{\mam (self-attention)} method where the classifiers are trained and tested across 20 runs in the same language setup for all LLMs, languages, and datasets. Note that Urdu (\ur) is not covered in G-MMLU.}
\label{tab:self-attn-ablation}
\end{table*}

\begin{table*}[ht]
\centering
\small
\resizebox{\textwidth}{!}{
\begin{tabular}{|l|l|c|c|c|c|c|}
\hline
\textbf{Dataset} & \textbf{Model} & \textbf{\en} & \textbf{\de} & \textbf{\hi} & \textbf{\bn} & \textbf{\ur} \\
\hline
\multirow{4}{*}{mTREx -- Capitals} & \mistralS & 0.789 ± 0.020 & 0.753 ± 0.020 & 0.771 ± 0.021 & 0.740 ± 0.024 & 0.728 ± 0.028 \\
 & \llamaS & 0.751 ± 0.021 & 0.731 ± 0.020 & 0.691 ± 0.026 & 0.728 ± 0.025 & 0.682 ± 0.024 \\
 & \mistralM & 0.742 ± 0.031 & 0.770 ± 0.022 & 0.694 ± 0.025 & 0.737 ± 0.021 & 0.708 ± 0.022 \\
 & \llamaM & 0.795 ± 0.022 & 0.743 ± 0.018 & 0.662 ± 0.026 & 0.697 ± 0.016 & 0.720 ± 0.024 \\
\hline
\multirow{4}{*}{mTREx -- Country} & \mistralS & 0.793 ± 0.014 & 0.857 ± 0.017 & 0.705 ± 0.044 & 0.830 ± 0.054 & 0.746 ± 0.049 \\
 & \llamaS & 0.835 ± 0.021 & 0.875 ± 0.012 & 0.764 ± 0.020 & 0.837 ± 0.021 & 0.772 ± 0.017 \\
 & \mistralM & 0.852 ± 0.015 & 0.931 ± 0.009 & 0.819 ± 0.016 & 0.856 ± 0.016 & 0.827 ± 0.015 \\
\hline
\end{tabular}
}
\caption{\small AUROC scores (mean ± standard error) [without multiplying by 100] for \textbf{\mam (fully connected activations)} method where the classifiers are trained and tested across 20 runs in the same language setup for all LLMs, languages, and datasets.}
\label{tab:fc-ablation}
\end{table*}

\begin{table*}[ht]
\centering
\small
\resizebox{\textwidth}{!}{
\begin{tabular}{|l|l|c|c|c|c|c|}
\hline
\textbf{Dataset} & \textbf{Model} & \textbf{\en} & \textbf{\de} & \textbf{\hi} & \textbf{\bn} & \textbf{\ur} \\
\hline
\multirow{3}{*}{mTREx -- Capitals} 
 & \mistralS   & 0.5785 $\pm$ 0.1443 & 0.5644 $\pm$ 0.1297 & 0.5377 $\pm$ 0.0979 & 0.5399 $\pm$ 0.0987 & 0.5234 $\pm$ 0.0710 \\
 & \llamaS   & 0.7102 $\pm$ 0.0047 & 0.7159 $\pm$ 0.0027 & 0.6768 $\pm$ 0.0033 & 0.6803 $\pm$ 0.0072 & 0.6907 $\pm$ 0.0038 \\
 & \mistralM  & 0.5904 $\pm$ 0.1854 & 0.5667 $\pm$ 0.1307 & 0.5439 $\pm$ 0.0975 & 0.5521 $\pm$ 0.1165 & 0.5579 $\pm$ 0.1242 \\
\hline
\multirow{3}{*}{mTREx -- Country} 
 & \mistralS   & 0.5875 $\pm$ 0.1724 & 0.5791 $\pm$ 0.1521 & 0.5641 $\pm$ 0.1172 & 0.5223 $\pm$ 0.0531 & 0.5437 $\pm$ 0.1027 \\
 & \llamaS   & 0.7792 $\pm$ 0.0037 & 0.7680 $\pm$ 0.0016 & 0.6656 $\pm$ 0.0020 & 0.5777 $\pm$ 0.0013 & 0.6414 $\pm$ 0.0065 \\
 & \mistralM  & 0.5981 $\pm$ 0.2050 & 0.6219 $\pm$ 0.2464 & 0.5587 $\pm$ 0.1327 & 0.5751 $\pm$ 0.1710 & 0.5308 $\pm$ 0.0828 \\
\hline
\multirow{3}{*}{mTREx -- Official Language} 
 & \mistralS   & 0.5575 $\pm$ 0.1180 & 0.5754 $\pm$ 0.1470 & 0.5299 $\pm$ 0.0894 & 0.5285 $\pm$ 0.0843 & 0.5295 $\pm$ 0.0585 \\
\hline
\end{tabular}
}
\caption{\small Average AUROC scores (mean $\pm$ standard error) [without multiplying by 100] for \textbf{SelfCheckGPT} method (\textbf{\sgm}) by varying the number of sample responses ($N \in \{10,15,20\}$) in high temperature settings for \mistralS, \llamaS, and \mistralM across all datasets and languages.}
\label{tab:sgm-ablation}
\end{table*}

\begin{table*}[ht]
\centering
\small
\resizebox{\textwidth}{!}{
\begin{tabular}{|l|l|c|c|c|c|c|}
\hline
\textbf{Dataset} & \textbf{Model} & \textbf{\en} & \textbf{\de} & \textbf{\hi} & \textbf{\bn} & \textbf{\ur} \\
\hline
\multirow{3}{*}{mTREx -- Capitals} 
 & \mistralS   & 0.6608 $\pm$ 0.0089 & 0.6596 $\pm$ 0.0019 & 0.5552 $\pm$ 0.0046 & 0.5812 $\pm$ 0.0058 & 0.5190 $\pm$ 0.0056 \\
 & \llamaS   & 0.7434 $\pm$ 0.0040 & 0.6684 $\pm$ 0.0078 & 0.6202 $\pm$ 0.0014 & 0.6095 $\pm$ 0.0032 & 0.5880 $\pm$ 0.0058 \\
 & \mistralM  & 0.7144 $\pm$ 0.0014 & 0.6148 $\pm$ 0.0037 & 0.5824 $\pm$ 0.0033 & 0.5928 $\pm$ 0.0043 & 0.5738 $\pm$ 0.0039 \\
\hline
\multirow{3}{*}{mTREx -- Country} 
 & \mistralS   & 0.7592 $\pm$ 0.0022 & 0.6981 $\pm$ 0.0041 & 0.7150 $\pm$ 0.0018 & 0.6595 $\pm$ 0.0077 & 0.6434 $\pm$ 0.0162 \\
 & \llamaS   & 0.8090 $\pm$ 0.0036 & 0.6839 $\pm$ 0.0017 & 0.7265 $\pm$ 0.0036 & 0.7504 $\pm$ 0.0094 & 0.7101 $\pm$ 0.0036 \\
 & \mistralM  & 0.7945 $\pm$ 0.0019 & 0.7406 $\pm$ 0.0070 & 0.7251 $\pm$ 0.0046 & 0.7047 $\pm$ 0.0018 & 0.6637 $\pm$ 0.0086 \\
\hline
\multirow{3}{*}{mTREx -- Official Language} 
 & \mistralS   & 0.6986 $\pm$ 0.0096 & 0.6836 $\pm$ 0.0029 & 0.6443 $\pm$ 0.0078 & 0.5952 $\pm$ 0.0117 & 0.5695 $\pm$ 0.0041 \\
\hline
\end{tabular}
}
\caption{\small Average AUROC scores (mean $\pm$ standard error) [without multiplying by 100] for \textbf{Semantic Entropy} method \textbf{(\sem)} by varying the number of sample responses ($N \in \{10,15,20\}$) in high temperature settings for \mistralS, \llamaS, and \mistralM across all datasets and languages.}
\label{tab:sem-ablation}
\end{table*}

%% file: final_tables/entropies_tables.tex
\begin{table*}[!htb]
\centering
\scriptsize
\resizebox{\textwidth}{!}{
\begin{tabular}{l|ccccc|ccccc|ccccc|}
\toprule
\multirow{2}{*}{\bf Models} & \multicolumn{5}{|c|}{\bf mTREx -- Capitals} & \multicolumn{5}{|c|}{\bf mTREx -- Country} & \multicolumn{5}{|c|}{\bf mTREx -- Official Language}
\\
 & \en & \de & \hi & \bn & \ur
 & \en & \de & \hi & \bn & \ur
 & \en & \de & \hi & \bn & \ur
\\
\midrule
\mistralS
 & 0.306 & 0.381 & 1.052 & 0.968 & 1.773
 & 0.154 & 0.625 & 1.310 & 1.344 & 1.205
 & 0.247 & 0.713 & 0.896 & 1.001 & 1.039
\\
\llamaS
 & 0.448 & 0.243 & 0.461 & 1.659 & 0.606
 & 0.485 & 0.673 & 0.351 & 2.101 & 0.466
 & 0.742 & 0.435 & 0.360 & 1.229 & 0.557
\\
\mistralM
 & 0.325 & 0.874 & 1.518 & 1.785 & 1.762
 & 0.304 & 0.722 & 1.066 & 1.521 & 1.050
 & 0.386 & 0.423 & 1.080 & 1.145 & 1.067
\\
\midrule
\avg
 & 0.360 & 0.499 & 1.010 & 1.471 & 1.380
 & 0.314 & 0.673 & 0.909 & 1.655 & 0.907
 & 0.458 & 0.524 & 0.778 & 1.125 & 0.887
\\
\bottomrule
\end{tabular}
}
\caption{\small Entropy of the softmax distribution for the first generated token for LLMs across languages over \mtrex datasets. Generator entropies are consistently low for high-resource languages and substantially higher for low-resource languages, indicating greater uncertainty during generation in low-resource ones.}
\label{table:generator_entropy}
\end{table*}

\begin{table*}[!htb]
\centering
\scriptsize
\resizebox{\textwidth}{!}{
\begin{tabular}{l|ccccc|ccccc|ccccc|}
\toprule
\multirow{2}{*}{\bf Models} & \multicolumn{5}{|c|}{\bf mTREx -- Capitals} & \multicolumn{5}{|c|}{\bf mTREx -- Country} & \multicolumn{5}{|c|}{\bf mTREx -- Official Language}
\\
 & \en & \de & \hi & \bn & \ur
 & \en & \de & \hi & \bn & \ur
 & \en & \de & \hi & \bn & \ur
\\
\midrule
\mistralS
 & 0.449 & 0.543 & 0.400 & 0.433 & 0.294
 & 0.342 & 0.442 & 0.203 & 0.124 & 0.183
 & 0.318 & 0.461 & 0.261 & 0.282 & 0.351
\\
\llamaS
 & 0.394 & 0.533 & 0.472 & 0.472 & 0.475
 & 0.303 & 0.394 & 0.486 & 0.436 & 0.473
 & 0.295 & 0.410 & 0.426 & 0.431 & 0.427
\\
\mistralM
 & 0.390 & 0.504 & 0.550 & 0.490 & 0.513
 & 0.330 & 0.290 & 0.515 & 0.469 & 0.473
 & 0.254 & 0.302 & 0.500 & 0.409 & 0.478
\\
\midrule
\avg
 & 0.411 & 0.527 & 0.474 & 0.465 & 0.427
 & 0.325 & 0.375 & 0.401 & 0.343 & 0.376
 & 0.289 & 0.391 & 0.396 & 0.374 & 0.419
\\
\bottomrule
\end{tabular}
}
\caption{\small Entropy of the detector's softmax distribution for \textbf{\mam (self-attention)} across languages and LLMs over \mtrex datasets. Unlike generators, detector entropies remain low and consistent across both high- and low-resource languages, indicating stable behavior that is independent of the language resource level.}
\label{table:mam_attention_entropy}
\end{table*}

\begin{table*}[!htb]
\centering
\scriptsize
\resizebox{\textwidth}{!}{
\begin{tabular}{l|ccccc|ccccc|ccccc|}
\toprule
\multirow{2}{*}{\bf Models} & \multicolumn{5}{|c|}{\bf mTREx -- Capitals} & \multicolumn{5}{|c|}{\bf mTREx -- Country} & \multicolumn{5}{|c|}{\bf mTREx -- Official Language}
\\
 & \en & \de & \hi & \bn & \ur
 & \en & \de & \hi & \bn & \ur
 & \en & \de & \hi & \bn & \ur
\\
\midrule
\mistralS
 & 0.482 & 0.570 & 0.400 & 0.510 & 0.304
 & 0.325 & 0.334 & 0.072 & 0.094 & 0.115
 & 0.258 & 0.461 & 0.148 & 0.222 & 0.219
\\
\llamaS
 & 0.242 & 0.523 & 0.434 & 0.459 & 0.374
 & 0.257 & 0.312 & 0.417 & 0.431 & 0.378
 & 0.268 & 0.399 & 0.356 & 0.420 & 0.338
\\
\mistralM
 & 0.212 & 0.541 & 0.469 & 0.484 & 0.455
 & 0.300 & 0.240 & 0.412 & 0.401 & 0.336
 & 0.214 & 0.287 & 0.375 & 0.372 & 0.396
\\
\midrule
\avg
 & 0.312 & 0.544 & 0.434 & 0.484 & 0.378
 & 0.294 & 0.295 & 0.300 & 0.309 & 0.277
 & 0.247 & 0.382 & 0.293 & 0.338 & 0.318
\\
\bottomrule
\end{tabular}
}
\caption{\small Entropy of the detector's softmax distribution for \textbf{\mam (fully connected activations)} across languages and LLMs over \mtrex datasets. Unlike generators, detector entropies remain low and consistent across both high- and low-resource languages, indicating stable behavior that is independent of the language resource level.}
\label{table:mam_fc_entropy}
\end{table*}

%% file: final_tables/cross_lingual_tables.tex
\begin{table*}[!htb]
\centering
\small
\resizebox{\textwidth}{!}{
\begin{tabular}{l|ccccc|ccccc|ccccc|cccc|cccc|}
\toprule
\multirow{2}{*}{\bf Models} & 
\multicolumn{5}{|c|}{\bf mTREx -- Capitals} & 
\multicolumn{5}{|c|}{\bf mTREx -- Country} & 
\multicolumn{5}{|c|}{\bf mTREx -- Official Language} & \multicolumn{4}{|c|}{\bf G-MMLU -- STEM} & 
\multicolumn{4}{|c|}{\bf G-MMLU -- Humanities} \\
 & \en & \de & \hi & \bn & \ur 
 & \en & \de & \hi & \bn & \ur 
 & \en & \de & \hi & \bn & \ur 
 & \en & \de & \hi & \bn 
 & \en & \de & \hi & \bn \\
\midrule
\mistralS
& 79 & 57 & 54 & 51 & 41
& 77 & 63 & 59 & 50 & 60
& 78 & 63 & 59 & 49 & 46
& 73 & 63 & 53 & 54
& 72 & 66 & 54 & 54
\\
\llamaS
& 79 & 67 & 53 & 56 & 57
& 86 & 59 & 63 & 65 & 54
& 88 & 69 & 47 & 66 & 49
& 77 & 72 & 67 & 65
& 78 & 65 & 62 & 63
\\
\mistralM
& 78 & 62 & 55 & 59 & 56
& 83 & 61 & 63 & 63 & 56
& 86 & 76 & 55 & 65 & 59
& 74 & 67 & 62 & 56
& 70 & 72 & 55 & 56
\\
\llamaM
& 82 & 65 & 54 & 57 & 55
& 86 & 68 & 56 & 71 & 57
& 82 & 63 & 55 & 61 & 54
& 83 & 79 & 70 & 73
& 86 & 76 & 70 & 68
\\
\midrule
\avg
& 80 & 62 & 54 & 56 & 52
& 83 & 62 & 60 & 62 & 56
& 84 & 67 & 54 & 60 & 52
& 77 & 70 & 62 & 62
& 76 & 69 & 60 & 60
\\
\bottomrule
\end{tabular}
}
\caption{\small AUROC scores of \textbf{\mam (self-attention)} method for \mistralS, \llamaS, \mistralM, and \llamaM across datasets and languages in the cross-lingual setting (train over \en data and test over target language data).}
\label{table:mam-self-attn-cross-lingual}
\end{table*}

\begin{table*}[!htb]
\centering
\small
\resizebox{\textwidth}{!}{
\begin{tabular}{l|ccccc|ccccc|ccccc|cccc|cccc|}
\toprule
\multirow{2}{*}{\bf Models} & \multicolumn{5}{|c|}{\bf mTREx -- Capitals} & \multicolumn{5}{|c|}{\bf mTREx -- Country} & \multicolumn{5}{|c|}{\bf mTREx -- Official Language} & \multicolumn{4}{|c|}{\bf G-MMLU -- STEM} & \multicolumn{4}{|c|}{\bf G-MMLU -- Humanities} \\
 & \en & \de & \hi & \bn & \ur & \en & \de & \hi & \bn & \ur & \en & \de & \hi & \bn & \ur & \en & \de & \hi & \bn & \en & \de & \hi & \bn \\
\midrule
\mistralS & 79 & \colorbox{red!30}{\scriptsize ↓28} & \colorbox{red!30}{\scriptsize ↓32} & \colorbox{red!30}{\scriptsize ↓35} & \colorbox{red!40}{\scriptsize ↓48} & 77 & \colorbox{red!20}{\scriptsize ↓18} & \colorbox{red!30}{\scriptsize ↓23} & \colorbox{red!30}{\scriptsize ↓35} & \colorbox{red!30}{\scriptsize ↓22} & 78 & \colorbox{red!20}{\scriptsize ↓19} & \colorbox{red!30}{\scriptsize ↓24} & \colorbox{red!30}{\scriptsize ↓37} & \colorbox{red!40}{\scriptsize ↓41} & 73 & \colorbox{red!20}{\scriptsize ↓14} & \colorbox{red!30}{\scriptsize ↓27} & \colorbox{red!30}{\scriptsize ↓26} & 72 & \colorbox{red!10}{\scriptsize ↓08} & \colorbox{red!30}{\scriptsize ↓25} & \colorbox{red!30}{\scriptsize ↓25} \\
\llamaS & 79 & \colorbox{red!20}{\scriptsize ↓15} & \colorbox{red!30}{\scriptsize ↓33} & \colorbox{red!30}{\scriptsize ↓29} & \colorbox{red!30}{\scriptsize ↓28} & 86 & \colorbox{red!30}{\scriptsize ↓31} & \colorbox{red!30}{\scriptsize ↓27} & \colorbox{red!30}{\scriptsize ↓24} & \colorbox{red!30}{\scriptsize ↓37} & 88 & \colorbox{red!30}{\scriptsize ↓22} & \colorbox{red!40}{\scriptsize ↓47} & \colorbox{red!30}{\scriptsize ↓25} & \colorbox{red!40}{\scriptsize ↓44} & 77 & \colorbox{red!10}{\scriptsize ↓06} & \colorbox{red!20}{\scriptsize ↓13} & \colorbox{red!20}{\scriptsize ↓16} & 78 & \colorbox{red!20}{\scriptsize ↓17} & \colorbox{red!30}{\scriptsize ↓21} & \colorbox{red!20}{\scriptsize ↓19} \\
\bottomrule
\end{tabular}
}
\caption{\small Percentage of increment \colorbox{green!17}{\scriptsize ↑} or decrement \colorbox{red!17}{\scriptsize ↓} in AUROC scores for \mistralS, \llamaS, \mistralM, and \llamaM across languages for \textbf{\mam (self-attention)} method w.r.t. the corresponding \en baseline for the model and dataset in the cross-lingual setting (train over \en data and test over target language data).}
\label{table:mam-self-attn-cross-lingual_delta_percentage}
\end{table*}

\begin{table*}[!htb]
\centering
\small
\resizebox{\textwidth}{!}{
\begin{tabular}{l|ccccc|ccccc|ccccc|cccc|cccc|}
\toprule
\multirow{2}{*}{\bf Models} & 
\multicolumn{5}{|c|}{\bf mTREx -- Capitals} & 
\multicolumn{5}{|c|}{\bf mTREx -- Country} & 
\multicolumn{5}{|c|}{\bf mTREx -- Official Language} & \multicolumn{4}{|c|}{\bf G-MMLU -- STEM} & 
\multicolumn{4}{|c|}{\bf G-MMLU -- Humanities} \\
 & \en & \de & \hi & \bn & \ur 
 & \en & \de & \hi & \bn & \ur 
 & \en & \de & \hi & \bn & \ur 
 & \en & \de & \hi & \bn 
 & \en & \de & \hi & \bn \\
\midrule
\mistralS
& 82 & 68 & 53 & 45 & 53
& 79 & 57 & 58 & 58 & 50
& 81 & 67 & 49 & 50 & 48
& 76 & 66 & 56 & 54
& 73 & 69 & 55 & 55
\\
\llamaS
& 79 & 67 & 54 & 61 & 52
& 83 & 53 & 63 & 63 & 55
& 88 & 65 & 58 & 50 & 49
& 78 & 73 & 69 & 64
& 78 & 71 & 63 & 60
\\
\mistralM
& 76 & 65 & 60 & 61 & 56
& 86 & 55 & 62 & 71 & 59
& 82 & 76 & 59 & 64 & 62
& 82 & 77 & 67 & 62
& 78 & 78 & 60 & 58
\\
\llamaM
& 80 & 66 & 59 & 46 & 59
& 87 & 66 & 61 & 65 & 52
& 82 & 61 & 58 & 55 & 56
& 84 & 79 & 79 & 68
& 86 & 77 & 71 & 68
\\
\midrule
\avg
& 79 & 66 & 56 & 53 & 54
& 84 & 57 & 60 & 64 & 54
& 83 & 67 & 56 & 54 & 53
& 80 & 73 & 67 & 61
& 79 & 73 & 62 & 60
\\
\bottomrule
\end{tabular}
}
\caption{\small AUROC scores of \textbf{\mam (fully connected activations)} method for \mistralS, \llamaS, \mistralM, and \llamaM across datasets and languages in the cross-lingual setting (train over \en data and test over target language data).}
\label{table:mam-fc-cross-lingual}
\end{table*}

\begin{table*}[!htb]
\centering
\small
\resizebox{\textwidth}{!}{
\begin{tabular}{l|ccccc|ccccc|ccccc|cccc|cccc|}
\toprule
\multirow{2}{*}{\bf Models} & \multicolumn{5}{|c|}{\bf mTREx -- Capitals} & \multicolumn{5}{|c|}{\bf mTREx -- Country} & \multicolumn{5}{|c|}{\bf mTREx -- Official Language} & \multicolumn{4}{|c|}{\bf G-MMLU -- STEM} & \multicolumn{4}{|c|}{\bf G-MMLU -- Humanities} \\
 & \en & \de & \hi & \bn & \ur & \en & \de & \hi & \bn & \ur & \en & \de & \hi & \bn & \ur & \en & \de & \hi & \bn & \en & \de & \hi & \bn \\
\midrule
\mistralS & 82 & \colorbox{red!20}{\scriptsize ↓17} & \colorbox{red!30}{\scriptsize ↓35} & \colorbox{red!40}{\scriptsize ↓45} & \colorbox{red!30}{\scriptsize ↓35} & 79 & \colorbox{red!30}{\scriptsize ↓28} & \colorbox{red!30}{\scriptsize ↓27} & \colorbox{red!30}{\scriptsize ↓27} & \colorbox{red!30}{\scriptsize ↓37} & 81 & \colorbox{red!20}{\scriptsize ↓17} & \colorbox{red!40}{\scriptsize ↓40} & \colorbox{red!30}{\scriptsize ↓38} & \colorbox{red!40}{\scriptsize ↓41} & 76 & \colorbox{red!20}{\scriptsize ↓13} & \colorbox{red!30}{\scriptsize ↓26} & \colorbox{red!30}{\scriptsize ↓29} & 73 & \colorbox{red!10}{\scriptsize ↓05} & \colorbox{red!30}{\scriptsize ↓25} & \colorbox{red!30}{\scriptsize ↓25} \\
\llamaS & 79 & \colorbox{red!20}{\scriptsize ↓15} & \colorbox{red!30}{\scriptsize ↓32} & \colorbox{red!30}{\scriptsize ↓23} & \colorbox{red!30}{\scriptsize ↓34} & 83 & \colorbox{red!30}{\scriptsize ↓36} & \colorbox{red!30}{\scriptsize ↓24} & \colorbox{red!30}{\scriptsize ↓24} & \colorbox{red!30}{\scriptsize ↓34} & 88 & \colorbox{red!30}{\scriptsize ↓26} & \colorbox{red!30}{\scriptsize ↓34} & \colorbox{red!40}{\scriptsize ↓43} & \colorbox{red!40}{\scriptsize ↓44} & 78 & \colorbox{red!10}{\scriptsize ↓06} & \colorbox{red!20}{\scriptsize ↓12} & \colorbox{red!20}{\scriptsize ↓18} & 78 & \colorbox{red!10}{\scriptsize ↓09} & \colorbox{red!20}{\scriptsize ↓19} & \colorbox{red!30}{\scriptsize ↓23} \\
\bottomrule
\end{tabular}
}
\caption{\small Percentage of increment \colorbox{green!17}{\scriptsize ↑} or decrement \colorbox{red!17}{\scriptsize ↓} in AUROC scores for \mistralS, \llamaS, \mistralM, and \llamaM across languages for \textbf{\mam (fully connected activations)} method w.r.t. the corresponding \en baseline for the model and dataset in the cross-lingual setting (train over \en data and test over target language data).}
\label{table:mam-fc-cross-lingual_delta_percentage}
\end{table*}

%% file: final_tables/multi_lingual_tables.tex
\begin{table*}[!htb]
\centering
\small
\resizebox{\textwidth}{!}{
\begin{tabular}{l|ccccc|ccccc|ccccc|cccc|cccc|}
\toprule
\multirow{2}{*}{\bf Models} & 
\multicolumn{5}{|c|}{\bf mTREx -- Capitals} & 
\multicolumn{5}{|c|}{\bf mTREx -- Country} & 
\multicolumn{5}{|c|}{\bf mTREx -- Official Language} & \multicolumn{4}{|c|}{\bf G-MMLU -- STEM} & 
\multicolumn{4}{|c|}{\bf G-MMLU -- Humanities} \\
 & \en & \de & \hi & \bn & \ur 
 & \en & \de & \hi & \bn & \ur 
 & \en & \de & \hi & \bn & \ur 
 & \en & \de & \hi & \bn 
 & \en & \de & \hi & \bn \\
\midrule
\mistralS
& 78 & 69 & 72 & 75 & 61
& 79 & 78 & 74 & 79 & 70
& 74 & 81 & 71 & 70 & 78
& 74 & 66 & 59 & 53
& 68 & 66 & 55 & 52
\\
\llamaS
& 80 & 71 & 65 & 70 & 67
& 83 & 84 & 78 & 83 & 77
& 88 & 84 & 83 & 79 & 82
& 77 & 74 & 69 & 69
& 76 & 74 & 65 & 63
\\
\mistralM
& 72 & 74 & 67 & 73 & 69
& 82 & 92 & 78 & 84 & 80
& 84 & 88 & 83 & 84 & 84
& 77 & 72 & 70 & 62
& 71 & 77 & 62 & 62
\\
\llamaM
& 82 & 75 & 68 & 69 & 73
& 86 & 92 & 82 & 77 & 82
& 78 & 87 & 85 & 78 & 86
& 85 & 81 & 85 & 75
& 85 & 78 & 80 & 72
\\
\midrule
\textbf{Average}
& 78 & 72 & 68 & 71 & 67
& 82 & 86 & 78 & 80 & 77
& 80 & 84 & 80 & 77 & 82
& 77 & 73 & 70 & 64
& 75 & 73 & 65 & 62
\\
\bottomrule
\end{tabular}
}
\caption{\small AUROC scores of \textbf{\mam (self-attention)} method for \mistralS, \llamaS, \mistralM, and \llamaM across datasets and languages in the multilingual setting (train over data across all languages and test over target language data).}
\label{table:mam-self-attn-multi-lingual}
\end{table*}

\begin{table*}[!htb]
\centering
\small
\resizebox{\textwidth}{!}{
\begin{tabular}{l|ccccc|ccccc|ccccc|cccc|cccc|}
\toprule
\multirow{2}{*}{\bf Models} & \multicolumn{5}{|c|}{\bf mTREx -- Capitals} & \multicolumn{5}{|c|}{\bf mTREx -- Country} & \multicolumn{5}{|c|}{\bf mTREx -- Official Language} & \multicolumn{4}{|c|}{\bf G-MMLU -- STEM} & \multicolumn{4}{|c|}{\bf G-MMLU -- Humanities} \\
 & \en & \de & \hi & \bn & \ur & \en & \de & \hi & \bn & \ur & \en & \de & \hi & \bn & \ur & \en & \de & \hi & \bn & \en & \de & \hi & \bn \\
\midrule
\mistralS & 78 & \colorbox{red!20}{\scriptsize ↓12} & \colorbox{red!10}{\scriptsize ↓08} & \colorbox{red!10}{\scriptsize ↓04} & \colorbox{red!30}{\scriptsize ↓22} & 79 & \colorbox{red!10}{\scriptsize ↓01} & \colorbox{red!10}{\scriptsize ↓06} & {\scriptsize ↓0} & \colorbox{red!20}{\scriptsize ↓11} & 74 & \colorbox{green!10}{\scriptsize ↑09} & \colorbox{red!10}{\scriptsize ↓04} & \colorbox{red!10}{\scriptsize ↓05} & \colorbox{green!10}{\scriptsize ↑05} & 74 & \colorbox{red!20}{\scriptsize ↓11} & \colorbox{red!30}{\scriptsize ↓20} & \colorbox{red!30}{\scriptsize ↓28} & 68 & \colorbox{red!10}{\scriptsize ↓03} & \colorbox{red!20}{\scriptsize ↓19} & \colorbox{red!30}{\scriptsize ↓24} \\
\llamaS & 80 & \colorbox{red!20}{\scriptsize ↓11} & \colorbox{red!20}{\scriptsize ↓19} & \colorbox{red!20}{\scriptsize ↓12} & \colorbox{red!20}{\scriptsize ↓16} & 83 & \colorbox{green!10}{\scriptsize ↑01} & \colorbox{red!10}{\scriptsize ↓06} & {\scriptsize ↓0} & \colorbox{red!10}{\scriptsize ↓07} & 88 & \colorbox{red!10}{\scriptsize ↓05} & \colorbox{red!10}{\scriptsize ↓06} & \colorbox{red!20}{\scriptsize ↓10} & \colorbox{red!10}{\scriptsize ↓07} & 77 & \colorbox{red!10}{\scriptsize ↓04} & \colorbox{red!20}{\scriptsize ↓10} & \colorbox{red!20}{\scriptsize ↓10} & 76 & \colorbox{red!10}{\scriptsize ↓03} & \colorbox{red!20}{\scriptsize ↓14} & \colorbox{red!20}{\scriptsize ↓17} \\
\bottomrule
\end{tabular}
}
\caption{\small Percentage of increment \colorbox{green!17}{\scriptsize ↑} or decrement \colorbox{red!17}{\scriptsize ↓} in AUROC scores for \mistralS, \llamaS, \mistralM, and \llamaM across languages for \textbf{\mam (self-attention)} method w.r.t. the corresponding \en baseline for the model and dataset in the multilingual setting (train over data across all languages and test over target language data).}
\label{table:mam-self-attn-multi-lingual_delta_percentage}
\end{table*}

\begin{table*}[!htb]
\centering
\small
\resizebox{\textwidth}{!}{
\begin{tabular}{l|ccccc|ccccc|ccccc|cccc|cccc|}
\toprule
\multirow{2}{*}{\bf Models} & 
\multicolumn{5}{|c|}{\bf mTREx -- Capitals} & 
\multicolumn{5}{|c|}{\bf mTREx -- Country} & 
\multicolumn{5}{|c|}{\bf mTREx -- Official Language} & \multicolumn{4}{|c|}{\bf G-MMLU -- STEM} & 
\multicolumn{4}{|c|}{\bf G-MMLU -- Humanities} \\
 & \en & \de & \hi & \bn & \ur 
 & \en & \de & \hi & \bn & \ur 
 & \en & \de & \hi & \bn & \ur 
 & \en & \de & \hi & \bn 
 & \en & \de & \hi & \bn \\
\midrule
\mistralS
& 79 & 72 & 72 & 74 & 67
& 80 & 85 & 74 & 79 & 78
& 79 & 83 & 71 & 74 & 80
& 75 & 66 & 58 & 55
& 71 & 68 & 57 & 51
\\
\llamaS
& 79 & 72 & 70 & 72 & 69
& 84 & 87 & 78 & 83 & 77
& 89 & 84 & 83 & 79 & 83
& 78 & 74 & 68 & 69
& 75 & 74 & 64 & 63
\\
\mistralM
& 79 & 75 & 72 & 77 & 71
& 84 & 94 & 79 & 86 & 83
& 84 & 89 & 84 & 85 & 85
& 82 & 78 & 71 & 69
& 76 & 80 & 66 & 63
\\
\llamaM
& 79 & 75 & 67 & 70 & 71
& 86 & 94 & 79 & 79 & 82
& 80 & 87 & 88 & 82 & 89
& 84 & 81 & 85 & 75
& 85 & 78 & 80 & 72
\\
\midrule
\textbf{Average}
& 78 & 73 & 70 & 73 & 69
& 83 & 89 & 77 & 81 & 79
& 83 & 85 & 81 & 79 & 84
& 80 & 75 & 70 & 66
& 76 & 74 & 66 & 62
\\
\bottomrule
\end{tabular}
}
\caption{\small AUROC scores of \textbf{\mam (fully connected activations)} method for \mistralS, \llamaS, \mistralM, and \llamaM across datasets and languages in the multilingual setting (train over data across all languages and test over target language data).}
\label{table:mam-fc-multi-lingual}
\end{table*}

\begin{table*}[!htb]
\centering
\small
\resizebox{\textwidth}{!}{
\begin{tabular}{l|ccccc|ccccc|ccccc|cccc|cccc|}
\toprule
\multirow{2}{*}{\bf Models} & \multicolumn{5}{|c|}{\bf mTREx -- Capitals} & \multicolumn{5}{|c|}{\bf mTREx -- Country} & \multicolumn{5}{|c|}{\bf mTREx -- Official Language} & \multicolumn{4}{|c|}{\bf G-MMLU -- STEM} & \multicolumn{4}{|c|}{\bf G-MMLU -- Humanities} \\
 & \en & \de & \hi & \bn & \ur & \en & \de & \hi & \bn & \ur & \en & \de & \hi & \bn & \ur & \en & \de & \hi & \bn & \en & \de & \hi & \bn \\
\midrule
\mistralS & 79 & \colorbox{red!10}{\scriptsize ↓09} & \colorbox{red!10}{\scriptsize ↓09} & \colorbox{red!10}{\scriptsize ↓06} & \colorbox{red!20}{\scriptsize ↓15} & 80 & \colorbox{green!10}{\scriptsize ↑06} & \colorbox{red!10}{\scriptsize ↓08} & \colorbox{red!10}{\scriptsize ↓01} & \colorbox{red!10}{\scriptsize ↓02} & 79 & \colorbox{green!10}{\scriptsize ↑05} & \colorbox{red!20}{\scriptsize ↓10} & \colorbox{red!10}{\scriptsize ↓06} & \colorbox{green!10}{\scriptsize ↑01} & 75 & \colorbox{red!20}{\scriptsize ↓12} & \colorbox{red!30}{\scriptsize ↓23} & \colorbox{red!30}{\scriptsize ↓27} & 71 & \colorbox{red!10}{\scriptsize ↓04} & \colorbox{red!30}{\scriptsize ↓20} & \colorbox{red!30}{\scriptsize ↓28} \\
\llamaS & 79 & \colorbox{red!10}{\scriptsize ↓09} & \colorbox{red!20}{\scriptsize ↓11} & \colorbox{red!10}{\scriptsize ↓09} & \colorbox{red!20}{\scriptsize ↓13} & 84 & \colorbox{green!10}{\scriptsize ↑04} & \colorbox{red!10}{\scriptsize ↓07} & \colorbox{red!10}{\scriptsize ↓01} & \colorbox{red!10}{\scriptsize ↓08} & 89 & \colorbox{red!10}{\scriptsize ↓06} & \colorbox{red!10}{\scriptsize ↓07} & \colorbox{red!20}{\scriptsize ↓11} & \colorbox{red!10}{\scriptsize ↓07} & 78 & \colorbox{red!10}{\scriptsize ↓05} & \colorbox{red!20}{\scriptsize ↓13} & \colorbox{red!20}{\scriptsize ↓12} & 75 & \colorbox{red!10}{\scriptsize ↓01} & \colorbox{red!20}{\scriptsize ↓15} & \colorbox{red!20}{\scriptsize ↓16} \\
\bottomrule
\end{tabular}
}
\caption{\small Percentage of increment \colorbox{green!17}{\scriptsize ↑} or decrement \colorbox{red!17}{\scriptsize ↓} in AUROC scores for \mistralS, \llamaS, \mistralM, and \llamaM across languages for \textbf{\mam (fully connected activations)} method w.r.t. the corresponding \en baseline for the model and dataset in the multilingual setting (train over data across all languages and test over target language data).}
\label{table:mam-fc-multi-lingual_delta_percentage}
\end{table*}

\begin{table*}[!htb]
\centering
\scriptsize
\resizebox{\textwidth}{!}{
\begin{tabular}{l|ccccc|ccccc|ccccc|}
\toprule
\multirow{2}{*}{\bf Models} & \multicolumn{5}{|c|}{\bf mTREx -- Capitals} & \multicolumn{5}{|c|}{\bf mTREx -- Country} & \multicolumn{5}{|c|}{\bf mTREx -- Official Language}
\\
 & \en & \de & \hi & \bn & \ur
 & \en & \de & \hi & \bn & \ur
 & \en & \de & \hi & \bn & \ur
\\
\midrule
\mistralS
 & 79 & 70 & 62 & 60 & 61
 & 82 & 60 & 70 & 57 & 63
 & 83 & 78 & 63 & 52 & 60
\\
\llamaS
 & 86 & 67 & 60 & 53 & 57
 & 84 & 72 & 74 & 48 & 68
 & 88 & 71 & 70 & 57 & 68
\\
\mistralM
 & 81 & 67 & 56 & 61 & 59
 & 81 & 76 & 66 & 66 & 64
 & 84 & 78 & 53 & 48 & 56
\\
\midrule
\avg
 & 82 & 67 & 59 & 58 & 58
 & 82 & 69 & 69 & 57 & 64
 & 85 & 75 & 62 & 51 & 61
\\
\bottomrule
\end{tabular}
}
\caption{\small AUROC scores of \textbf{\mam (self-attention)} method [when averaging artifacts across multiple generated tokens] for \mistralS, \llamaS, and \mistralM over \mtrex datasets across languages in the cross-lingual setting (train over \en data and test over target language data).}
\label{table:mam_attention_avg_tkns_cross_lingual}
\end{table*}

\begin{table*}[!htb]
\centering
\small
\resizebox{\textwidth}{!}{
\begin{tabular}{l|ccccc|ccccc|ccccc|}
\toprule
\multirow{2}{*}{\bf Models} & \multicolumn{5}{|c|}{\bf mTREx -- Capitals} & \multicolumn{5}{|c|}{\bf mTREx -- Country} & \multicolumn{5}{|c|}{\bf mTREx -- Official Language} \\
 & \en & \de & \hi & \bn & \ur & \en & \de & \hi & \bn & \ur & \en & \de & \hi & \bn & \ur \\
\midrule
\mistralS & 79 & \colorbox{red!20}{\scriptsize ↓11} & \colorbox{red!30}{\scriptsize ↓22} & \colorbox{red!30}{\scriptsize ↓24} & \colorbox{red!30}{\scriptsize ↓23} & 82 & \colorbox{red!30}{\scriptsize ↓27} & \colorbox{red!20}{\scriptsize ↓15} & \colorbox{red!30}{\scriptsize ↓30} & \colorbox{red!30}{\scriptsize ↓23} & 83 & \colorbox{red!10}{\scriptsize ↓06} & \colorbox{red!30}{\scriptsize ↓24} & \colorbox{red!30}{\scriptsize ↓37} & \colorbox{red!30}{\scriptsize ↓28} \\
\llamaS & 86 & \colorbox{red!30}{\scriptsize ↓22} & \colorbox{red!30}{\scriptsize ↓30} & \colorbox{red!30}{\scriptsize ↓38} & \colorbox{red!30}{\scriptsize ↓34} & 84 & \colorbox{red!20}{\scriptsize ↓14} & \colorbox{red!20}{\scriptsize ↓12} & \colorbox{red!40}{\scriptsize ↓43} & \colorbox{red!20}{\scriptsize ↓19} & 88 & \colorbox{red!20}{\scriptsize ↓19} & \colorbox{red!30}{\scriptsize ↓20} & \colorbox{red!30}{\scriptsize ↓35} & \colorbox{red!30}{\scriptsize ↓23} \\
\mistralM & 81 & \colorbox{red!20}{\scriptsize ↓17} & \colorbox{red!30}{\scriptsize ↓31} & \colorbox{red!30}{\scriptsize ↓25} & \colorbox{red!30}{\scriptsize ↓27} & 81 & \colorbox{red!10}{\scriptsize ↓06} & \colorbox{red!20}{\scriptsize ↓19} & \colorbox{red!20}{\scriptsize ↓19} & \colorbox{red!30}{\scriptsize ↓21} & 84 & \colorbox{red!10}{\scriptsize ↓07} & \colorbox{red!30}{\scriptsize ↓37} & \colorbox{red!40}{\scriptsize ↓43} & \colorbox{red!30}{\scriptsize ↓33} \\
\midrule
\avg & 82 & \colorbox{red!20}{\scriptsize ↓18} & \colorbox{red!30}{\scriptsize ↓28} & \colorbox{red!30}{\scriptsize ↓29} & \colorbox{red!30}{\scriptsize ↓29} & 82 & \colorbox{red!20}{\scriptsize ↓16} & \colorbox{red!20}{\scriptsize ↓16} & \colorbox{red!30}{\scriptsize ↓30} & \colorbox{red!30}{\scriptsize ↓22} & 85 & \colorbox{red!20}{\scriptsize ↓12} & \colorbox{red!30}{\scriptsize ↓27} & \colorbox{red!40}{\scriptsize ↓40} & \colorbox{red!30}{\scriptsize ↓28} \\
\bottomrule
\end{tabular}
}
\caption{\small Percentage of increment \colorbox{green!17}{\scriptsize ↑} or decrement \colorbox{red!17}{\scriptsize ↓} in AUROC scores for \mistralS, \llamaS, and \mistralM across languages for \textbf{\mam (self-attention)} method [when averaging artifacts across multiple generated tokens] w.r.t. the corresponding \en baseline for the model and dataset in the cross-lingual setting (train over \en data and test over target language data).}
\label{table:mam_attention_avg_tkns_cross_lingual_delta_percentage}
\end{table*}

\begin{table*}[!htb]
\centering
\scriptsize
\resizebox{\textwidth}{!}{
\begin{tabular}{l|ccccc|ccccc|ccccc|}
\toprule
\multirow{2}{*}{\bf Models} & \multicolumn{5}{|c|}{\bf mTREx -- Capitals} & \multicolumn{5}{|c|}{\bf mTREx -- Country} & \multicolumn{5}{|c|}{\bf mTREx -- Official Language}
\\
 & \en & \de & \hi & \bn & \ur
 & \en & \de & \hi & \bn & \ur
 & \en & \de & \hi & \bn & \ur
\\
\midrule
\mistralS
 & 82 & 71 & 59 & 61 & 65
 & 85 & 69 & 62 & 53 & 71
 & 84 & 80 & 66 & 53 & 62
\\
\llamaS
 & 82 & 67 & 51 & 60 & 52
 & 83 & 71 & 71 & 69 & 68
 & 89 & 75 & 58 & 48 & 57 
\\
\mistralM
 & 83 & 71 & 57 & 58 & 59
 & 85 & 72 & 67 & 62 & 68
 & 86 & 80 & 75 & 65 & 73
\\
\midrule
\avg
 & 82 & 69 & 55 & 59 & 58
 & 84 & 70 & 66 & 61 & 69
 & 86 & 78 & 66 & 55 & 63
\\
\bottomrule
\end{tabular}
}
\caption{\small AUROC scores of \textbf{\mam (fully connected activations)} method [when averaging artifacts across multiple generated tokens] for \mistralS, \llamaS, and \mistralM over \mtrex datasets across languages in the cross-lingual setting (train over \en data and test over target language data).}
\label{table:mam_fc_activation_avg_tkns_cross_lingual}
\end{table*}

\begin{table*}[!htb]
\centering
\small
\resizebox{\textwidth}{!}{
\begin{tabular}{l|ccccc|ccccc|ccccc|}
\toprule
\multirow{2}{*}{\bf Models} & \multicolumn{5}{|c|}{\bf mTREx -- Capitals} & \multicolumn{5}{|c|}{\bf mTREx -- Country} & \multicolumn{5}{|c|}{\bf mTREx -- Official Language} \\
 & \en & \de & \hi & \bn & \ur & \en & \de & \hi & \bn & \ur & \en & \de & \hi & \bn & \ur \\
\midrule
\mistralS & 82 & \colorbox{red!20}{\scriptsize ↓13} & \colorbox{red!30}{\scriptsize ↓28} & \colorbox{red!30}{\scriptsize ↓26} & \colorbox{red!30}{\scriptsize ↓21} & 85 & \colorbox{red!20}{\scriptsize ↓19} & \colorbox{red!30}{\scriptsize ↓27} & \colorbox{red!30}{\scriptsize ↓38} & \colorbox{red!20}{\scriptsize ↓16} & 84 & \colorbox{red!10}{\scriptsize ↓05} & \colorbox{red!30}{\scriptsize ↓21} & \colorbox{red!30}{\scriptsize ↓37} & \colorbox{red!30}{\scriptsize ↓26} \\
\llamaS & 82 & \colorbox{red!20}{\scriptsize ↓18} & \colorbox{red!30}{\scriptsize ↓38} & \colorbox{red!30}{\scriptsize ↓27} & \colorbox{red!30}{\scriptsize ↓37} & 83 & \colorbox{red!20}{\scriptsize ↓14} & \colorbox{red!20}{\scriptsize ↓14} & \colorbox{red!20}{\scriptsize ↓17} & \colorbox{red!20}{\scriptsize ↓18} & 89 & \colorbox{red!20}{\scriptsize ↓16} & \colorbox{red!30}{\scriptsize ↓35} & \colorbox{red!40}{\scriptsize ↓46} & \colorbox{red!30}{\scriptsize ↓36} \\
\mistralM & 83 & \colorbox{red!20}{\scriptsize ↓14} & \colorbox{red!30}{\scriptsize ↓31} & \colorbox{red!30}{\scriptsize ↓30} & \colorbox{red!30}{\scriptsize ↓29} & 85 & \colorbox{red!20}{\scriptsize ↓15} & \colorbox{red!30}{\scriptsize ↓21} & \colorbox{red!30}{\scriptsize ↓27} & \colorbox{red!30}{\scriptsize ↓20} & 86 & \colorbox{red!10}{\scriptsize ↓07} & \colorbox{red!20}{\scriptsize ↓13} & \colorbox{red!30}{\scriptsize ↓24} & \colorbox{red!20}{\scriptsize ↓15} \\
\midrule
\avg & 82 & \colorbox{red!20}{\scriptsize ↓16} & \colorbox{red!30}{\scriptsize ↓33} & \colorbox{red!30}{\scriptsize ↓28} & \colorbox{red!30}{\scriptsize ↓29} & 84 & \colorbox{red!20}{\scriptsize ↓17} & \colorbox{red!30}{\scriptsize ↓21} & \colorbox{red!30}{\scriptsize ↓27} & \colorbox{red!20}{\scriptsize ↓18} & 86 & \colorbox{red!10}{\scriptsize ↓09} & \colorbox{red!30}{\scriptsize ↓23} & \colorbox{red!30}{\scriptsize ↓36} & \colorbox{red!30}{\scriptsize ↓27} \\
\bottomrule
\end{tabular}
}
\caption{\small Percentage of increment \colorbox{green!17}{\scriptsize ↑} or decrement \colorbox{red!17}{\scriptsize ↓} in AUROC scores for \mistralS, \llamaS, and \mistralM across languages for \textbf{\mam (fully connected activations)} method [when averaging artifacts across multiple generated tokens] w.r.t. the corresponding \en baseline for the model and dataset in the cross-lingual setting (train over \en data and test over target language data).}
\label{table:mam_fc_activation_avg_tkns_cross_lingual_delta_percentage}
\end{table*}

\begin{table*}[!htb]
\centering
\scriptsize
\resizebox{\textwidth}{!}{
\begin{tabular}{l|ccccc|ccccc|ccccc|}
\toprule
\multirow{2}{*}{\bf Models} & \multicolumn{5}{|c|}{\bf mTREx -- Capitals} & \multicolumn{5}{|c|}{\bf mTREx -- Country} & \multicolumn{5}{|c|}{\bf mTREx -- Official Language}
\\
 & \en & \de & \hi & \bn & \ur
 & \en & \de & \hi & \bn & \ur
 & \en & \de & \hi & \bn & \ur
\\
\midrule
\mistralS
 & 76 & 69 & 74 & 72 & 73
 & 78 & 82 & 75 & 78 & 81
 & 81 & 79 & 73 & 69 & 75
\\
\llamaS
 & 82 & 72 & 73 & 74 & 76
 & 84 & 84 & 84 & 85 & 83
 & 89 & 81 & 84 & 80 & 83
\\
\mistralM
 & 78 & 75 & 71 & 77 & 70
 & 80 & 91 & 81 & 84 & 83
 & 82 & 87 & 85 & 86 & 85
\\
\midrule
\avg
 & 78 & 72 & 72 & 74 & 72
 & 80 & 85 & 79 & 82 & 82
 & 83 & 82 & 80 & 78 & 81
\\
\bottomrule
\end{tabular}
}
\caption{\small AUROC scores of \textbf{\mam (self-attention)} method [when averaging artifacts across multiple generated tokens] for \mistralS, \llamaS, and \mistralM over \mtrex datasets across languages in the multilingual setting (train over data across all languages and test over target language data).}
\label{table:mam_attention_avg_tkns_multi_lingual}
\end{table*}

\begin{table*}[!htb]
\centering
\small
\resizebox{\textwidth}{!}{
\begin{tabular}{l|ccccc|ccccc|ccccc|}
\toprule
\multirow{2}{*}{\bf Models} & \multicolumn{5}{|c|}{\bf mTREx -- Capitals} & \multicolumn{5}{|c|}{\bf mTREx -- Country} & \multicolumn{5}{|c|}{\bf mTREx -- Official Language} \\
 & \en & \de & \hi & \bn & \ur & \en & \de & \hi & \bn & \ur & \en & \de & \hi & \bn & \ur \\
\midrule
\mistralS & 76 & \colorbox{red!10}{\scriptsize ↓09} & \colorbox{red!10}{\scriptsize ↓03} & \colorbox{red!10}{\scriptsize ↓05} & \colorbox{red!10}{\scriptsize ↓04} & 78 & \colorbox{green!10}{\scriptsize ↑05} & \colorbox{red!10}{\scriptsize ↓04} & {\scriptsize ↓0} & \colorbox{green!10}{\scriptsize ↑04} & 81 & \colorbox{red!10}{\scriptsize ↓02} & \colorbox{red!20}{\scriptsize ↓10} & \colorbox{red!20}{\scriptsize ↓15} & \colorbox{red!10}{\scriptsize ↓07} \\
\llamaS & 82 & \colorbox{red!20}{\scriptsize ↓12} & \colorbox{red!20}{\scriptsize ↓11} & \colorbox{red!20}{\scriptsize ↓10} & \colorbox{red!10}{\scriptsize ↓07} & 84 & {\scriptsize ↓0} & {\scriptsize ↓0} & \colorbox{green!10}{\scriptsize ↑01} & \colorbox{red!10}{\scriptsize ↓01} & 89 & \colorbox{red!10}{\scriptsize ↓09} & \colorbox{red!10}{\scriptsize ↓06} & \colorbox{red!20}{\scriptsize ↓10} & \colorbox{red!10}{\scriptsize ↓07} \\
\mistralM & 78 & \colorbox{red!10}{\scriptsize ↓04} & \colorbox{red!10}{\scriptsize ↓09} & \colorbox{red!10}{\scriptsize ↓01} & \colorbox{red!20}{\scriptsize ↓10} & 80 & \colorbox{green!20}{\scriptsize ↑14} & \colorbox{green!10}{\scriptsize ↑01} & \colorbox{green!10}{\scriptsize ↑05} & \colorbox{green!10}{\scriptsize ↑04} & 82 & \colorbox{green!10}{\scriptsize ↑06} & \colorbox{green!10}{\scriptsize ↑04} & \colorbox{green!10}{\scriptsize ↑05} & \colorbox{green!10}{\scriptsize ↑04} \\
\midrule
\avg & 78 & \colorbox{red!10}{\scriptsize ↓08} & \colorbox{red!10}{\scriptsize ↓08} & \colorbox{red!10}{\scriptsize ↓05} & \colorbox{red!10}{\scriptsize ↓08} & 80 & \colorbox{green!10}{\scriptsize ↑06} & \colorbox{red!10}{\scriptsize ↓01} & \colorbox{green!10}{\scriptsize ↑02} & \colorbox{green!10}{\scriptsize ↑02} & 83 & \colorbox{red!10}{\scriptsize ↓01} & \colorbox{red!10}{\scriptsize ↓04} & \colorbox{red!10}{\scriptsize ↓06} & \colorbox{red!10}{\scriptsize ↓02} \\
\bottomrule
\end{tabular}
}
\caption{\small Percentage of increment \colorbox{green!17}{\scriptsize ↑} or decrement \colorbox{red!17}{\scriptsize ↓} in AUROC scores for \mistralS, \llamaS, and \mistralM across languages for \textbf{\mam (self-attention)} method [when averaging artifacts across multiple generated tokens] w.r.t. the corresponding \en baseline for the model and dataset in the multilingual setting (train over data across all languages and test over target language data).}
\label{table:mam_attention_avg_tkns_multi_lingual_delta_percentage}
\end{table*}

\begin{table*}[!htb]
\centering
\scriptsize
\resizebox{\textwidth}{!}{
\begin{tabular}{l|ccccc|ccccc|ccccc|}
\toprule
\multirow{2}{*}{\bf Models} & \multicolumn{5}{|c|}{\bf mTREx -- Capitals} & \multicolumn{5}{|c|}{\bf mTREx -- Country} & \multicolumn{5}{|c|}{\bf mTREx -- Official Language}
\\
 & \en & \de & \hi & \bn & \ur
 & \en & \de & \hi & \bn & \ur
 & \en & \de & \hi & \bn & \ur
\\
\midrule
\mistralS
 & 80 & 74 & 75 & 74 & 76
 & 84 & 88 & 76 & 80 & 88
 & 83 & 83 & 77 & 71 & 80
\\
\llamaS
 & 82 & 73 & 75 & 75 & 76
 & 83 & 87 & 84 & 86 & 84
 & 89 & 83 & 86 & 82 & 85  
\\
\mistralM
 & 82 & 77 & 74 & 79 & 70
 & 83 & 92 & 83 & 87 & 88
 & 84 & 89 & 86 & 87 & 86
\\
\midrule
\avg
 & 81 & 74 & 74 & 76 & 73
 & 83 & 88 & 81 & 84 & 86
 & 85 & 85 & 82 & 80 & 83
\\
\bottomrule
\end{tabular}
}
\caption{\small AUROC scores of \textbf{\mam (fully connected activations)} method [when averaging artifacts across multiple generated tokens] for \mistralS, \llamaS, and \mistralM over \mtrex datasets across languages in the multilingual setting (train over data across all languages and test over target language data).}
\label{table:mam_fc_activation_avg_tkns_multi_lingual}
\end{table*}

\begin{table*}[!htb]
\centering
\small
\resizebox{\textwidth}{!}{
\begin{tabular}{l|ccccc|ccccc|ccccc|}
\toprule
\multirow{2}{*}{\bf Models} & \multicolumn{5}{|c|}{\bf mTREx -- Capitals} & \multicolumn{5}{|c|}{\bf mTREx -- Country} & \multicolumn{5}{|c|}{\bf mTREx -- Official Language} \\
 & \en & \de & \hi & \bn & \ur & \en & \de & \hi & \bn & \ur & \en & \de & \hi & \bn & \ur \\
\midrule
\mistralS & 80 & \colorbox{red!10}{\scriptsize ↓08} & \colorbox{red!10}{\scriptsize ↓06} & \colorbox{red!10}{\scriptsize ↓08} & \colorbox{red!10}{\scriptsize ↓05} & 84 & \colorbox{green!10}{\scriptsize ↑05} & \colorbox{red!20}{\scriptsize ↓10} & \colorbox{red!10}{\scriptsize ↓05} & \colorbox{green!10}{\scriptsize ↑05} & 83 & {\scriptsize ↓0} & \colorbox{red!10}{\scriptsize ↓07} & \colorbox{red!20}{\scriptsize ↓14} & \colorbox{red!10}{\scriptsize ↓04} \\
\llamaS & 82 & \colorbox{red!20}{\scriptsize ↓11} & \colorbox{red!10}{\scriptsize ↓09} & \colorbox{red!10}{\scriptsize ↓09} & \colorbox{red!10}{\scriptsize ↓07} & 83 & \colorbox{green!10}{\scriptsize ↑05} & \colorbox{green!10}{\scriptsize ↑01} & \colorbox{green!10}{\scriptsize ↑04} & \colorbox{green!10}{\scriptsize ↑01} & 89 & \colorbox{red!10}{\scriptsize ↓07} & \colorbox{red!10}{\scriptsize ↓03} & \colorbox{red!10}{\scriptsize ↓08} & \colorbox{red!10}{\scriptsize ↓04} \\
\mistralM & 82 & \colorbox{red!10}{\scriptsize ↓06} & \colorbox{red!20}{\scriptsize ↓10} & \colorbox{red!10}{\scriptsize ↓04} & \colorbox{red!20}{\scriptsize ↓15} & 83 & \colorbox{green!20}{\scriptsize ↑11} & {\scriptsize ↓0} & \colorbox{green!10}{\scriptsize ↑05} & \colorbox{green!10}{\scriptsize ↑06} & 84 & \colorbox{green!10}{\scriptsize ↑06} & \colorbox{green!10}{\scriptsize ↑02} & \colorbox{green!10}{\scriptsize ↑04} & \colorbox{green!10}{\scriptsize ↑02} \\
\midrule
\avg & 81 & \colorbox{red!10}{\scriptsize ↓09} & \colorbox{red!10}{\scriptsize ↓09} & \colorbox{red!10}{\scriptsize ↓06} & \colorbox{red!20}{\scriptsize ↓10} & 83 & \colorbox{green!10}{\scriptsize ↑06} & \colorbox{red!10}{\scriptsize ↓02} & \colorbox{green!10}{\scriptsize ↑01} & \colorbox{green!10}{\scriptsize ↑04} & 85 & {\scriptsize ↓0} & \colorbox{red!10}{\scriptsize ↓04} & \colorbox{red!10}{\scriptsize ↓06} & \colorbox{red!10}{\scriptsize ↓02} \\
\bottomrule
\end{tabular}
}
\caption{\small Percentage of increment \colorbox{green!17}{\scriptsize ↑} or decrement \colorbox{red!17}{\scriptsize ↓} in AUROC scores for \mistralS, \llamaS, and \mistralM across languages for \textbf{\mam (fully connected activations)} method [when averaging artifacts across multiple generated tokens] w.r.t. the corresponding \en baseline for the model and dataset in the multilingual setting (train over data across all languages and test over target language data).}
\label{table:mam_fc_activation_avg_tkns_multi_lingual_delta_percentage}
\end{table*}

%% file: tables/appendix_snyder_attention_final.tex

\begin{table*}[!htb]
\centering
\small
\resizebox{\textwidth}{!}{
\begin{tabular}{l|ccccc|ccccc|ccccc|cccc|cccc|}
\toprule
\multirow{2}{*}{\bf Models} & \multicolumn{5}{|c|}{\bf mTREx -- Capitals} & \multicolumn{5}{|c|}{\bf mTREx -- Country} & \multicolumn{5}{|c|}{\bf mTREx -- Official Language} & \multicolumn{4}{|c|}{\bf G-MMLU -- STEM} & \multicolumn{4}{|c|}{\bf G-MMLU -- Humanities} \\
 & \en & \de & \hi & \bn & \ur & \en & \de & \hi & \bn & \ur & \en & \de & \hi & \bn & \ur & \en & \de & \hi & \bn & \en & \de & \hi & \bn \\
\midrule
\mistralS & 74 & 65~\colorbox{red!17}{\scriptsize ↓12} & 80~\colorbox{green!17}{\scriptsize ↑08} & 80~\colorbox{green!17}{\scriptsize ↑08} & 88~\colorbox{green!17}{\scriptsize ↑19} & 79 & 72~\colorbox{red!17}{\scriptsize ↓09} & 92~\colorbox{green!17}{\scriptsize ↑16} & 96~\colorbox{green!17}{\scriptsize ↑22} & 95~\colorbox{green!17}{\scriptsize ↑20} & 86 & 78~\colorbox{red!17}{\scriptsize ↓09} & 90~\colorbox{green!17}{\scriptsize ↑05} & 91~\colorbox{green!17}{\scriptsize ↑06} & 89~\colorbox{green!17}{\scriptsize ↑03} & 69 & 67~\colorbox{red!17}{\scriptsize ↓03} & 77~\colorbox{green!17}{\scriptsize ↑12} & 76~\colorbox{green!17}{\scriptsize ↑10} & 66 & 63~\colorbox{red!17}{\scriptsize ↓05} & 67~\colorbox{green!17}{\scriptsize ↑02} & 70~\colorbox{green!17}{\scriptsize ↑06} \\
\llamaS & 81 & 70~\colorbox{red!17}{\scriptsize ↓14} & 67~\colorbox{red!17}{\scriptsize ↓17} & 74~\colorbox{red!17}{\scriptsize ↓09} & 73~\colorbox{red!17}{\scriptsize ↓10} & 83 & 80~\colorbox{red!17}{\scriptsize ↓04} & 72~\colorbox{red!17}{\scriptsize ↓13} & 80~\colorbox{red!17}{\scriptsize ↓04} & 74~\colorbox{red!17}{\scriptsize ↓11} & 85 & 80~\colorbox{red!17}{\scriptsize ↓06} & 75~\colorbox{red!17}{\scriptsize ↓12} & 75~\colorbox{red!17}{\scriptsize ↓12} & 75~\colorbox{red!17}{\scriptsize ↓12} & 68 & 67~\colorbox{red!17}{\scriptsize ↓01} & 64~\colorbox{red!17}{\scriptsize ↓06} & 68 & 71 & 67~\colorbox{red!17}{\scriptsize ↓06} & 59~\colorbox{red!17}{\scriptsize ↓17} & 65~\colorbox{red!17}{\scriptsize ↓08} \\
\mistralM & 80 & 72~\colorbox{red!17}{\scriptsize ↓10} & 74~\colorbox{red!17}{\scriptsize ↓08} & 72~\colorbox{red!17}{\scriptsize ↓10} & 72~\colorbox{red!17}{\scriptsize ↓10} & 84 & 87~\colorbox{green!17}{\scriptsize ↑04} & 75~\colorbox{red!17}{\scriptsize ↓11} & 81~\colorbox{red!17}{\scriptsize ↓04} & 77~\colorbox{red!17}{\scriptsize ↓08} & 89 & 86~\colorbox{red!17}{\scriptsize ↓03} & 76~\colorbox{red!17}{\scriptsize ↓15} & 77~\colorbox{red!17}{\scriptsize ↓13} & 76~\colorbox{red!17}{\scriptsize ↓15} & 70 & 68~\colorbox{red!17}{\scriptsize ↓03} & 66~\colorbox{red!17}{\scriptsize ↓06} & 60~\colorbox{red!17}{\scriptsize ↓14} & 73 & 68~\colorbox{red!17}{\scriptsize ↓07} & 61~\colorbox{red!17}{\scriptsize ↓16} & 63~\colorbox{red!17}{\scriptsize ↓14} \\
\llamaM & 89 & 77~\colorbox{red!17}{\scriptsize ↓13} & 62~\colorbox{red!17}{\scriptsize ↓30} & 67~\colorbox{red!17}{\scriptsize ↓25} & 70~\colorbox{red!17}{\scriptsize ↓21} & 88 & 88~{\scriptsize ↓0} & 74~\colorbox{red!17}{\scriptsize ↓16} & 74~\colorbox{red!17}{\scriptsize ↓16} & 76~\colorbox{red!17}{\scriptsize ↓14} & 89 & 89~{\scriptsize ↓0} & 81~\colorbox{red!17}{\scriptsize ↓09} & 75~\colorbox{red!17}{\scriptsize ↓16} & 80~\colorbox{red!17}{\scriptsize ↓10} & 78 & 75~\colorbox{red!17}{\scriptsize ↓04} & 74~\colorbox{red!17}{\scriptsize ↓05} & 69~\colorbox{red!17}{\scriptsize ↓12} & 83 & 76~\colorbox{red!17}{\scriptsize ↓08} & 75~\colorbox{red!17}{\scriptsize ↓10} & 65~\colorbox{red!17}{\scriptsize ↓22} \\
\bottomrule
\end{tabular}
}
\caption{\small Classifier's accuracy of the \textbf{\mam (self-attention)} method for \mistralS, \llamaS, \mistralM, and \llamaM across datasets in English (\en), German (\de), Hindi (\hi), Bengali (\bn), and Urdu (\ur) languages. A value of \colorbox{green!17}{\scriptsize ↑X} or \colorbox{red!17}{\scriptsize ↓Y} represents an X\% increase or Y\% decrease in score w.r.t. the corresponding \en baseline for a model and dataset.}
\label{table:appendix_snyder_attention_delta}
\end{table*}

%% file: tables/appendix_snyder_fully_connected_final.tex
\begin{table*}[!htb]
\centering
\small
\resizebox{\textwidth}{!}{
\begin{tabular}{l|ccccc|ccccc|ccccc|cccc|cccc|}
\toprule
\multirow{2}{*}{\bf Models} & \multicolumn{5}{|c|}{\bf mTREx -- Capitals} & \multicolumn{5}{|c|}{\bf mTREx -- Country} & \multicolumn{5}{|c|}{\bf mTREx -- Official Language} & \multicolumn{4}{|c|}{\bf G-MMLU -- STEM} & \multicolumn{4}{|c|}{\bf G-MMLU -- Humanities} \\
 & \en & \de & \hi & \bn & \ur & \en & \de & \hi & \bn & \ur & \en & \de & \hi & \bn & \ur & \en & \de & \hi & \bn & \en & \de & \hi & \bn \\
\midrule
\mistralS & 74 & 68~\colorbox{red!17}{\scriptsize ↓08} & 79~\colorbox{green!17}{\scriptsize ↑07} & 79~\colorbox{green!17}{\scriptsize ↑07} & 89~\colorbox{green!17}{\scriptsize ↑20} & 81 & 78~\colorbox{red!17}{\scriptsize ↓04} & 92~\colorbox{green!17}{\scriptsize ↑14} & 96~\colorbox{green!17}{\scriptsize ↑19} & 95~\colorbox{green!17}{\scriptsize ↑17} & 87 & 81~\colorbox{red!17}{\scriptsize ↓07} & 90~\colorbox{green!17}{\scriptsize ↑03} & 91~\colorbox{green!17}{\scriptsize ↑05} & 91~\colorbox{green!17}{\scriptsize ↑05} & 71 & 67~\colorbox{red!17}{\scriptsize ↓06} & 77~\colorbox{green!17}{\scriptsize ↑08} & 76~\colorbox{green!17}{\scriptsize ↑07} & 68 & 65~\colorbox{red!17}{\scriptsize ↓04} & 63~\colorbox{red!17}{\scriptsize ↓07} & 70~\colorbox{green!17}{\scriptsize ↑03} \\
\llamaS & 79 & 70~\colorbox{red!17}{\scriptsize ↓11} & 70~\colorbox{red!17}{\scriptsize ↓11} & 76~\colorbox{red!17}{\scriptsize ↓04} & 74~\colorbox{red!17}{\scriptsize ↓06} & 82 & 79~\colorbox{red!17}{\scriptsize ↓04} & 72~\colorbox{red!17}{\scriptsize ↓12} & 80~\colorbox{red!17}{\scriptsize ↓02} & 75~\colorbox{red!17}{\scriptsize ↓09} & 86 & 80~\colorbox{red!17}{\scriptsize ↓07} & 78~\colorbox{red!17}{\scriptsize ↓09} & 75~\colorbox{red!17}{\scriptsize ↓13} & 76~\colorbox{red!17}{\scriptsize ↓12} & 69 & 68~\colorbox{red!17}{\scriptsize ↓01} & 63~\colorbox{red!17}{\scriptsize ↓09} & 68~\colorbox{red!17}{\scriptsize ↓01} & 70 & 68~\colorbox{red!17}{\scriptsize ↓03} & 58~\colorbox{red!17}{\scriptsize ↓17} & 65~\colorbox{red!17}{\scriptsize ↓07} \\
\mistralM & 80 & 71~\colorbox{red!17}{\scriptsize ↓11} & 71~\colorbox{red!17}{\scriptsize ↓11} & 71~\colorbox{red!17}{\scriptsize ↓11} & 73~\colorbox{red!17}{\scriptsize ↓09} & 83 & 88~\colorbox{green!17}{\scriptsize ↑06} & 75~\colorbox{red!17}{\scriptsize ↓10} & 82~\colorbox{red!17}{\scriptsize ↓01} & 79~\colorbox{red!17}{\scriptsize ↓05} & 88 & 84~\colorbox{red!17}{\scriptsize ↓05} & 79~\colorbox{red!17}{\scriptsize ↓10} & 77~\colorbox{red!17}{\scriptsize ↓12} & 79~\colorbox{red!17}{\scriptsize ↓10} & 75 & 72~\colorbox{red!17}{\scriptsize ↓04} & 67~\colorbox{red!17}{\scriptsize ↓11} & 65~\colorbox{red!17}{\scriptsize ↓13} & 77 & 74~\colorbox{red!17}{\scriptsize ↓04} & 64~\colorbox{red!17}{\scriptsize ↓17} & 63~\colorbox{red!17}{\scriptsize ↓18} \\
\llamaM & 88 & 77~\colorbox{red!17}{\scriptsize ↓12} & 61~\colorbox{red!17}{\scriptsize ↓31} & 67~\colorbox{red!17}{\scriptsize ↓24} & 73~\colorbox{red!17}{\scriptsize ↓17} & 86 & 87~\colorbox{green!17}{\scriptsize ↑01} & 72~\colorbox{red!17}{\scriptsize ↓16} & 71~\colorbox{red!17}{\scriptsize ↓17} & 72~\colorbox{red!17}{\scriptsize ↓16} & 88 & 87~\colorbox{red!17}{\scriptsize ↓01} & 81~\colorbox{red!17}{\scriptsize ↓08} & 77~\colorbox{red!17}{\scriptsize ↓12} & 79~\colorbox{red!17}{\scriptsize ↓10} & 76 & 75~\colorbox{red!17}{\scriptsize ↓01} & 74~\colorbox{red!17}{\scriptsize ↓03} & 68~\colorbox{red!17}{\scriptsize ↓11} & 83 & 76~\colorbox{red!17}{\scriptsize ↓08} & 73~\colorbox{red!17}{\scriptsize ↓12} & 67~\colorbox{red!17}{\scriptsize ↓19} \\
\bottomrule
\end{tabular}
}
\caption{\small Classifier's accuracy of the \textbf{\mam (fully connected activations)} method for \mistralS, \llamaS, \mistralM, and \llamaM across datasets in English (\en), German (\de), Hindi (\hi), Bengali (\bn), and Urdu (\ur) languages. A value of \colorbox{green!17}{\scriptsize ↑X} or \colorbox{red!17}{\scriptsize ↓Y} represents an X\% increase or Y\% decrease in score w.r.t. the corresponding \en baseline for a model and dataset.}
\label{table:appendix_snyder_fullyconnected_delta}
\end{table*}

%% file: final_tables/tphr_tables_main.tex
\begin{table*}[t]
\centering
\small
\resizebox{\textwidth}{!}{
\begin{tabular}{|l|cccc|cccc|cccc|ccc|ccc|}
\toprule
\multirow{2}{*}{\bf Models} & \multicolumn{4}{|c|}{\bf mTREx -- Capitals} & \multicolumn{4}{|c|}{\bf mTREx -- Country} & \multicolumn{4}{|c|}{\bf mTREx -- Official Language} & \multicolumn{3}{|c|}{\bf G-MMLU -- STEM} & \multicolumn{3}{|c|}{\bf G-MMLU -- Humanities} \\
 &  \de & \hi & \bn & \ur &  \de & \hi & \bn & \ur &  \de & \hi & \bn & \ur &  \de & \hi & \bn &  \de & \hi & \bn \\
\midrule
\multicolumn{19}{|c|}{\textbf{\mam (fully connected activations)}} \\
\hline
\mistralS  & \colorbox{MidnightBlue!10}{1.2} & \colorbox{MidnightBlue!20}{7.0} & \colorbox{MidnightBlue!20}{6.2} & \colorbox{MidnightBlue!20}{5.0}  & \colorbox{MidnightBlue!10}{3.8} & \colorbox{MidnightBlue!20}{5.1} & \colorbox{MidnightBlue!30}{17.0} & \colorbox{MidnightBlue!40}{34.0}  & \colorbox{MidnightBlue!10}{3.7} & \colorbox{MidnightBlue!20}{11.7} & \colorbox{MidnightBlue!30}{17.8} & \colorbox{MidnightBlue!20}{13.6}  & \colorbox{MidnightBlue!10}{0.8} & \colorbox{MidnightBlue!10}{1.0} & \colorbox{MidnightBlue!10}{0.8}  & \colorbox{MidnightBlue!10}{1.7} & \colorbox{MidnightBlue!10}{1.5} & \colorbox{MidnightBlue!10}{1.2} \\
\llamaS  & \colorbox{MidnightBlue!10}{1.6} & \colorbox{MidnightBlue!20}{6.0} & \colorbox{MidnightBlue!20}{9.2} & \colorbox{MidnightBlue!20}{5.1}  & \colorbox{MidnightBlue!20}{7.0} & \colorbox{MidnightBlue!20}{6.7} & \colorbox{MidnightBlue!40}{46.0} & \colorbox{MidnightBlue!20}{7.5}  & \colorbox{MidnightBlue!10}{0.0} & \colorbox{MidnightBlue!20}{6.5} & \colorbox{MidnightBlue!20}{5.1} & \colorbox{MidnightBlue!20}{7.5}  & \colorbox{MidnightBlue!10}{1.8} & \colorbox{MidnightBlue!10}{1.2} & \colorbox{MidnightBlue!10}{1.8}  & \colorbox{MidnightBlue!10}{2.5} & \colorbox{MidnightBlue!10}{1.1} & \colorbox{MidnightBlue!10}{1.6} \\
\mistralM  & \colorbox{MidnightBlue!30}{20.0} & \colorbox{MidnightBlue!20}{12.5} & \colorbox{MidnightBlue!50}{51.0} & \colorbox{MidnightBlue!20}{8.5}  & \colorbox{MidnightBlue!10}{3.5} & \colorbox{MidnightBlue!20}{9.8} & \colorbox{MidnightBlue!40}{46.0} & \colorbox{MidnightBlue!20}{12.0}  & \colorbox{MidnightBlue!10}{1.9} & \colorbox{MidnightBlue!20}{9.8} & \colorbox{MidnightBlue!20}{11.8} & \colorbox{MidnightBlue!20}{10.2}  & \colorbox{MidnightBlue!10}{2.0} & \colorbox{MidnightBlue!10}{2.3} & \colorbox{MidnightBlue!10}{1.9}  & \colorbox{MidnightBlue!10}{4.0} & \colorbox{MidnightBlue!10}{2.1} & \colorbox{MidnightBlue!10}{2.3} \\
\llamaM  & \colorbox{MidnightBlue!10}{4.0} & \colorbox{MidnightBlue!10}{2.8} & \colorbox{MidnightBlue!10}{4.5} & \colorbox{MidnightBlue!20}{6.1}  & \colorbox{MidnightBlue!10}{3.9} & \colorbox{MidnightBlue!10}{3.1} & \colorbox{MidnightBlue!10}{4.3} & \colorbox{MidnightBlue!20}{5.9}  & \colorbox{MidnightBlue!10}{1.0} & \colorbox{MidnightBlue!10}{4.2} & NA & \colorbox{MidnightBlue!20}{8.5}  & \colorbox{MidnightBlue!10}{1.2} & \colorbox{MidnightBlue!30}{20.0} & \colorbox{MidnightBlue!10}{1.1}  & \colorbox{MidnightBlue!10}{1.4} & \colorbox{MidnightBlue!10}{2.8} & \colorbox{MidnightBlue!10}{1.2} \\
\midrule
\avg  & \colorbox{MidnightBlue!10}{3.2} & \colorbox{MidnightBlue!20}{5.6} & \colorbox{MidnightBlue!20}{9.8} & \colorbox{MidnightBlue!20}{5.7}  & \colorbox{MidnightBlue!10}{4.5} & \colorbox{MidnightBlue!20}{5.5} & \colorbox{MidnightBlue!30}{25.0} & \colorbox{MidnightBlue!20}{10.2}  & \colorbox{MidnightBlue!10}{2.3} & NA & \colorbox{MidnightBlue!50}{50.0} & \colorbox{MidnightBlue!20}{14.3}  & \colorbox{MidnightBlue!10}{1.4} & \colorbox{MidnightBlue!10}{1.9} & \colorbox{MidnightBlue!10}{1.4}  & \colorbox{MidnightBlue!10}{2.5} & \colorbox{MidnightBlue!10}{1.5} & \colorbox{MidnightBlue!10}{1.4} \\
\midrule
\multicolumn{19}{|c|}{\textbf{\mam (self-attention)}} \\
\hline
\mistralS  & \colorbox{MidnightBlue!10}{1.3} & \colorbox{MidnightBlue!20}{9.8} & \colorbox{MidnightBlue!20}{10.0} & \colorbox{MidnightBlue!20}{6.9}  & \colorbox{MidnightBlue!20}{5.8} & \colorbox{MidnightBlue!30}{16.5} & \colorbox{MidnightBlue!20}{9.7} & \colorbox{MidnightBlue!20}{13.6}  & \colorbox{MidnightBlue!10}{2.2} & \colorbox{MidnightBlue!40}{35.0} & \colorbox{MidnightBlue!30}{17.8} & \colorbox{MidnightBlue!30}{17.0}  & \colorbox{MidnightBlue!10}{1.0} & \colorbox{MidnightBlue!10}{1.0} & \colorbox{MidnightBlue!10}{0.9}  & \colorbox{MidnightBlue!10}{1.7} & \colorbox{MidnightBlue!10}{1.4} & \colorbox{MidnightBlue!10}{1.4} \\
\llamaS  & \colorbox{MidnightBlue!10}{1.6} & \colorbox{MidnightBlue!10}{3.8} & \colorbox{MidnightBlue!10}{4.6} & \colorbox{MidnightBlue!10}{4.2}  & \colorbox{MidnightBlue!30}{28.0} & \colorbox{MidnightBlue!10}{4.4} & \colorbox{MidnightBlue!30}{15.3} & \colorbox{MidnightBlue!20}{5.6}  & \colorbox{MidnightBlue!10}{0.0} & \colorbox{MidnightBlue!20}{5.2} & \colorbox{MidnightBlue!20}{5.1} & \colorbox{MidnightBlue!20}{10.0}  & \colorbox{MidnightBlue!10}{2.3} & \colorbox{MidnightBlue!10}{1.3} & \colorbox{MidnightBlue!10}{1.9}  & \colorbox{MidnightBlue!10}{2.0} & \colorbox{MidnightBlue!10}{1.3} & \colorbox{MidnightBlue!10}{1.7} \\
\mistralM  & \colorbox{MidnightBlue!20}{10.0} & \colorbox{MidnightBlue!20}{7.1} & \colorbox{MidnightBlue!20}{10.2} & \colorbox{MidnightBlue!20}{6.4}  & \colorbox{MidnightBlue!10}{2.8} & \colorbox{MidnightBlue!20}{13.0} & \colorbox{MidnightBlue!30}{23.0} & \colorbox{MidnightBlue!40}{48.0}  & \colorbox{MidnightBlue!10}{4.3} & \colorbox{MidnightBlue!30}{19.5} & \colorbox{MidnightBlue!40}{47.0} & NA  & \colorbox{MidnightBlue!10}{3.0} & \colorbox{MidnightBlue!20}{10.5} & \colorbox{MidnightBlue!10}{2.1}  & \colorbox{MidnightBlue!10}{2.0} & \colorbox{MidnightBlue!10}{4.5} & \colorbox{MidnightBlue!10}{4.0} \\
\llamaM  & \colorbox{MidnightBlue!10}{1.7} & \colorbox{MidnightBlue!10}{2.6} & \colorbox{MidnightBlue!10}{3.8} & \colorbox{MidnightBlue!10}{4.6}  & \colorbox{MidnightBlue!10}{3.9} & \colorbox{MidnightBlue!20}{7.0} & \colorbox{MidnightBlue!20}{5.6} & \colorbox{MidnightBlue!20}{10.2}  & \colorbox{MidnightBlue!10}{0.9} & \colorbox{MidnightBlue!20}{5.0} & \colorbox{MidnightBlue!40}{47.0} & \colorbox{MidnightBlue!10}{4.9}  & \colorbox{MidnightBlue!10}{1.7} & NA & \colorbox{MidnightBlue!10}{1.2}  & \colorbox{MidnightBlue!10}{1.2} & \colorbox{MidnightBlue!10}{2.3} & \colorbox{MidnightBlue!10}{1.2} \\
\midrule
\avg  & \colorbox{MidnightBlue!10}{2.2} & \colorbox{MidnightBlue!10}{4.5} & \colorbox{MidnightBlue!20}{5.4} & \colorbox{MidnightBlue!20}{5.1}  & \colorbox{MidnightBlue!20}{5.4} & \colorbox{MidnightBlue!20}{8.8} & NA & \colorbox{MidnightBlue!20}{10.2}  & \colorbox{MidnightBlue!10}{3.5} & NA & \colorbox{MidnightBlue!20}{12.5} & \colorbox{MidnightBlue!30}{21.5}  & \colorbox{MidnightBlue!10}{1.8} & \colorbox{MidnightBlue!10}{2.4} & \colorbox{MidnightBlue!10}{1.5}  & \colorbox{MidnightBlue!10}{3.3} & \colorbox{MidnightBlue!10}{2.0} & \colorbox{MidnightBlue!10}{1.7} \\
\midrule
\multicolumn{19}{|c|}{\textbf{SelfCheckGPT}} \\
\hline
\mistralS  & \colorbox{MidnightBlue!20}{6.0} & \colorbox{MidnightBlue!20}{9.8} & \colorbox{MidnightBlue!20}{10.0} & \colorbox{MidnightBlue!20}{6.1}  & \colorbox{MidnightBlue!20}{5.8} & \colorbox{MidnightBlue!20}{7.3} & \colorbox{MidnightBlue!10}{3.8} & \colorbox{MidnightBlue!20}{6.8}  & \colorbox{MidnightBlue!10}{2.8} & \colorbox{MidnightBlue!30}{23.3} & \colorbox{MidnightBlue!30}{17.8} & \colorbox{MidnightBlue!20}{6.8}  & \colorbox{MidnightBlue!10}{3.5} & \colorbox{MidnightBlue!10}{4.8} & \colorbox{MidnightBlue!20}{5.0}  & \colorbox{MidnightBlue!20}{10.0} & \colorbox{MidnightBlue!10}{4.8} & \colorbox{MidnightBlue!10}{4.5} \\
\llamaS  & \colorbox{MidnightBlue!20}{8.0} & \colorbox{MidnightBlue!20}{14.0} & \colorbox{MidnightBlue!30}{15.3} & \colorbox{MidnightBlue!30}{23.0}  & \colorbox{MidnightBlue!30}{28.0} & \colorbox{MidnightBlue!10}{3.6} & \colorbox{MidnightBlue!10}{2.3} & \colorbox{MidnightBlue!10}{3.5}  & \colorbox{MidnightBlue!10}{0.0} & \colorbox{MidnightBlue!10}{2.2} & \colorbox{MidnightBlue!20}{5.1} & \colorbox{MidnightBlue!10}{3.0}  & \colorbox{MidnightBlue!20}{7.0} & \colorbox{MidnightBlue!10}{2.3} & \colorbox{MidnightBlue!10}{2.6}  & \colorbox{MidnightBlue!10}{2.5} & \colorbox{MidnightBlue!10}{2.4} & \colorbox{MidnightBlue!10}{3.0} \\
\mistralM  & \colorbox{MidnightBlue!10}{2.2} & \colorbox{MidnightBlue!10}{3.8} & \colorbox{MidnightBlue!20}{5.1} & \colorbox{MidnightBlue!20}{5.7}  & \colorbox{MidnightBlue!10}{4.7} & \colorbox{MidnightBlue!10}{3.9} & \colorbox{MidnightBlue!20}{11.5} & \colorbox{MidnightBlue!10}{2.7}  & \colorbox{MidnightBlue!10}{1.0} & \colorbox{MidnightBlue!10}{4.9} & \colorbox{MidnightBlue!20}{5.9} & \colorbox{MidnightBlue!20}{13.7}  & NA & \colorbox{MidnightBlue!10}{2.3} & \colorbox{MidnightBlue!10}{2.8}  & \colorbox{MidnightBlue!20}{12.0} & NA & \colorbox{MidnightBlue!20}{8.0} \\
\llamaM  & \colorbox{MidnightBlue!10}{2.0} & \colorbox{MidnightBlue!10}{3.5} & \colorbox{MidnightBlue!10}{4.5} & \colorbox{MidnightBlue!20}{11.0}  & \colorbox{MidnightBlue!10}{2.1} & \colorbox{MidnightBlue!10}{1.3} & \colorbox{MidnightBlue!10}{4.9} & \colorbox{MidnightBlue!20}{5.9}  & \colorbox{MidnightBlue!10}{0.7} & \colorbox{MidnightBlue!20}{12.5} & \colorbox{MidnightBlue!20}{6.7} & \colorbox{MidnightBlue!20}{11.3}  & \colorbox{MidnightBlue!10}{2.5} & \colorbox{MidnightBlue!30}{20.0} & \colorbox{MidnightBlue!20}{10.0}  & \colorbox{MidnightBlue!20}{10.0} & \colorbox{MidnightBlue!20}{7.0} & \colorbox{MidnightBlue!30}{17.0} \\
\midrule
\avg  & \colorbox{MidnightBlue!10}{3.2} & \colorbox{MidnightBlue!20}{5.0} & \colorbox{MidnightBlue!20}{6.1} & \colorbox{MidnightBlue!20}{7.3}  & \colorbox{MidnightBlue!20}{6.8} & \colorbox{MidnightBlue!10}{3.4} & \colorbox{MidnightBlue!10}{4.2} & \colorbox{MidnightBlue!20}{6.4}  & \colorbox{MidnightBlue!10}{1.4} & \colorbox{MidnightBlue!30}{20.0} & \colorbox{MidnightBlue!50}{50.0} & \colorbox{MidnightBlue!20}{14.3}  & \colorbox{MidnightBlue!20}{7.0} & \colorbox{MidnightBlue!10}{3.8} & \colorbox{MidnightBlue!10}{3.8}  & \colorbox{MidnightBlue!20}{5.0} & \colorbox{MidnightBlue!20}{5.0} & \colorbox{MidnightBlue!10}{4.8} \\
\midrule
\multicolumn{19}{|c|}{\textbf{Semantic Entropy}} \\
\hline
\mistralS  & \colorbox{MidnightBlue!20}{6.0} & \colorbox{MidnightBlue!10}{4.1} & \colorbox{MidnightBlue!20}{5.6} & \colorbox{MidnightBlue!10}{3.9}  & \colorbox{MidnightBlue!10}{3.3} & \colorbox{MidnightBlue!20}{13.2} & \colorbox{MidnightBlue!20}{7.6} & \colorbox{MidnightBlue!20}{5.7}  & \colorbox{MidnightBlue!10}{3.7} & \colorbox{MidnightBlue!20}{11.7} & \colorbox{MidnightBlue!20}{6.5} & \colorbox{MidnightBlue!10}{4.9}  & \colorbox{MidnightBlue!10}{3.5} & \colorbox{MidnightBlue!10}{2.7} & \colorbox{MidnightBlue!10}{2.5}  & \colorbox{MidnightBlue!10}{3.3} & \colorbox{MidnightBlue!10}{2.7} & \colorbox{MidnightBlue!10}{2.7} \\
\llamaS  & \colorbox{MidnightBlue!10}{1.3} & \colorbox{MidnightBlue!10}{3.5} & \colorbox{MidnightBlue!10}{3.3} & \colorbox{MidnightBlue!10}{2.9}  & \colorbox{MidnightBlue!10}{2.2} & \colorbox{MidnightBlue!10}{4.4} & \colorbox{MidnightBlue!20}{7.7} & \colorbox{MidnightBlue!10}{4.5}  & \colorbox{MidnightBlue!10}{0.0} & \colorbox{MidnightBlue!20}{8.7} & \colorbox{MidnightBlue!40}{36.0} & \colorbox{MidnightBlue!20}{10.0}  & \colorbox{MidnightBlue!10}{3.5} & \colorbox{MidnightBlue!10}{1.6} & \colorbox{MidnightBlue!10}{1.6}  & \colorbox{MidnightBlue!20}{10.0} & \colorbox{MidnightBlue!10}{2.4} & \colorbox{MidnightBlue!10}{2.4} \\
\mistralM  & \colorbox{MidnightBlue!10}{2.0} & \colorbox{MidnightBlue!10}{3.8} & \colorbox{MidnightBlue!10}{4.6} & \colorbox{MidnightBlue!10}{3.6}  & \colorbox{MidnightBlue!10}{4.7} & \colorbox{MidnightBlue!20}{6.5} & \colorbox{MidnightBlue!20}{5.1} & \colorbox{MidnightBlue!10}{4.4}  & \colorbox{MidnightBlue!10}{2.6} & \colorbox{MidnightBlue!10}{3.9} & \colorbox{MidnightBlue!20}{9.4} & \colorbox{MidnightBlue!10}{4.1}  & \colorbox{MidnightBlue!10}{1.2} & \colorbox{MidnightBlue!20}{5.2} & \colorbox{MidnightBlue!20}{8.3}  & \colorbox{MidnightBlue!10}{2.4} & \colorbox{MidnightBlue!30}{27.0} & \colorbox{MidnightBlue!40}{32.0} \\
\llamaM  & \colorbox{MidnightBlue!10}{1.7} & \colorbox{MidnightBlue!10}{4.7} & \colorbox{MidnightBlue!20}{12.2} & \colorbox{MidnightBlue!30}{27.5}  & \colorbox{MidnightBlue!10}{3.0} & \colorbox{MidnightBlue!10}{3.1} & \colorbox{MidnightBlue!40}{39.0} & \colorbox{MidnightBlue!20}{6.8}  & \colorbox{MidnightBlue!10}{2.0} & \colorbox{MidnightBlue!20}{8.3} & \colorbox{MidnightBlue!40}{47.0} & \colorbox{MidnightBlue!10}{3.1}  & NA & \colorbox{MidnightBlue!30}{20.0} & \colorbox{MidnightBlue!20}{5.0}  & NA & \colorbox{MidnightBlue!20}{14.0} & NA \\
\midrule
\avg  & \colorbox{MidnightBlue!10}{1.9} & \colorbox{MidnightBlue!10}{3.8} & \colorbox{MidnightBlue!10}{4.9} & \colorbox{MidnightBlue!10}{4.2}  & \colorbox{MidnightBlue!10}{3.0} & \colorbox{MidnightBlue!20}{6.3} & \colorbox{MidnightBlue!20}{7.1} & \colorbox{MidnightBlue!20}{7.3}  & \colorbox{MidnightBlue!10}{3.5} & \colorbox{MidnightBlue!40}{40.0} & \colorbox{MidnightBlue!30}{25.0} & \colorbox{MidnightBlue!40}{43.0}  & \colorbox{MidnightBlue!20}{7.0} & \colorbox{MidnightBlue!10}{3.8} & \colorbox{MidnightBlue!10}{3.2}  & \colorbox{MidnightBlue!20}{10.0} & \colorbox{MidnightBlue!20}{5.0} & \colorbox{MidnightBlue!10}{4.8} \\
\bottomrule
\end{tabular}
}
\caption{Ratio of model's task accuracy delta to HD's AUROC delta w.r.t \en (i.e., $10^{\metric}$ values) for \textbf{\mam (fully connected activations)}, \textbf{\mam (self-attention)}, \textbf{SelfCheckGPT}, and \textbf{Semantic Entropy} methods. Values where the AUROC delta is zero are marked as NA. Darker color shades indicate a higher value, signifying that although the task accuracy for these languages drops drastically, the HD's performance remains much more stable.}
\label{table:ratio_acc_auc_tphr_original}
\end{table*}

%% file: main_ArXiv_xelatex.bbl
\begin{thebibliography}{39}
\providecommand{\natexlab}[1]{#1}

\bibitem[{Abdaljalil et~al.(2025)Abdaljalil, Kurban, and Serpedin}]{Abdaljalil2025HalluVerse25FM}
Samir Abdaljalil, Hasan Kurban, and Erchin Serpedin. 2025.
\newblock \href {https://api.semanticscholar.org/CorpusID:276928068} {Halluverse25: Fine-grained multilingual benchmark dataset for llm hallucinations}.
\newblock \emph{ArXiv}, abs/2503.07833.

\bibitem[{Adlakha et~al.(2024)Adlakha, BehnamGhader, Lu, Meade, and Reddy}]{adlakha-etal-2024-evaluating}
Vaibhav Adlakha, Parishad BehnamGhader, Xing~Han Lu, Nicholas Meade, and Siva Reddy. 2024.
\newblock \href {https://doi.org/10.1162/tacl_a_00667} {Evaluating correctness and faithfulness of instruction-following models for question answering}.
\newblock \emph{Transactions of the Association for Computational Linguistics}, 12:681--699.

\bibitem[{Atari et~al.(2023)Atari, Xue, Park, Blasi, and Henrich}]{atari2023humans}
Mohammad Atari, Mona~J. Xue, Peter~S. Park, Dami{\'a}n~E. Blasi, and Joseph Henrich. 2023.
\newblock \href {https://psyarxiv.com/5b26t} {Which humans?}

\bibitem[{Azaria and Mitchell(2023)}]{azaria-mitchell-2023-internal}
Amos Azaria and Tom Mitchell. 2023.
\newblock \href {https://doi.org/10.18653/v1/2023.findings-emnlp.68} {The internal state of an {LLM} knows when it{'}s lying}.
\newblock In \emph{Findings of the Association for Computational Linguistics: EMNLP 2023}, pages 967--976, Singapore. Association for Computational Linguistics.

\bibitem[{Barocas et~al.(2023)Barocas, Hardt, and Narayanan}]{barocas-hardt-narayanan}
Solon Barocas, Moritz Hardt, and Arvind Narayanan. 2023.
\newblock \emph{Fairness and Machine Learning: Limitations and Opportunities}.
\newblock MIT Press.

\bibitem[{Bender et~al.(2021)Bender, Gebru, McMillan-Major, and Shmitchell}]{bender2021dangers}
Emily~M Bender, Timnit Gebru, Angelina McMillan-Major, and Shmargaret Shmitchell. 2021.
\newblock On the dangers of stochastic parrots: Can language models be too big?
\newblock In \emph{Proceedings of the 2021 ACM conference on fairness, accountability, and transparency}, pages 610--623.

\bibitem[{Chen et~al.(2018)Chen, Johansson, and Sontag}]{chen2018my}
Irene Chen, Fredrik~D Johansson, and David Sontag. 2018.
\newblock Why is my classifier discriminatory?
\newblock \emph{Advances in neural information processing systems}, 31.

\bibitem[{Datta et~al.(2023)Datta, Soni, Mukherjee, and Ghosh}]{datta-etal-2023-mildsum}
Debtanu Datta, Shubham Soni, Rajdeep Mukherjee, and Saptarshi Ghosh. 2023.
\newblock \href {https://doi.org/10.18653/v1/2023.emnlp-main.321} {{MILDS}um: A novel benchmark dataset for multilingual summarization of {I}ndian legal case judgments}.
\newblock In \emph{Proceedings of the 2023 Conference on Empirical Methods in Natural Language Processing}, pages 5291--5302, Singapore. Association for Computational Linguistics.

\bibitem[{Elsahar et~al.(2018)Elsahar, Vougiouklis, Remaci, Gravier, Hare, Laforest, and Simperl}]{trex_paper}
Hady Elsahar, Pavlos Vougiouklis, Arslen Remaci, Christophe Gravier, Jonathon Hare, Frederique Laforest, and Elena Simperl. 2018.
\newblock \href {https://aclanthology.org/L18-1544/} {{T}-{RE}x: A large scale alignment of natural language with knowledge base triples}.
\newblock In \emph{Proceedings of the Eleventh International Conference on Language Resources and Evaluation ({LREC} 2018)}, Miyazaki, Japan. European Language Resources Association (ELRA).

\bibitem[{Farquhar et~al.(2024)Farquhar, Kossen, Kuhn, and Gal}]{semantic_entropy_paper}
Sebastian Farquhar, Jannik Kossen, Lorenz Kuhn, and Yarin Gal. 2024.
\newblock \href {https://api.semanticscholar.org/CorpusID:270615909} {Detecting hallucinations in large language models using semantic entropy}.
\newblock \emph{Nature}, 630:625 -- 630.

\bibitem[{Feng et~al.(2022)Feng, Yang, Cer, Arivazhagan, and Wang}]{labse_paper}
Fangxiaoyu Feng, Yinfei Yang, Daniel Cer, Naveen Arivazhagan, and Wei Wang. 2022.
\newblock \href {https://doi.org/10.18653/v1/2022.acl-long.62} {Language-agnostic {BERT} sentence embedding}.
\newblock In \emph{Proceedings of the 60th Annual Meeting of the Association for Computational Linguistics (Volume 1: Long Papers)}, pages 878--891, Dublin, Ireland. Association for Computational Linguistics.

\bibitem[{Ferrando et~al.(2024)Ferrando, Obeso, Rajamanoharan, and Nanda}]{ferrando2024know}
Javier Ferrando, Oscar Obeso, Senthooran Rajamanoharan, and Neel Nanda. 2024.
\newblock Do i know this entity? knowledge awareness and hallucinations in language models.
\newblock \emph{arXiv preprint arXiv:2411.14257}.

\bibitem[{Gala et~al.(2023)Gala, Chitale, AK, Gumma, Doddapaneni, Kumar, Nawale, Sujatha, Puduppully, Raghavan, Kumar, Khapra, Dabre, and Kunchukuttan}]{gala2023indictrans2highqualityaccessiblemachine}
Jay Gala, Pranjal~A. Chitale, Raghavan AK, Varun Gumma, Sumanth Doddapaneni, Aswanth Kumar, Janki Nawale, Anupama Sujatha, Ratish Puduppully, Vivek Raghavan, Pratyush Kumar, Mitesh~M. Khapra, Raj Dabre, and Anoop Kunchukuttan. 2023.
\newblock \href {https://arxiv.org/abs/2305.16307} {Indictrans2: Towards high-quality and accessible machine translation models for all 22 scheduled indian languages}.
\newblock \emph{Preprint}, arXiv:2305.16307.

\bibitem[{Gasca(2023)}]{google_ai_search}
David Gasca. 2023.
\newblock \href {https://blog.google/products/search/search-labs-ai-announcement-/} {Help us build the future of search with search labs}.

\bibitem[{Gottweis et~al.(2025)Gottweis, Weng, Daryin, Tu, Palepu, Sirkovic, Myaskovsky, Weissenberger, Rong, Tanno, Saab, Popovici, Blum, Zhang, Chou, Hassidim, Gokturk, Vahdat, Kohli, Matias, Carroll, Kulkarni, Tomasev, Guan, Dhillon, Vaishnav, Lee, Costa, Penadés, Peltz, Xu, Pawlosky, Karthikesalingam, and Natarajan}]{gottweis2025aicoscientist}
Juraj Gottweis, Wei-Hung Weng, Alexander Daryin, Tao Tu, Anil Palepu, Petar Sirkovic, Artiom Myaskovsky, Felix Weissenberger, Keran Rong, Ryutaro Tanno, Khaled Saab, Dan Popovici, Jacob Blum, Fan Zhang, Katherine Chou, Avinatan Hassidim, Burak Gokturk, Amin Vahdat, Pushmeet Kohli, Yossi Matias, Andrew Carroll, Kavita Kulkarni, Nenad Tomasev, Yuan Guan, Vikram Dhillon, Eeshit~Dhaval Vaishnav, Byron Lee, Tiago R~D Costa, José~R Penadés, Gary Peltz, Yunhan Xu, Annalisa Pawlosky, Alan Karthikesalingam, and Vivek Natarajan. 2025.
\newblock \href {https://arxiv.org/abs/2502.18864} {Towards an ai co-scientist}.
\newblock \emph{Preprint}, arXiv:2502.18864.

\bibitem[{Haddow et~al.(2022)Haddow, Bawden, Miceli~Barone, Helcl, and Birch}]{haddow-etal-2022-survey}
Barry Haddow, Rachel Bawden, Antonio~Valerio Miceli~Barone, Jind{\v{r}}ich Helcl, and Alexandra Birch. 2022.
\newblock \href {https://doi.org/10.1162/coli_a_00446} {Survey of low-resource machine translation}.
\newblock \emph{Computational Linguistics}, 48(3):673--732.

\bibitem[{Hendrycks et~al.(2021)Hendrycks, Burns, Basart, Zou, Mazeika, Song, and Steinhardt}]{hendrycks2021measuringmassivemultitasklanguage}
Dan Hendrycks, Collin Burns, Steven Basart, Andy Zou, Mantas Mazeika, Dawn Song, and Jacob Steinhardt. 2021.
\newblock \href {https://arxiv.org/abs/2009.03300} {Measuring massive multitask language understanding}.
\newblock \emph{Preprint}, arXiv:2009.03300.

\bibitem[{Ji et~al.(2023)Ji, Lee, Frieske, Yu, Su, Xu, Ishii, Bang, Madotto, and Fung}]{ji2023survey}
Ziwei Ji, Nayeon Lee, Rita Frieske, Tiezheng Yu, Dan Su, Yan Xu, Etsuko Ishii, Ye~Jin Bang, Andrea Madotto, and Pascale Fung. 2023.
\newblock Survey of hallucination in natural language generation.
\newblock \emph{ACM computing surveys}, 55(12):1--38.

\bibitem[{Jiang et~al.(2020)Jiang, Anastasopoulos, Araki, Ding, and Neubig}]{jiang-etal-2020-multilingual}
Zhengbao Jiang, Antonios Anastasopoulos, Jun Araki, Haibo Ding, and Graham Neubig. 2020.
\newblock \href {https://doi.org/10.18653/v1/2020.emnlp-main.479} {{X}-{FACTR}: Multilingual factual knowledge retrieval from pretrained language models}.
\newblock In \emph{Proceedings of the 2020 Conference on Empirical Methods in Natural Language Processing (EMNLP)}, pages 5943--5959, Online. Association for Computational Linguistics.

\bibitem[{Liang et~al.(2023)Liang, Bommasani, Lee, Tsipras, Soylu, Yasunaga, Zhang, Narayanan, Wu, Kumar, Newman, Yuan, Yan, Zhang, Cosgrove, Manning, Re, Acosta-Navas, Hudson, Zelikman, Durmus, Ladhak, Rong, Ren, Yao, WANG, Santhanam, Orr, Zheng, Yuksekgonul, Suzgun, Kim, Guha, Chatterji, Khattab, Henderson, Huang, Chi, Xie, Santurkar, Ganguli, Hashimoto, Icard, Zhang, Chaudhary, Wang, Li, Mai, Zhang, and Koreeda}]{liang2023holistic}
Percy Liang, Rishi Bommasani, Tony Lee, Dimitris Tsipras, Dilara Soylu, Michihiro Yasunaga, Yian Zhang, Deepak Narayanan, Yuhuai Wu, Ananya Kumar, Benjamin Newman, Binhang Yuan, Bobby Yan, Ce~Zhang, Christian~Alexander Cosgrove, Christopher~D Manning, Christopher Re, Diana Acosta-Navas, Drew~Arad Hudson, Eric Zelikman, Esin Durmus, Faisal Ladhak, Frieda Rong, Hongyu Ren, Huaxiu Yao, Jue WANG, Keshav Santhanam, Laurel Orr, Lucia Zheng, Mert Yuksekgonul, Mirac Suzgun, Nathan Kim, Neel Guha, Niladri~S. Chatterji, Omar Khattab, Peter Henderson, Qian Huang, Ryan~Andrew Chi, Sang~Michael Xie, Shibani Santurkar, Surya Ganguli, Tatsunori Hashimoto, Thomas Icard, Tianyi Zhang, Vishrav Chaudhary, William Wang, Xuechen Li, Yifan Mai, Yuhui Zhang, and Yuta Koreeda. 2023.
\newblock \href {https://openreview.net/forum?id=iO4LZibEqW} {Holistic evaluation of language models}.
\newblock \emph{Transactions on Machine Learning Research}.
\newblock Featured Certification, Expert Certification.

\bibitem[{Lin et~al.(2021)Lin, Hilton, and Evans}]{lin2021truthfulqa}
Stephanie Lin, Jacob Hilton, and Owain Evans. 2021.
\newblock Truthfulqa: Measuring how models mimic human falsehoods.
\newblock \emph{arXiv preprint arXiv:2109.07958}.

\bibitem[{Mahapatra et~al.(2025)Mahapatra, Datta, Soni, Goswami, and Ghosh}]{10.1145/3748313}
Sayan Mahapatra, Debtanu Datta, Shubham Soni, Adrijit Goswami, and Saptarshi Ghosh. 2025.
\newblock \href {https://doi.org/10.1145/3748313} {Milpac: A novel benchmark for evaluating translation of legal text to indian languages}.
\newblock \emph{ACM Trans. Asian Low-Resour. Lang. Inf. Process.}, 24(8).

\bibitem[{Manakul et~al.(2023)Manakul, Liusie, and Gales}]{manakul-etal-2023-selfcheckgpt}
Potsawee Manakul, Adian Liusie, and Mark Gales. 2023.
\newblock \href {https://doi.org/10.18653/v1/2023.emnlp-main.557} {{S}elf{C}heck{GPT}: Zero-resource black-box hallucination detection for generative large language models}.
\newblock In \emph{Proceedings of the 2023 Conference on Empirical Methods in Natural Language Processing}, pages 9004--9017, Singapore. Association for Computational Linguistics.

\bibitem[{Moayeri et~al.(2024)Moayeri, Tabassi, and Feizi}]{moayeri2024worldbench}
Mazda Moayeri, Elham Tabassi, and Soheil Feizi. 2024.
\newblock Worldbench: Quantifying geographic disparities in llm factual recall.
\newblock In \emph{Proceedings of the 2024 ACM Conference on Fairness, Accountability, and Transparency}, pages 1211--1228.

\bibitem[{Neitemeier et~al.(2025)Neitemeier, Deiseroth, Eichenberg, and Balles}]{neitemeier2025hierarchical}
Pit Neitemeier, Bj{\"o}rn Deiseroth, Constantin Eichenberg, and Lukas Balles. 2025.
\newblock \href {https://openreview.net/forum?id=tU074jg2vS} {Hierarchical autoregressive transformers: Combining byte- and word-level processing for robust, adaptable language models}.
\newblock In \emph{The Thirteenth International Conference on Learning Representations}.

\bibitem[{Pava et~al.(2025)Pava, Meinhardt, Zaman, Friedman, Truong, Zhang, Marivate, and Koyejo}]{low-resource-paper}
Juan~N. Pava, Caroline Meinhardt, Haifa Badi~Uz Zaman, Toni Friedman, Sang~T. Truong, Daniel Zhang, Vukosi Marivate, and Sanmi Koyejo. 2025.
\newblock \href {https://hai.stanford.edu/policy/mind-the-language-gap-mapping-the-challenges-of-llm-development-in-low-resource-language-contexts} {Mind the (language) gap: Mapping the challenges of {LLM} development in low-resource language contexts}.
\newblock Last accessed on 2025-07-28.

\bibitem[{Peng et~al.(2023)Peng, Kalliamvakou, Cihon, and Demirer}]{Peng2023TheIO}
Sida Peng, Eirini Kalliamvakou, Peter Cihon, and Mert Demirer. 2023.
\newblock \href {https://api.semanticscholar.org/CorpusID:256826778} {The impact of ai on developer productivity: Evidence from github copilot}.
\newblock \emph{ArXiv}, abs/2302.06590.

\bibitem[{Rawte et~al.(2023)Rawte, Sheth, and Das}]{rawte2023survey}
Vipula Rawte, Amit Sheth, and Amitava Das. 2023.
\newblock A survey of hallucination in large foundation models.
\newblock \emph{arXiv preprint arXiv:2309.05922}.

\bibitem[{Schw{\"o}bel et~al.(2023)Schw{\"o}bel, Golebiowski, Donini, Archambeau, and Pruthi}]{schwobel-etal-2023-geographical}
Pola Schw{\"o}bel, Jacek Golebiowski, Michele Donini, Cedric Archambeau, and Danish Pruthi. 2023.
\newblock \href {https://doi.org/10.18653/v1/2023.findings-emnlp.823} {Geographical erasure in language generation}.
\newblock In \emph{Findings of the Association for Computational Linguistics: EMNLP 2023}, pages 12310--12324, Singapore. Association for Computational Linguistics.

\bibitem[{Shao et~al.(2024)Shao, Jiang, Kanell, Xu, Khattab, and Lam}]{shao-etal-2024-assisting}
Yijia Shao, Yucheng Jiang, Theodore Kanell, Peter Xu, Omar Khattab, and Monica Lam. 2024.
\newblock \href {https://doi.org/10.18653/v1/2024.naacl-long.347} {Assisting in writing {W}ikipedia-like articles from scratch with large language models}.
\newblock In \emph{Proceedings of the 2024 Conference of the North American Chapter of the Association for Computational Linguistics: Human Language Technologies (Volume 1: Long Papers)}, pages 6252--6278, Mexico City, Mexico. Association for Computational Linguistics.

\bibitem[{Simhi et~al.(2025)Simhi, Itzhak, Barez, Stanovsky, and Belinkov}]{simhi2025trustmeimwrong}
Adi Simhi, Itay Itzhak, Fazl Barez, Gabriel Stanovsky, and Yonatan Belinkov. 2025.
\newblock \href {https://arxiv.org/abs/2502.12964} {Trust me, i'm wrong: High-certainty hallucinations in llms}.
\newblock \emph{Preprint}, arXiv:2502.12964.

\bibitem[{Singh et~al.(2025)Singh, Romanou, Fourrier, Adelani, Ngui, Vila-Suero, Limkonchotiwat, Marchisio, Leong, Susanto, Ng, Longpre, Ko, Ruder, Smith, Bosselut, Oh, Martins, Choshen, Ippolito, Ferrante, Fadaee, Ermis, and Hooker}]{G-MMLU_paper}
Shivalika Singh, Angelika Romanou, Clémentine Fourrier, David~I. Adelani, Jian~Gang Ngui, Daniel Vila-Suero, Peerat Limkonchotiwat, Kelly Marchisio, Wei~Qi Leong, Yosephine Susanto, Raymond Ng, Shayne Longpre, Wei-Yin Ko, Sebastian Ruder, Madeline Smith, Antoine Bosselut, Alice Oh, Andre F.~T. Martins, Leshem Choshen, Daphne Ippolito, Enzo Ferrante, Marzieh Fadaee, Beyza Ermis, and Sara Hooker. 2025.
\newblock \href {https://arxiv.org/abs/2412.03304} {Global mmlu: Understanding and addressing cultural and linguistic biases in multilingual evaluation}.
\newblock \emph{Preprint}, arXiv:2412.03304.

\bibitem[{Snyder et~al.(2024)Snyder, Moisescu, and Zafar}]{snyder_early_2024}
Ben Snyder, Marius Moisescu, and Muhammad~Bilal Zafar. 2024.
\newblock \href {https://doi.org/10.1145/3637528.3671796} {On early detection of hallucinations in factual question answering}.
\newblock In \emph{Proceedings of the 30th ACM SIGKDD Conference on Knowledge Discovery and Data Mining}, KDD '24, page 2721–2732, New York, NY, USA. Association for Computing Machinery.

\bibitem[{Song et~al.(2025)Song, Li, Lothritz, Ezzini, Sleem, Gentile, State, Bissyandé, and Klein}]{song2025llmsilverbulletlowresource}
Yewei Song, Lujun Li, Cedric Lothritz, Saad Ezzini, Lama Sleem, Niccolo Gentile, Radu State, Tegawendé~F. Bissyandé, and Jacques Klein. 2025.
\newblock \href {https://arxiv.org/abs/2503.24102} {Is llm the silver bullet to low-resource languages machine translation?}
\newblock \emph{Preprint}, arXiv:2503.24102.

\bibitem[{ul~Islam et~al.(2025)ul~Islam, Lauscher, and Glavaš}]{islam2025llmshallucinatelanguagesmultilingual}
Saad~Obaid ul~Islam, Anne Lauscher, and Goran Glavaš. 2025.
\newblock \href {https://arxiv.org/abs/2502.12769} {How much do llms hallucinate across languages? on multilingual estimation of llm hallucination in the wild}.
\newblock \emph{Preprint}, arXiv:2502.12769.

\bibitem[{Vazquez et~al.(2025)Vazquez, Mickus, Zosa, Vahtola, Tiedemann, Sinha, Segonne, Sanchez~Vega, Raganato, Libovick{\'y}, Karlgren, Ji, Helcl, Guillou, De~Gibert, Bengoetxea, Attieh, and Apidianaki}]{vazquez-etal-2025-semeval}
Raul Vazquez, Timothee Mickus, Elaine Zosa, Teemu Vahtola, J{\"o}rg Tiedemann, Aman Sinha, Vincent Segonne, Fernando Sanchez~Vega, Alessandro Raganato, Jind{\v{r}}ich Libovick{\'y}, Jussi Karlgren, Shaoxiong Ji, Jind{\v{r}}ich Helcl, Liane Guillou, Ona De~Gibert, Jaione Bengoetxea, Joseph Attieh, and Marianna Apidianaki. 2025.
\newblock \href {https://aclanthology.org/2025.semeval-1.322/} {{S}em{E}val-2025 task 3: Mu-{SHROOM}, the multilingual shared-task on hallucinations and related observable overgeneration mistakes}.
\newblock In \emph{Proceedings of the 19th International Workshop on Semantic Evaluation (SemEval-2025)}, pages 2472--2497, Vienna, Austria. Association for Computational Linguistics.

\bibitem[{Wang et~al.(2025)Wang, Adel, Lange, Liu, Nie, Str{\"o}tgen, and Schuetze}]{wang-etal-2025-lost-multilinguality}
Mingyang Wang, Heike Adel, Lukas Lange, Yihong Liu, Ercong Nie, Jannik Str{\"o}tgen, and Hinrich Schuetze. 2025.
\newblock \href {https://aclanthology.org/2025.acl-long.253/} {Lost in multilinguality: Dissecting cross-lingual factual inconsistency in transformer language models}.
\newblock In \emph{Proceedings of the 63rd Annual Meeting of the Association for Computational Linguistics (Volume 1: Long Papers)}, pages 5075--5094, Vienna, Austria. Association for Computational Linguistics.

\bibitem[{Xiao and Wang(2021)}]{xiao-wang-2021-hallucination}
Yijun Xiao and William~Yang Wang. 2021.
\newblock \href {https://doi.org/10.18653/v1/2021.eacl-main.236} {On hallucination and predictive uncertainty in conditional language generation}.
\newblock In \emph{Proceedings of the 16th Conference of the European Chapter of the Association for Computational Linguistics: Main Volume}, pages 2734--2744, Online. Association for Computational Linguistics.

\bibitem[{Zhou et~al.(2022)Zhou, Ethayarajh, and Jurafsky}]{zhou2022richer}
Kaitlyn Zhou, Kawin Ethayarajh, and Dan Jurafsky. 2022.
\newblock Richer countries and richer representations.
\newblock \emph{arXiv preprint arXiv:2205.05093}.

\end{thebibliography}
